\documentclass[acmsmall,nonacm]{acmart}

\AtBeginDocument{%
  }

\acmJournal{CSUR}
\setcopyright{none}
\acmYear{2026}
\acmDOI{}
\usepackage{booktabs}
\usepackage{xurl}
\usepackage{subcaption}
\usepackage{wrapfig}
\usepackage{needspace}
\usepackage{placeins}
\usepackage{multirow}
\usepackage{xcolor}
\usepackage{enumitem}
\usepackage{tabularx}
\usepackage{colortbl}
\usepackage{tcolorbox}
\usepackage{pifont}
\usepackage{bm}
\usepackage{bbm}
\newcommand{\ind}{\mathbbm{1}}
\tcbuselibrary{breakable,skins}
\usetikzlibrary{calc,arrows.meta,decorations.pathreplacing,decorations.text,positioning}

\definecolor{groupAcolor}{HTML}{2563EB}    
\definecolor{groupOcolor}{HTML}{7C3AED}    
\definecolor{groupRcolor}{HTML}{0891B2}    

\definecolor{lcink}{HTML}{1F2933}
\definecolor{lcmute}{HTML}{8A94A3}
\newcommand{\lcF}{\tikz[baseline=-0.62ex]\fill[lcink] (0,0) circle (0.58ex);}
\newcommand{\lcH}{\tikz[baseline=-0.62ex]{\fill[lcink] (0,0) -- (90:0.58ex) arc (90:270:0.58ex) -- cycle;
  \draw[lcink, line width=0.35pt] (0,0) circle (0.58ex);}}
\newcommand{\lcN}{\tikz[baseline=-0.62ex]\draw[lcmute, line width=0.45pt] (0,0) circle (0.58ex);}

\definecolor{tabband}{HTML}{EEF2F7}   
\definecolor{gainc}{HTML}{15803D}     
\newcommand{\gain}[1]{\textcolor{gainc}{$+$#1}}
\newcommand{\nr}{\textcolor{gray}{n/r}}      
\newsavebox{\fittablebox}
\newenvironment{fittable}{\begin{lrbox}{\fittablebox}}{%
\end{lrbox}\ifdim\wd\fittablebox>\linewidth
\resizebox{\linewidth}{!}{\usebox{\fittablebox}}%
\else\usebox{\fittablebox}\fi}
\newcommand{\na}{\textcolor{gray}{---}}      
\newlength{\qbarmax}

\begin{document}

\title{Large Language Models for AI Infrastructure}

\author{Bolin Shen}
\email{blshen@fsu.edu}
\affiliation{%
  \institution{Florida State University}
  \department{Computer Science}
  \city{Tallahassee}
  \state{FL}
  \country{USA}
}

\author{Shaoyi Huang}
\email{shuang59@stevens.edu}
\affiliation{%
  \institution{Stevens Institute of Technology}
  \department{Computer Science}
  \city{Hoboken}
  \state{NJ}
  \country{USA}
}

\author{Yushun Dong}
\email{yd24f@fsu.edu}
\authornote{Corresponding author.}
\affiliation{%
  \institution{Florida State University}
  \department{Computer Science}
  \city{Tallahassee}
  \state{FL}
  \country{USA}
}


\begin{abstract}
Large language models (LLMs) are increasingly used to design, optimize, and operate the infrastructure that supports AI training and serving, yet the literature remains fragmented across hardware design, machine learning systems, electronic design automation, and high-performance computing. We present a comprehensive survey of LLM-assisted AI infrastructure across this lifecycle, from accelerator design and kernel engineering to compilation, deployment, operations, data management, and energy optimization. We organize the literature by the primary resource managed by the LLM into three groups: artifact-centric methods for computational artifacts, operational-centric methods for deployment context and system state, and resource-centric methods for data and energy. Their categories span the stages of the lifecycle and locate AI-specific concerns such as distributed-training configuration, inference serving, GPU-cluster scheduling, and data loading. Building on this taxonomy, we conduct a cross-study quantitative synthesis of the performance reported for LLM-driven methods. Because individual studies adopt heterogeneous benchmarks, baselines, hardware platforms, and metrics, we compare methods horizontally only where evaluation conditions permit and align their results along common dimensions: functional correctness, execution performance, throughput, resource and energy consumption, operational outcomes, and optimization cost. Each result retains its native metric, baseline, and evaluation setting rather than being collapsed into a single score. This multidimensional comparison shows where LLM-driven methods yield consistent gains, how the size of reported gains in workload orchestration depends on the evaluation environment, workload, and reference, from simulators to real clusters and production, and where evidence is still lacking. Together, the taxonomy, the lifecycle perspective, and the quantitative evaluation provide an evidence-grounded reference for LLM-driven engineering across the AI infrastructure lifecycle.

\end{abstract}



\maketitle


\section{Introduction}
\label{sec:intro}

Large language models (LLMs)~\cite{zhao2023llmsurvey,minaee2024large} are
increasingly studied as tools for designing, optimizing, and operating the
infrastructure used to train and serve AI models, which spans accelerator
hardware, graphics processing unit (GPU) kernels, compilers and parallel
runtimes, deployment and operational systems, and the data and energy that AI
workloads consume. On the artifact side, LLMs generate register-transfer-level
(RTL) designs from natural-language
specifications~\cite{liu2023verilogeval,lu2023rtllm,chang2023chipgpt,liu2023rtlcoder},
synthesize and optimize GPU
kernels~\cite{kernelbench2025,tritonbench2025,wen2025multikernelbench}, tune
compiler optimizations~\cite{cummins2023compiler,lin2025awarecompiler,chen2026magellan},
and translate programs across parallel programming
models~\cite{bitan2025unipar,davis2025parevalrepo,jhaveri2025accellium}. On the
operational side, they deploy research artifacts and generate software build
recipes~\cite{wang2026deploybench,rp251105626}, detect and remediate Kubernetes
misconfigurations~\cite{malul2024genkubesec}, diagnose failures in production
and AI clusters~\cite{yang2026tsguard,shi2024clusterdiag,sun2026metakube}, and,
most recently, make the runtime decisions specific to AI workloads:
parallelism and sharding plans for distributed
training~\cite{ding2025asap,raina2026primus}, routing and cache policies for
inference serving~\cite{hamadanian2025glia,poudel2026cachecraft}, and
GPU-cluster scheduling~\cite{wang2025semantic,novikov2025alphaevolve}. On the
resource side, they build data-preparation pipelines for model
training~\cite{dataflow2025prep}, catalog data lakes and extract structured
records~\cite{an2025ledd,arora2023evaporate}, generate energy-efficient
code~\cite{peng2024energycode,dearing2026lassiee}, reduce the power of RTL
designs and deep-learning accelerators~\cite{ping2026poet,rp260617461}, and
forecast data-center cooling load~\cite{jiang2025hyperload}.

\begin{wrapfigure}{l}{0.33\textwidth}
\centering
\includegraphics[width=\linewidth]{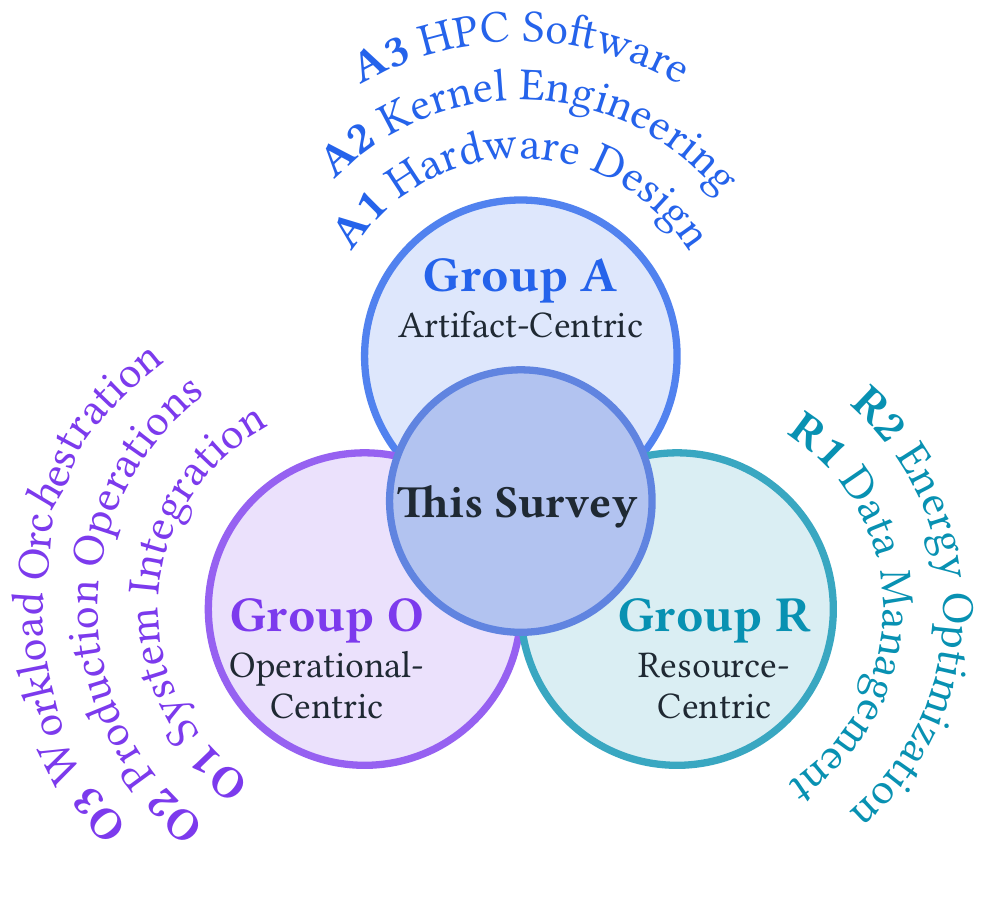}
\caption{Scope of this survey across artifact-centric, operational-centric, and resource-centric AI infrastructure.}
\Description{Three overlapping circles labeled Group A artifact-centric, Group O operational-centric, and Group R resource-centric surround a central circle labeled This Survey. Curved labels along each circle list its categories: A1 to A3 above Group A, O1 to O3 to the left of Group O, and R1 and R2 to the right of Group R.}
\label{fig:position}
\end{wrapfigure}

These results establish a broad research agenda rather than evidence that LLMs
can autonomously replace infrastructure specialists, and the literature behind
them remains fragmented across machine learning, electronic design automation
(EDA), computer systems, and high-performance computing (HPC). Existing surveys examine individual layers in
depth, but they organize work by model adaptation, design stage, programming
task, or operational workflow, and none follows an AI model through the
infrastructure that builds, runs, and sustains it. We therefore view AI
infrastructure as a lifecycle: hardware and software artifacts are first
built, training and serving workloads are then operated on them, and data and
energy are supplied to those workloads throughout. Along this lifecycle, we
characterize each LLM-driven method by the resource that the LLM's principal
action manages, the feedback available to evaluate that action, and the
outcome used to assess it. This view yields a taxonomy whose three groups
correspond to the three phases of the lifecycle, and it allows methods to be
compared across infrastructure layers while preserving differences in their
experimental evidence.

The proposed taxonomy comprises three groups and eight categories.
\emph{\textcolor{groupAcolor}{Artifact-centric}} work (Group~A) generates or modifies computational
artifacts: \emph{\textcolor{groupAcolor}{Hardware Design}} (A1), \emph{\textcolor{groupAcolor}{Kernel Engineering}} (A2), and
\emph{\textcolor{groupAcolor}{HPC Software}} (A3). Compilers, simulators, numerical tests, and execution
harnesses provide feedback for these tasks. Their verdicts are conditional on
the test coverage and execution environment; successful evaluation is not a
general correctness guarantee.
\emph{\textcolor{groupOcolor}{Operational-centric}} work (Group~O) uses deployment context or system
state for \emph{\textcolor{groupOcolor}{System Integration}} (O1), \emph{\textcolor{groupOcolor}{Production Operations}}
(O2), and \emph{\textcolor{groupOcolor}{Workload Orchestration}} (O3). Its outputs include configurations,
diagnoses, operational actions, and the parallelization, serving, and scheduling
policies that determine how training and serving workloads use a cluster.
\emph{\textcolor{groupRcolor}{Resource-centric}} work (Group~R) addresses \emph{\textcolor{groupRcolor}{Data Management}} (R1)
and \emph{\textcolor{groupRcolor}{Energy Optimization}} (R2), including the quality and resource costs
of data processing and the energy consequences of generated code or policies.
Figure~\ref{fig:position} summarizes these concerns. A study may address more
than one concern; its primary category follows its principal contribution,
with secondary objectives retained in the discussion.

This infrastructure-wide perspective complements specialized surveys by
connecting design, deployment, operation, and resource efficiency within one
analytical framework. A lifecycle map (Section~\ref{sec:prelim:lifecycle})
places every category on seven stages from accelerator design to power and
marks which stage hand-offs LLM methods have evaluated, and a table in
Appendix~\ref{app:ai-concerns} places every AI-specific concern, including distributed training, inference
serving, GPU-cluster scheduling, and data loading, on the same stages.
Table~\ref{tab:survey-scope} situates the resulting coverage against selected
related surveys. Surveys of training and serving systems are marked as related
to O3 because they review systems built for AI workloads rather than LLMs that
engineer such systems.

\noindent In summary, this survey makes three contributions:
\begin{itemize}[leftmargin=1.2em, topsep=2pt, itemsep=2pt]
  \item \textbf{A cross-lifecycle survey of LLMs for AI infrastructure.} To our
        knowledge, it is the first survey that spans the lifecycle of training and
        serving AI models, organized as seven stages from accelerator design, kernel
        development, and compilation through deployment and serving and cluster
        operation to data supply and power management. It covers
        the method families, benchmarks, and AI-specific concerns of every stage,
        connecting work that has so far been studied one stage at a time into a
        single end-to-end view of the lifecycle.

  \item \textbf{A lifecycle-aligned taxonomy of LLM-driven methods.} The
        taxonomy organizes the field into eight categories across three groups:
        artifact-centric methods generate hardware, kernels, and parallel
        software; operational-centric methods integrate, operate, and orchestrate
        infrastructure; and resource-centric methods manage data and energy.
        Grouping methods by what the LLM manages keeps those with shared feedback
        and evaluation together and aligns each group with a lifecycle phase.

  \item \textbf{A quantitative analysis of LLMs' practical impact on AI
        infrastructure.} It characterizes the effectiveness and efficiency gains
        reported across artifact-centric, operational-centric, and
        resource-centric settings, and it compares LLM-driven workload
        orchestration across studies, separating simulated, real-cluster, and
        production evidence.
\end{itemize}

\noindent\textbf{Scope.}
This survey covers work in which an LLM generates an infrastructure artifact,
proposes an infrastructure action, or interprets system evidence. General
compiler, HPC, and data-infrastructure studies are included when their methods
apply to infrastructure used by AI workloads, but their results are not
presented as measurements on AI workloads unless the source evaluates that
setting; systems that optimize LLM execution without an LLM acting as the
engineering agent are treated as context. The review aims to cover every
method family, benchmark, and AI-specific concern in each category rather than
to enumerate every eligible paper, so its coverage of fast-moving subfields is
broad but not exhaustive. Quantitative comparisons include only studies that
report a recoverable metric, result, and evaluation setting.

\noindent\textbf{Organization.}
Section~\ref{sec:prelim} introduces the lifecycle of AI infrastructure and
the common task formulation. Section~\ref{sec:taxonomy}
develops the taxonomy, places the AI-specific concerns within it, and reviews
methods by family. Section~\ref{sec:experiments} synthesizes reported evaluations.
Section~\ref{sec:related} discusses related surveys, and
Section~\ref{sec:conclusion} presents conclusions and research priorities.


\section{Background and Problem Formulation}
\label{sec:prelim}

\subsection{The AI Infrastructure Lifecycle}
\label{sec:prelim:lifecycle}
\label{sec:prelim:infra}
AI infrastructure comprises the hardware, software, and operational services
that train and serve AI models at scale, extending from accelerators and the
software stack that programs them to the management of clusters and the supply
of data and power. It differs from general-purpose computing infrastructure in
the two workloads it must sustain. \emph{Distributed training} partitions a
model and its data across many accelerators and must remain efficient and fault
tolerant over long-running jobs~\cite{duan2024trainsurvey}, whereas
\emph{inference serving} must meet latency and throughput objectives under
fluctuating demand~\cite{miao2023servingsurvey}. Both workloads share
accelerator clusters, and their efficiency depends on tightly coupled decisions
across hardware, software, operations, and resource supply rather than on any
single component.

\begin{figure}[t]
\centering
\includegraphics[width=\linewidth]{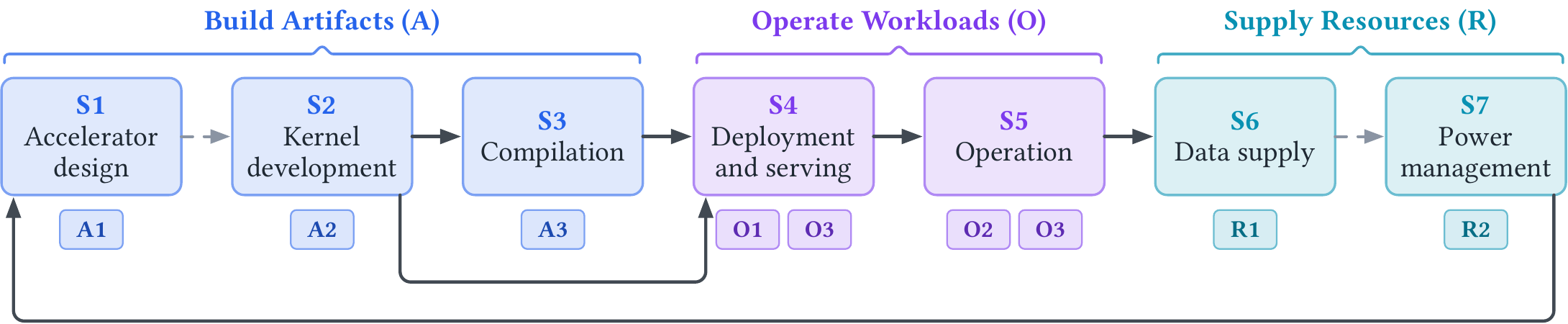}
\caption{Lifecycle of AI infrastructure engineering. Seven stages form three phases that
place the artifact-centric (A), operational-centric (O), and resource-centric (R) groups on the
lifecycle; the chips below each stage list the categories whose primary contribution falls in
it. Solid arrows mark hand-offs at which the output of an LLM-driven method in one stage has been
evaluated in the next; dashed arrows mark hand-offs without such an end-to-end evaluation.}
\Description{A left-to-right flowchart of seven stages: S1 accelerator design, S2 kernel development, and S3 compilation under Build Artifacts (A); S4 deployment and serving and S5 operation under Operate Workloads (O); S6 data supply and S7 power management under Supply Resources (R). Category chips below the stages read A1, A2, A3, O1 and O3, O2 and O3, R1, and R2. Arrows connect consecutive stages; the arrows from S1 to S2 and from S6 to S7 are dashed. Two edges below the row lead from S2 to S4 and from S7 back to S1.}
\label{fig:lifecycle}
\end{figure}

We therefore organize AI infrastructure as a lifecycle of seven stages that
follows a model from the hardware that executes it to the resources it
consumes (Figure~\ref{fig:lifecycle}). \emph{Accelerator design} (S1)
produces the circuits that execute AI operators, \emph{kernel development} (S2)
implements these operators for a target device, and \emph{compilation} (S3)
lowers model graphs into executable programs. \emph{Deployment and serving}
(S4) partitions a model across devices and configures its runtime policies,
and \emph{operation} (S5) schedules jobs on shared clusters and diagnoses and
recovers from failures. \emph{Data supply} (S6) prepares and loads the data
that workloads consume, and \emph{power management} (S7) accounts for and
reduces their energy demand. The stages form three phases, namely building
artifacts (S1--S3), operating workloads (S4--S5), and supplying resources
(S6--S7), which correspond to the artifact-centric, operational-centric, and
resource-centric groups of the taxonomy in Section~\ref{sec:taxonomy}; each
category is placed on the stage of its primary contribution.

Consecutive stages are linked by hand-offs, in which the output of one stage
becomes the input of the next. Two further links complete the lifecycle:
serving kernels are frequently integrated into inference engines as libraries,
which connects S2 directly to S4, and the energy measured in S7 informs the
design objectives of the next accelerator generation, which closes the cycle
from S7 to S1. In Figure~\ref{fig:lifecycle}, a hand-off is drawn solid when the
output of an LLM-driven method in one stage has been evaluated in the next, and
dashed when no such end-to-end evaluation exists. The two cross-stage edges are
supported by FlashInfer-Bench~\cite{xing2026flashinferbench} and
AgentCompile~\cite{rp260607665}, which evaluate generated kernels inside serving
engines (S2 to S4), and by POET~\cite{ping2026poet} and
AUTOGATE~\cite{rp260617461}, which optimize RTL and accelerator designs for
power (S7 to S1). Current work concentrates on the
build-artifacts phase. LLM-driven methods for the runtime decisions of S4 and S5
are emerging, whereas concerns such as checkpointing, straggler mitigation,
quantization choice, and data loading remain largely unexplored
(Appendix~\ref{app:ai-concerns}). Most methods also operate within a single
stage, and the links from accelerator design to kernel development and from
data supply to power management have yet to be examined.


\providecolor{posink}{HTML}{1F2933}
\providecolor{posneutral}{HTML}{5C6675}

\newcommand{\covfull}[1]{\tikz[baseline=-0.5ex]{\fill[#1] (0,0) circle (0.62ex);}}
\newcommand{\covpart}[1]{%
  \tikz[baseline=-0.5ex]{%
    \fill[#1] (0,0) -- (90:0.62ex) arc (90:270:0.62ex) -- cycle;
    \draw[#1, line width=0.35pt] (0,0) circle (0.62ex);}}
\newcommand{\covnone}{\tikz[baseline=-0.5ex]{\draw[black!32, line width=0.35pt] (0,0) circle (0.62ex);}}
\newcommand{\fA}{\covfull{groupAcolor}}   \newcommand{\pA}{\covpart{groupAcolor}}
\newcommand{\fO}{\covfull{groupOcolor}}   \newcommand{\pO}{\covpart{groupOcolor}}
\newcommand{\fR}{\covfull{groupRcolor}}   \newcommand{\pR}{\covpart{groupRcolor}}
\newcommand{\nn}{\covnone}

\providecommand{\posnohyph}{\hyphenpenalty=10000 \exhyphenpenalty=10000
                            \tolerance=9999 \emergencystretch=1em\relax}

\providecommand{\poschip}[4]{\textbf{\textcolor{#1}{#2}} #3\\#4}

\begin{table}[t]
\centering
\footnotesize
\caption{Principal scope of selected related surveys under the proposed taxonomy.
\protect\covfull{posink}~primary topic; \protect\covpart{posink}~related coverage;
\protect\covnone~not a principal topic. Colors identify the three groups.}
\label{tab:survey-scope}
\setlength{\tabcolsep}{5pt}
\renewcommand{\arraystretch}{1.12}
\begin{tabularx}{\linewidth}{@{}>{\raggedright\arraybackslash}p{2.6cm}>{\raggedright\arraybackslash}p{1.5cm}ccc@{\hskip 8pt}ccc@{\hskip 8pt}cc@{\hskip 12pt}>{\raggedright\arraybackslash\posnohyph}X@{}}
\toprule
& &
\multicolumn{3}{c@{\hskip 8pt}}{\shortstack{\textcolor{groupAcolor}{\textbf{Group A}}\\[-1pt]\scriptsize artifact}} &
\multicolumn{3}{c@{\hskip 8pt}}{\shortstack{\textcolor{groupOcolor}{\textbf{Group O}}\\[-1pt]\scriptsize operational}} &
\multicolumn{2}{c}{\shortstack{\textcolor{groupRcolor}{\textbf{Group R}}\\[-1pt]\scriptsize resource}} & \\
\cmidrule(lr){3-5}\cmidrule(lr){6-8}\cmidrule(lr){9-10}
\textbf{Survey} & \textbf{Venue}
& \textbf{A1} & \textbf{A2} & \textbf{A3}
& \textbf{O1} & \textbf{O2} & \textbf{O3}
& \textbf{R1} & \textbf{R2}
& \textbf{Primary scope} \\
\midrule
Pan et al.~\cite{pan2025llmeda}
  & arXiv '25
  & \fA & \nn & \nn & \nn & \nn & \nn & \nn & \nn
  & LLMs for electronic design automation \\
Yang et al.~\cite{yang2026verilog}
  & CSUR '26
  & \fA & \nn & \nn & \nn & \nn & \nn & \nn & \nn
  & Verilog code generation \\
Yu et al.~\cite{yu2026kernel}
  & arXiv '26
  & \nn & \fA & \nn & \nn & \nn & \nn & \nn & \nn
  & GPU kernel generation \\
Ljaljevi\'{c} et al.~\cite{ljaljevic2026hpc}
  & FGCS '26
  & \nn & \pA & \fA & \nn & \nn & \nn & \nn & \nn
  & LLMs in HPC programming \\
Jiang et al.~\cite{jiang2024codellm}
  & arXiv '24
  & \pA & \pA & \pA & \nn & \nn & \nn & \nn & \nn
  & LLMs for code generation \\
Pahl et al.~\cite{pahl2026iac}
  & Electron.\ '26
  & \nn & \nn & \nn & \fO & \pO & \nn & \nn & \nn
  & AI for infrastructure-as-code \\
Zhang et al.~\cite{zhang2025aiops}
  & CSUR '25
  & \nn & \nn & \nn & \pO & \fO & \nn & \nn & \nn
  & AIOps in the era of LLMs \\
Duan et al.~\cite{duan2024trainsurvey}
  & arXiv '24
  & \nn & \nn & \pA & \nn & \nn & \pO & \nn & \nn
  & Systems for distributed LLM training \\
Miao et al.~\cite{miao2023servingsurvey}
  & arXiv '23
  & \nn & \pA & \nn & \nn & \nn & \pO & \nn & \nn
  & Systems for efficient LLM serving \\
Hong et al.~\cite{hong2025text2sql}
  & TKDE '25
  & \nn & \nn & \nn & \nn & \nn & \nn & \pR & \nn
  & LLM-based Text-to-SQL \\
Ali et al.~\cite{masthanali2026sustainable}
  & arXiv '26
  & \nn & \nn & \nn & \nn & \nn & \nn & \nn & \fR
  & Sustainable code generation \\
\midrule
\textbf{This survey}
  & ---
  & \fA & \fA & \fA & \fO & \fO & \fO & \fR & \fR
  & \textbf{LLMs across the AI infrastructure lifecycle} \\
\bottomrule
\end{tabularx}
\end{table}

\subsection{A Common Task Formulation}
\label{sec:prelim:formal}
We represent an infrastructure task by $(\mathcal{I},\mathcal{A},\mathcal{E},F,J)$,
where $\mathcal{I}$ is the space of task inputs, such as specifications,
existing code, telemetry, and resource constraints; $\mathcal{A}$ is the space
of candidate actions, such as programs, configuration edits, and diagnoses;
$\mathcal{E}$ is the environment in which an action is evaluated, including
tools, hardware, and system state; $F$ is the feedback mechanism; and $J$ is
the task-specific cost. For an input $I\in\mathcal{I}$, evaluating a candidate
$a\in\mathcal{A}$ yields feedback $f=F(\mathcal{E},a;I)$, such as diagnostics,
test outcomes, or measured performance, and the task is the constrained
optimization
\begin{equation}
  \min_{a\in\mathcal{A}} J(\mathcal{E},a;I)
  \quad\text{subject to}\quad
  \mathrm{Feasible}(\mathcal{E},a;I)=\top ,
  \label{eq:problem}
\end{equation}
where the predicate $\mathrm{Feasible}$ indicates whether $a$ satisfies the
stated correctness, security, and service requirements and $\top$ denotes
true. Tests and measurements provide evidence for this predicate only to the
extent of their coverage. More generally, the feedback $F$ observed by a method
is a proxy for $\mathrm{Feasible}$ and $J$: it may rely on visible rather than
held-out tests, or on a simulator, surrogate, or recorded trace in place of the
deployment environment, so a candidate that scores well under $F$ may still
violate the requirements or incur a higher cost when deployed. Benefits such as accuracy or throughput enter $J$
through a change of sign or a corresponding loss, and multi-objective studies
may retain separate costs instead of a scalar
weighting~\cite{miettinen1998multiobjective}.

A method evaluates a sequence of candidates $a_1,\ldots,a_k$ within an attempt
budget $k$ and observes $f_i=F(\mathcal{E},a_i;I)$ for each candidate; later
candidates may depend on earlier observations, and one-shot generation
corresponds to $k=1$. The observations need not be independent or
deterministic, since timing noise, concurrency, and changing system state can
affect them. The mechanism by which the LLM proposes these candidates
constitutes the method's \emph{adaptation strategy}, which ranges over
prompting and agentic loops, retrieval augmentation, supervised fine-tuning,
reinforcement learning, evolutionary or tree search, and multi-agent
collaboration. The formulation thus provides a common vocabulary across tasks
without assuming a common optimizer, and the comparison tables of
Section~\ref{sec:taxonomy} characterize each system by $\mathcal{I}$,
$\mathcal{A}$, $\mathcal{E}$, $F$, $J$, and its adaptation strategy.


\FloatBarrier
\section{Landscape of LLM-Assisted AI Infrastructure Engineering}
\label{sec:taxonomy}

The formulation in Equation~\eqref{eq:problem} identifies common components
without equating their implementations. This section organizes LLM-driven methods by the primary
managed resource, places the AI-specific
concerns of Section~\ref{sec:prelim:lifecycle} within that organization, and
compares the inputs, action spaces, feedback mechanisms, and adaptation
strategies of the systems in each category.

\subsection{Taxonomy Overview}
\label{sec:taxonomy:rule}

\textbf{Classification rule.}
A study's primary category is determined by its principal contribution and
main evaluation objective. The classification is an analytical convention;
individual systems may combine artifact generation, deployment, and resource
optimization. Programming language, lifecycle stage, and model-adaptation
strategy are retained as attributes rather than used as the top-level
partition. Figure~\ref{fig:taxonomy-framework} shows the hierarchy with every
system it covers, Appendix~\ref{app:taxonomy-table} provides the category
definitions, and Tables~\ref{tab:cmp-a1}--\ref{tab:cmp-r2} compare the systems
of each category along the components of Section~\ref{sec:prelim:formal}.


\usetikzlibrary{calc}

\providecommand{\chipfont}{\fontsize{6.2}{7.2}\selectfont}
\providecommand{\chipsep}{\hskip 1.5pt plus 0.5pt minus 0.25pt\relax}
\providecommand{\chipbody}{\chipfont\lineskip=1.9pt\lineskiplimit=1000pt\relax}

\definecolor{chipink}{HTML}{1F2933}       
\definecolor{treeneutral}{HTML}{5C6675}   

\tikzset{
  refchip/.style={rectangle, rounded corners=1.5pt, line width=0.3pt,
                  inner xsep=2.1pt, inner ysep=1.2pt, outer sep=0pt,
                  text=chipink, font=\chipfont}
}

\providecommand{\refchip}[3]{%
  \begin{tikzpicture}[baseline=(rc.base)]%
    \node[refchip, fill=#1, draw=#2] (rc) {#3};%
  \end{tikzpicture}\chipsep}
\providecommand{\chipA}[1]{\refchip{groupAcolor!16}{groupAcolor!55}{#1}}
\providecommand{\chipO}[1]{\refchip{groupOcolor!16}{groupOcolor!55}{#1}}
\providecommand{\chipR}[1]{\refchip{groupRcolor!16}{groupRcolor!55}{#1}}

\begin{figure}[t]
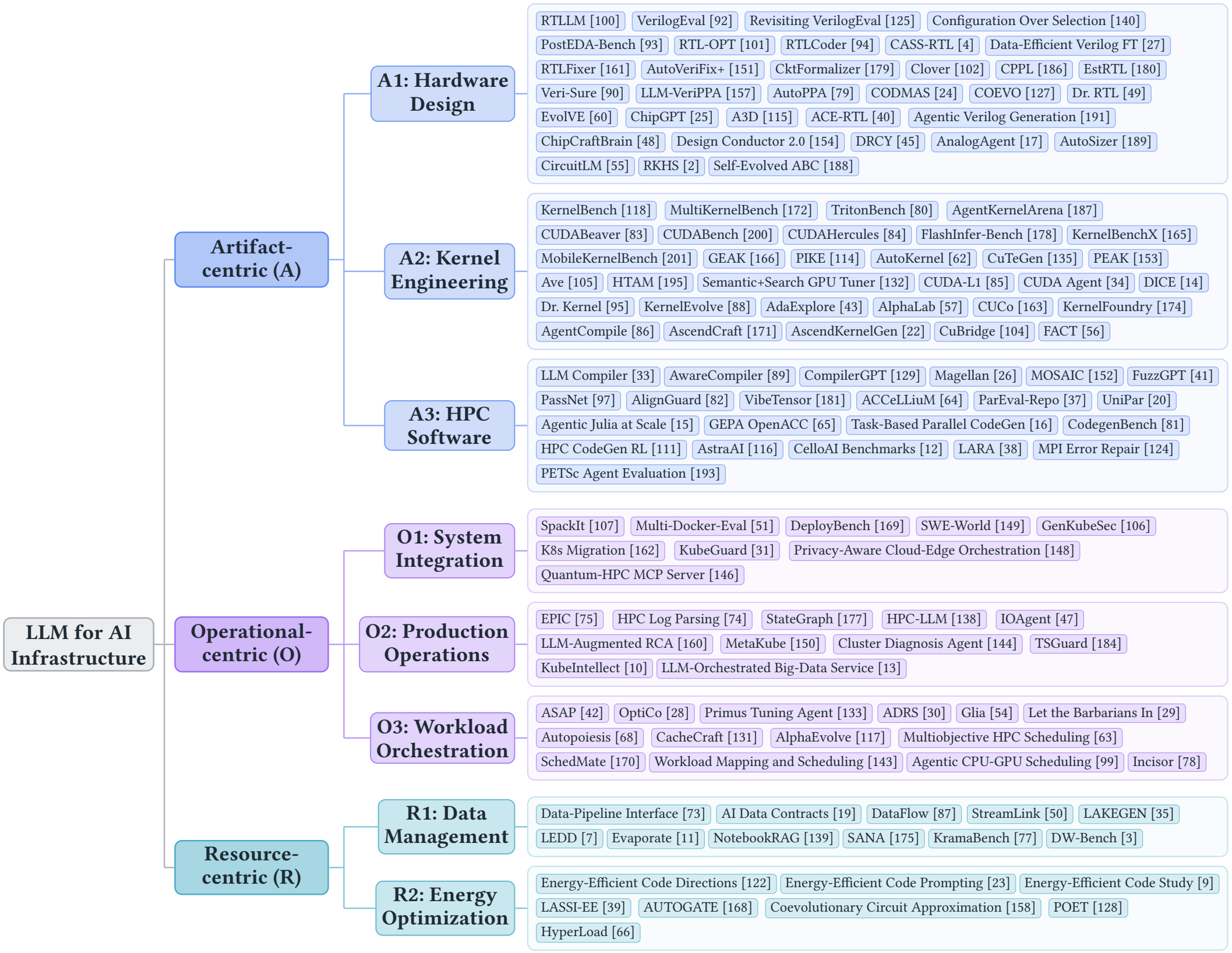

\centering
\hbadness=10000
\resizebox{\linewidth}{!}{%
\begin{tikzpicture}[
  every node/.style={font=\small},
  root/.style={rectangle, rounded corners=3pt, line width=0.6pt,
               draw=treeneutral!55, fill=treeneutral!12, text=chipink,
               minimum width=2.05cm, minimum height=0.72cm,
               inner xsep=3pt, inner ysep=2.5pt, align=center, font=\small},
  group/.style={rectangle, rounded corners=3pt, line width=0.6pt,
                text=chipink, minimum width=2.30cm, minimum height=0.68cm,
                inner xsep=3pt, inner ysep=2.5pt, align=center, font=\small},
  groupA/.style={group, fill=groupAcolor!36, draw=groupAcolor!75},
  groupO/.style={group, fill=groupOcolor!36, draw=groupOcolor!75},
  groupR/.style={group, fill=groupRcolor!36, draw=groupRcolor!75},
  category/.style={rectangle, rounded corners=3pt, line width=0.5pt,
                   text=chipink, minimum width=1.95cm, minimum height=0.68cm,
                   inner xsep=2.5pt, inner ysep=2.5pt, align=center, font=\small},
  catA/.style={category, fill=groupAcolor!22, draw=groupAcolor!62},
  catO/.style={category, fill=groupOcolor!22, draw=groupOcolor!62},
  catR/.style={category, fill=groupRcolor!22, draw=groupRcolor!62},
  panel/.style={rectangle, rounded corners=3pt, line width=0.5pt,
                align=left, text width=10.26cm,
                inner xsep=3.5pt, inner ysep=3.2pt,
                font=\chipbody},
  panelA/.style={panel, fill=groupAcolor!3, draw=groupAcolor!25},
  panelO/.style={panel, fill=groupOcolor!3, draw=groupOcolor!25},
  panelR/.style={panel, fill=groupRcolor!3, draw=groupRcolor!25},
  rootedge/.style={line width=0.6pt, draw=treeneutral!50},
  edgeA/.style={line width=0.6pt, draw=groupAcolor!50},
  edgeO/.style={line width=0.6pt, draw=groupOcolor!50},
  edgeR/.style={line width=0.6pt, draw=groupRcolor!50}
]

\node[panelA, anchor=north west] (a1papers) at (6.65,0) {%
  \input{sections/tables/chips_A1}
};

\node[panelA, anchor=north west] (a2papers) at ([yshift=-0.13cm]a1papers.south west) {%
  \input{sections/tables/chips_A2}
};

\node[panelA, anchor=north west] (a3papers) at ([yshift=-0.13cm]a2papers.south west) {%
  \input{sections/tables/chips_A3}
};

\node[panelO, anchor=north west] (o1papers) at ([yshift=-0.24cm]a3papers.south west) {%
  \input{sections/tables/chips_O1}
};

\node[panelO, anchor=north west] (o2papers) at ([yshift=-0.13cm]o1papers.south west) {%
  \input{sections/tables/chips_O2}
};

\node[panelO, anchor=north west] (o3papers) at ([yshift=-0.13cm]o2papers.south west) {%
  \input{sections/tables/chips_O3}
};

\node[panelR, anchor=north west] (r1papers) at ([yshift=-0.24cm]o3papers.south west) {%
  \input{sections/tables/chips_R1}
};

\node[panelR, anchor=north west] (r2papers) at ([yshift=-0.13cm]r1papers.south west) {%
  \input{sections/tables/chips_R2}
};

\node[catA, anchor=east] (a1) at ([xshift=-0.18cm]a1papers.west) {\textbf{A1: Hardware}\\[-1pt]\textbf{Design}};
\node[catA, anchor=east] (a2) at ([xshift=-0.18cm]a2papers.west) {\textbf{A2: Kernel}\\[-1pt]\textbf{Engineering}};
\node[catA, anchor=east] (a3) at ([xshift=-0.18cm]a3papers.west) {\textbf{A3: HPC}\\[-1pt]\textbf{Software}};

\node[catO, anchor=east] (o1) at ([xshift=-0.18cm]o1papers.west) {\textbf{O1: System}\\[-1pt]\textbf{Integration}};
\node[catO, anchor=east] (o2) at ([xshift=-0.18cm]o2papers.west) {\textbf{O2: Production}\\[-1pt]\textbf{Operations}};
\node[catO, anchor=east] (o3) at ([xshift=-0.18cm]o3papers.west) {\textbf{O3: Workload}\\[-1pt]\textbf{Orchestration}};

\node[catR, anchor=east] (r1) at ([xshift=-0.18cm]r1papers.west) {\textbf{R1: Data}\\[-1pt]\textbf{Management}};
\node[catR, anchor=east] (r2) at ([xshift=-0.18cm]r2papers.west) {\textbf{R2: Energy}\\[-1pt]\textbf{Optimization}};

\path let \p1=(a1.west), \p2=(a2.west), \p3=(a3.west),
          \p4=(o1.west), \p5=(o2.west), \p6=(r1.west), \p7=(r2.west), \p8=(o3.west)
      in coordinate (categoryLeft) at
         ({min(\x1,\x2,\x3,\x4,\x5,\x6,\x7,\x8)},0);
\coordinate (categoryBusX) at ([xshift=-0.225cm]categoryLeft);
\coordinate (groupAmid) at ($(a1.center)!0.5!(a3.center)$);
\coordinate (groupOmid) at ($(o1.center)!0.5!(o3.center)$);
\coordinate (groupRmid) at ($(r1.center)!0.5!(r2.center)$);
\node[groupA, anchor=east] (groupA) at ([xshift=-0.45cm]categoryLeft |- groupAmid) {\textbf{Artifact-}\\[-1pt]\textbf{centric (A)}};
\node[groupO, anchor=east] (groupO) at ([xshift=-0.45cm]categoryLeft |- groupOmid) {\textbf{Operational-}\\[-1pt]\textbf{centric (O)}};
\node[groupR, anchor=east] (groupR) at ([xshift=-0.45cm]categoryLeft |- groupRmid) {\textbf{Resource-}\\[-1pt]\textbf{centric (R)}};

\path let \p1=(groupA.west), \p2=(groupO.west), \p3=(groupR.west)
      in coordinate (groupLeft) at ({min(\x1,\x2,\x3)},0);
\coordinate (rootBusX) at ([xshift=-0.15cm]groupLeft);
\node[root, anchor=east] (root) at ([xshift=-0.30cm]groupLeft |- groupO.center) {\textbf{LLM for AI}\\\textbf{Infrastructure}};

\coordinate (rootbus) at (rootBusX |- root.east);
\draw[rootedge] (root.east) -- (rootbus);
\draw[rootedge] (rootbus |- groupA.west) -- (rootbus |- groupR.west);
\draw[rootedge] (rootbus |- groupA.west) -- (groupA.west);
\draw[rootedge] (rootbus |- groupO.west) -- (groupO.west);
\draw[rootedge] (rootbus |- groupR.west) -- (groupR.west);

\coordinate (groupAbus) at (categoryBusX |- groupA.east);
\draw[edgeA] (groupA.east) -- (groupAbus);
\draw[edgeA] (groupAbus |- a1.west) -- (groupAbus |- a3.west);
\draw[edgeA] (groupAbus |- a1.west) -- (a1.west);
\draw[edgeA] (groupAbus |- a2.west) -- (a2.west);
\draw[edgeA] (groupAbus |- a3.west) -- (a3.west);

\coordinate (groupObus) at (categoryBusX |- groupO.east);
\draw[edgeO] (groupO.east) -- (groupObus);
\draw[edgeO] (groupObus |- o1.west) -- (groupObus |- o3.west);
\draw[edgeO] (groupObus |- o1.west) -- (o1.west);
\draw[edgeO] (groupObus |- o2.west) -- (o2.west);
\draw[edgeO] (groupObus |- o3.west) -- (o3.west);

\coordinate (groupRbus) at (categoryBusX |- groupR.east);
\draw[edgeR] (groupR.east) -- (groupRbus);
\draw[edgeR] (groupRbus |- r1.west) -- (groupRbus |- r2.west);
\draw[edgeR] (groupRbus |- r1.west) -- (r1.west);
\draw[edgeR] (groupRbus |- r2.west) -- (r2.west);

\draw[edgeA] (a1.east) -- (a1papers.west);
\draw[edgeA] (a2.east) -- (a2papers.west);
\draw[edgeA] (a3.east) -- (a3papers.west);

\draw[edgeO] (o1.east) -- (o1papers.west);
\draw[edgeO] (o2.east) -- (o2papers.west);
\draw[edgeO] (o3.east) -- (o3papers.west);

\draw[edgeR] (r1.east) -- (r1papers.west);
\draw[edgeR] (r2.east) -- (r2papers.west);

\end{tikzpicture}%
}
\caption{Our taxonomy organizes LLM-driven AI infrastructure by its primary managed resource into
\textcolor{groupAcolor}{artifact-centric}, \textcolor{groupOcolor}{operational-centric},
and \textcolor{groupRcolor}{resource-centric} groups and eight categories. Each category is expanded with all systems reviewed in it, ordered by method family as in
Tables~\ref{tab:cmp-a1}--\ref{tab:cmp-r2}.}
\Description{A left-to-right taxonomy tree. LLM for AI Infrastructure branches into Artifact-centric, Operational-centric, and Resource-centric groups; these groups branch into eight categories, each of which holds a grid of representative-work blocks.}
\label{fig:taxonomy-framework}
\end{figure}

\textbf{Groups and categories.}
Artifact-centric work (Group~A) produces or evaluates computational artifacts.
Hardware Design (A1) concerns RTL, high-level synthesis (HLS), circuits, and
EDA decisions. Kernel Engineering (A2) concerns device-level kernels and
accelerator transformations. HPC Software (A3) concerns compiler decisions,
parallel programs, deep-learning frameworks, and distributed scientific
software. Operational-centric work (Group~O) depends on deployment context or
system observations. System Integration (O1) connects artifacts to an
execution environment, Production Operations (O2) interprets the state of
deployed systems to diagnose and remediate it, and Workload Orchestration (O3)
decides how training and serving workloads use shared accelerators, memory,
and interconnect over time. O3 covers parallelization and sharding plans,
request routing and batching, KV-cache and expert-placement policies, job
scheduling and placement, and fault-tolerance policies; its outputs are
policies or plans evaluated by workload-level performance under a trace or
live load. Resource-centric work (Group~R) emphasizes data processing and
access in Data Management (R1), or energy, power, and thermal objectives in
Energy Optimization (R2). R1 includes data-quality tasks that support
resource-aware processing, even when the study does not directly optimize
bytes transferred.

\textbf{Assignment of boundary cases.}
The main objective takes precedence over auxiliary measurements. RTL
generation with power as one of several PPA metrics remains in A1, whereas
code generation primarily evaluated by energy savings belongs to R2. Diagnosis
of an I/O incident from traces belongs to O2, while data extraction or access
planning belongs to R1. Compiler-generated accelerator code is assigned by the
contribution evaluated: kernel construction in A2, or general compiler
heuristics in A3. Parallel translation may target both host and accelerator
code and is assigned to A3 when cross-paradigm translation is the central
contribution. The three operational categories are separated by the kind of
decision and its evaluation. A one-time configuration judged by whether the
service deploys and runs belongs to O1; a diagnosis or explanation of observed
behavior belongs to O2, including the diagnosis of training failures; and a
plan or policy judged by performance under load, such as a parallelism
configuration, a routing policy, or a scheduling heuristic, belongs to O3.
Kernels and compiler passes written for serving engines remain in A2 and A3,
while a cluster-level partitioning of a model is an O3 decision. Frameworks
that report several infrastructure case studies are assigned by the case with
the strongest infrastructure evidence, and their other cases are discussed as
secondary contributions. These rules preserve a primary category without
asserting that the underlying technical problems are disjoint.

\textbf{Placement of AI-specific workloads.}
The lifecycle of Section~\ref{sec:prelim:lifecycle} is orthogonal to this
resource-based partition, so each AI-specific concern is placed where its
managed resource lies. Decisions about how training and serving workloads
occupy a cluster, including parallelism and sharding, request routing,
batching, KV-cache eviction, model placement, and GPU-cluster scheduling, are
O3 policies; fused communication kernels and serving runtimes are artifacts in
A2 and A3; the diagnosis of training failures interprets operational evidence
in O2; and training-data preparation and loading belong to R1. Systems
evaluated on an AI model, workload, or cluster are starred in
Tables~\ref{tab:cmp-a1}--\ref{tab:cmp-r2}, and
Appendix~\ref{app:ai-concerns} maps every concern to its stage, category,
and the LLM-driven systems that address it.

\textbf{Reading the comparison tables.}
Tables~\ref{tab:cmp-a1}--\ref{tab:cmp-r2} instantiate the formulation of
Section~\ref{sec:prelim:formal} for every system. The input
$\mathcal{I}$ distinguishes specifications, existing artifacts, telemetry, and
constraints; the action space $\mathcal{A}$ records what the LLM emits; the
environment $\mathcal{E}$ records where candidate actions are executed or
checked; the feedback $F$ lists the signals available to the method during
generation, which may differ from the signals used only to evaluate it; and
the column $J$ records the outcomes that the evaluation reports. In this
column, Corr denotes a reported correctness rate, which measures the
$\mathrm{Feasible}$ predicate of Equation~\eqref{eq:problem} rather than a
cost; systems that enforce correctness without reporting its rate list only
their cost. The
adaptation column records how the LLM proposes candidates: prompting,
retrieval, fine-tuning, reinforcement learning, search, or multi-agent
decomposition. Section~\ref{sec:experiments} compares reported results where
task sets, baselines, budgets, and hardware permit.

\subsection{Artifact-Centric Automation}
\label{sec:artifact}\label{sec:compute}

Artifact-centric methods propose code or design transformations and evaluate
them through compilers, simulators, numerical checks, or execution harnesses.
The input $I$ is a specification or existing artifact, and $a$ is a candidate
or edit. Feedback $F$ may include acceptance diagnostics, test outcomes, and
measured costs. Repeated evaluation is often feasible in a controlled
environment, enabling repair and search. Its reliability depends on test
coverage, toolchain versions, and measurement stability; a machine-generated
verdict is not necessarily deterministic or complete. This group comprises
hardware design, kernel engineering, and HPC software.

\subsubsection{Hardware Design (A1)}
\label{sec:hw}

\textbf{Task and formulation.}
Hardware-design applications generate or analyze RTL, HLS specifications,
formal properties, testbenches, and circuit descriptions. Their evaluation
uses simulation, synthesis, timing analysis, formal verification, or circuit
simulation~\cite{liu2023verilogeval,lu2023rtllm,pan2025llmeda}.
Syntax acceptance and simulation success provide distinct checks. Neither
alone establishes that a design satisfies its intended specification or is
ready for fabrication. For validated candidates, a possible scalarization of
power, performance, and area (PPA) is
\begin{equation}
 \min_a\;\alpha P(a)+\beta/f_{\max}(a)+\gamma A(a)
 \quad\text{subject to}\quad\mathrm{Feasible}(\mathcal{E},a;I)=\top,
 \label{eq:ppa}
\end{equation}
where $P(a)$ is design power, $f_{\max}(a)$ is the maximum clock frequency,
and $A(a)$ is the area. The nonnegative weights $\alpha$, $\beta$, and $\gamma$
include the normalization required to combine these units. This expression
illustrates a design trade-off; studies may instead use constraints or Pareto
comparisons~\cite{miettinen1998multiobjective}. Correctness checking and PPA
estimation can require separate tools and different evaluation costs.

\textbf{Benchmarks and evaluation protocols.}
VerilogEval and RTLLM established specification-to-RTL evaluation with
simulation-based functional checks and, in RTLLM, an additional design-quality
goal~\cite{liu2023verilogeval,lu2023rtllm}. Subsequent benchmarks refine the
protocol itself. Revisiting VerilogEval~\cite{pinckney2024revisiting} revises
the task format and adds failure classification, so that results on the
original and revised suites are not interchangeable, and Configuration Over
Selection~\cite{rp260417102} sweeps decoding configurations across open models
to show that inference-time settings can outweigh the choice of model. Two
newer suites move beyond functional generation: RTL-OPT~\cite{rp260101765}
pairs suboptimal designs with human-optimized references to measure PPA
improvement, and PostEDA-Bench~\cite{rp260506936} evaluates agents on
post-synthesis design-rule repair and PPA convergence with machine-checkable
EDA toolchains. Together, these benchmarks shift A1 evaluation from code that
simulates correctly toward designs that close timing, power, and sign-off
checks.

\textbf{Domain-adapted and internally guided models.}
RTLCoder~\cite{liu2023rtlcoder} fine-tunes a compact open model on
synthesized instruction--code pairs. Data-efficient
fine-tuning~\cite{rp260415388} generates testbenches with a multi-agent
workflow so that only verified pairs are used for supervision.
CASS-RTL~\cite{rp260605680} leaves the weights unchanged: it identifies
attention heads whose activations separate correct from incorrect RTL and
steers them at inference time. These methods change the proposal distribution
rather than the evaluation loop.

\textbf{Verification-guided generation and repair.}
A second family strengthens the feedback channel. RTLFixer~\cite{rp231116543}
combines retrieval with ReAct-style~\cite{yao2023react} prompting over compiler
diagnostics to repair syntax errors. AutoVeriFix+~\cite{rp260311489} first
derives a Python reference model and then uses concolic testing with
cycle-accurate traces to expose corner cases and prune functionally redundant
logic. Clover~\cite{luo2026clover} casts RTL repair as a stochastic
tree-of-thoughts search that dispatches subproblems either to LLM agents or to
symbolic solvers. Veri-Sure~\cite{contracts2025ai} aligns agents through an
explicit design contract and combines trace-driven temporal analysis with
formal verification, while EstRTL~\cite{rp260609867} estimates functional
correctness statically before deciding whether to regenerate or repair. Other
systems change the representation that the LLM edits:
CktFormalizer~\cite{rp260507782} embeds a typed hardware language in Lean so
that compiler and proof-state feedback guide repair and equivalence proving,
and CPPL~\cite{rp260517892} routes generation through a JSON circuit
intermediate representation (IR) that is checked and lowered to CIRCT. These
methods share a common direction: feedback that arrives earlier, is more
structured, and is formal where possible.

\textbf{PPA optimization and search.}
LLM-VeriPPA~\cite{thorat2025verippa} adds a PPA-oriented optimization stage
after functional repair. Search-based systems instead explore populations of
candidates: EvolVE~\cite{hsin2026evolve} compares evolution strategies and
pairs Monte Carlo tree search (MCTS)~\cite{browne2012mcts} for functional
correctness with idea-guided refinement for optimization;
COEVO~\cite{ping2026coevo} retains partially correct but architecturally
promising candidates in a Pareto search over correctness, area, delay, and
power; and AutoPPA~\cite{li2026autoppa} induces reusable optimization rules by
contrasting generated code pairs. CODMAS~\cite{rp260317204} uses two dialectic
agents to plan and verify pipelining and clock-gating transformations, and
Dr.~RTL~\cite{rp260414989} performs closed-loop timing optimization with an
industrial synthesis flow, combining critical-path analysis, parallel RTL
rewriting, and a library of reusable optimization skills. The evaluation
environments of this family range from manually degraded textbook circuits to
industrial flows.

\textbf{Agentic design flows and accelerator design.}
ChipGPT~\cite{chang2023chipgpt} first combined generation, output management,
and candidate selection in a zero-code design flow. ACE-RTL~\cite{deng2026acertl}
couples an RTL-specialized model with a frontier reasoning model through
generator, reflector, and coordinator roles, and
ChipCraftBrain~\cite{eryilmaz2026chipcraftbrain} orchestrates six specialized
agents with a learned policy and solves logic-minimization subproblems
symbolically. A controlled study~\cite{rp260319347} on the
CVDP~\cite{pinckney2025cvdp} benchmark examines how prompting structure and
tool design affect agentic Verilog generation and shows that unstructured agent
loops can underperform carefully prompted single-pass generation. Agentic flows
also extend beyond single modules. A3D~\cite{nallathambi2026a3d} automates
workload analysis, HLS refactoring, and micro-architecture exploration for
scientific accelerators, and DRCY~\cite{rp260315672} reviews schematic
connections against component datasheets within a production design-review
workflow. Design Conductor 2.0~\cite{rp260505170} provides the most direct
AI-specific example: starting from the TurboQuant~\cite{zandieh2025turboquant}
paper, its multi-agent harness autonomously designed VerTQ, an LLM inference
accelerator. AlphaEvolve's Verilog simplification of a tensor processing unit
(TPU) arithmetic circuit, discussed with its primary O3 contribution, is a
second example of LLM-driven design for AI
accelerators~\cite{novikov2025alphaevolve}.

\textbf{EDA tools, analog, and circuit design.}
Another line of work optimizes EDA tools rather than designs. Self-Evolved
ABC~\cite{rp260415082} employs LLM agents to rewrite components of the ABC
logic-synthesis codebase within a loop driven by correctness and quality of
results, and RKHS~\cite{wang2026energykernel} synthesizes HLS list-scheduling
heuristics from compact templates with retrieval and iterative refinement.
Analog design requires circuit-level feedback: AnalogAgent~\cite{bao2026analogagent}
distills SPICE feedback into a self-evolving memory shared by generation,
optimization, and knowledge-curation agents; AutoSizer~\cite{yu2026autosizer}
embeds numerical transistor sizing in an LLM-driven meta-optimizer that
reshapes the search space from simulation feedback; and
CircuitLM~\cite{rp260104505} grounds schematic generation in a component
knowledge base and checks the result with an electrical-rule checker and an
LLM judge.

\textbf{Limitations.}
The principal evaluation limitation is the gap between a finite test suite and
the intended design contract. Underspecified prompts and weak testbenches can
leave functional errors undetected, and benchmark revisions, decoding
configuration, and candidate budgets change reported success rates. Evaluation
on AI accelerators remains scarce in A1, and the transfer of these methods to
the datapaths and memory systems of training and inference hardware is largely
untested. The quantitative synthesis therefore retains benchmark
identity, the meaning of syntax and functional rates, and search budgets
(Section~\ref{sec:exp-artifact}). Table~\ref{tab:cmp-a1} compares the A1 systems.

\begin{table}[tp]
\centering
\caption{Hardware design (A1) systems under the task formulation of Eq.~(\ref{eq:problem}), grouped by method family. $\mathcal{I}$: S specification, C existing code or artifact, T telemetry or traces, K constraints; $\mathcal{A}$: Code program or design code, Cfg configuration, Diag diagnosis or answer; $\mathcal{E}$: CPU or GPU execution on that device, RTL RTL simulator or formal tool, EDA synthesis or physical-design flow, SPICE circuit simulator, Prod production system, Off offline corpus; $F$: St static check, Co compilation or synthesis, Si simulation, Ex execution, LJ LLM judge, Hu human; $J$: Corr correctness, PPA power-performance-area, Perf runtime or throughput, Acc accuracy, Succ task success; Adapt.: P prompting, RAG retrieval, SFT fine-tuning, RL reinforcement learning, Evo evolutionary or tree search, MA multi-agent, Int internal-state intervention; --- none. $^{\star}$ evaluation on an AI model, workload, or cluster.}
\label{tab:cmp-a1}
\scriptsize
\setlength{\tabcolsep}{2.6pt}
\renewcommand{\arraystretch}{1.04}
\begin{tabularx}{\linewidth}{@{}>{\raggedright\arraybackslash}X c>{\raggedright\arraybackslash}p{0.75cm}>{\raggedright\arraybackslash}p{1.05cm}>{\raggedright\arraybackslash}p{1.25cm}>{\raggedright\arraybackslash}p{1.05cm}>{\raggedright\arraybackslash}p{1.25cm}>{\raggedright\arraybackslash}p{1.25cm}@{}}
\toprule
\textbf{System} & \textbf{Year} & $\boldsymbol{\mathcal{I}}$ & $\boldsymbol{\mathcal{A}}$ & $\boldsymbol{\mathcal{E}}$ & $\boldsymbol{F}$ & $\boldsymbol{J}$ & \textbf{Adapt.}\\
\midrule
\rowcolor{tabband}\multicolumn{8}{@{}l}{\textit{Benchmarks and evaluation protocols}}\\
RTLLM~\cite{lu2023rtllm} & 2023 & S & Code & RTL, EDA & --- & Corr, PPA & P\\
VerilogEval~\cite{liu2023verilogeval} & 2023 & S & Code & RTL & --- & Corr & P, SFT\\
Revisiting VerilogEval~\cite{pinckney2024revisiting} & 2024 & S & Code & RTL & --- & Corr & P\\
Configuration Over Selection~\cite{rp260417102} & 2026 & S & Code & RTL, EDA & --- & Corr & P\\
PostEDA-Bench~\cite{rp260506936} & 2026 & C, T & Code, Cfg & EDA & Co & Succ & P, MA\\
RTL-OPT~\cite{rp260101765} & 2026 & C & Code & RTL, EDA & --- & PPA & P\\
\addlinespace[1pt]
\rowcolor{tabband}\multicolumn{8}{@{}l}{\textit{Domain-adapted and internally guided models}}\\
RTLCoder~\cite{liu2023rtlcoder} & 2023 & S & Code & RTL & --- & Corr & SFT\\
CASS-RTL~\cite{rp260605680} & 2026 & S & Code & RTL & --- & Corr & Int\\
Data-Efficient Verilog FT~\cite{rp260415388} & 2026 & S & Code & RTL & Si & Corr & SFT, MA\\
\addlinespace[1pt]
\rowcolor{tabband}\multicolumn{8}{@{}l}{\textit{Verification-guided generation and repair}}\\
RTLFixer~\cite{rp231116543} & 2023 & C, T & Code & RTL & Co & Corr & RAG, P\\
AutoVeriFix+~\cite{rp260311489} & 2026 & S & Code & RTL & Co, Si & Corr & P, MA\\
CktFormalizer~\cite{rp260507782} & 2026 & S & Code & RTL, EDA & St, Co, Si & Corr & MA\\
Clover~\cite{luo2026clover} & 2026 & C, T & Code & RTL & St, Si & Corr & Evo, MA\\
CPPL~\cite{rp260517892} & 2026 & S & Code & RTL & St, Co & Corr & P\\
EstRTL~\cite{rp260609867} & 2026 & S & Code & RTL & LJ & Corr & MA\\
Veri-Sure~\cite{contracts2025ai} & 2026 & S & Code & RTL & St, Si & Corr & MA\\
\addlinespace[1pt]
\rowcolor{tabband}\multicolumn{8}{@{}l}{\textit{PPA optimization and search}}\\
LLM-VeriPPA~\cite{thorat2025verippa} & 2025 & S & Code & RTL, EDA & Co, Si & Corr, PPA & P\\
AutoPPA~\cite{li2026autoppa} & 2026 & C & Code & RTL, EDA & Co, Si & PPA & Evo\\
CODMAS~\cite{rp260317204} & 2026 & C & Code & RTL, EDA & Co, Si & PPA & MA\\
COEVO~\cite{ping2026coevo} & 2026 & S & Code & RTL, EDA & Si, Co & Corr, PPA & Evo\\
Dr.\ RTL~\cite{rp260414989} & 2026 & C & Code & EDA & Co & PPA & MA\\
EvolVE~\cite{hsin2026evolve} & 2026 & S & Code & RTL, EDA & Si, Co & Corr, PPA & Evo\\
\addlinespace[1pt]
\rowcolor{tabband}\multicolumn{8}{@{}l}{\textit{Agentic design flows and accelerator design}}\\
ChipGPT~\cite{chang2023chipgpt} & 2023 & S & Code & EDA & Co & PPA & P\\
A3D~\cite{nallathambi2026a3d} & 2026 & C & Code & EDA & Co, Ex & Perf & MA, RAG\\
ACE-RTL~\cite{deng2026acertl} & 2026 & S & Code & RTL & Si & Corr & SFT, MA\\
Agentic Verilog Generation~\cite{rp260319347} & 2026 & S & Code & RTL & Si, Co & Corr & P\\
ChipCraftBrain~\cite{eryilmaz2026chipcraftbrain} & 2026 & S & Code & RTL, EDA & Si, Co & Corr & MA, RL, RAG\\
Design Conductor 2.0$^{\star}$~\cite{rp260505170} & 2026 & S & Code & RTL, EDA & Si, Co & Corr, PPA & MA\\
DRCY~\cite{rp260315672} & 2026 & C & Diag & Prod & St, Hu & Acc & MA, RAG\\
\addlinespace[1pt]
\rowcolor{tabband}\multicolumn{8}{@{}l}{\textit{EDA tools, analog, and circuit design}}\\
AnalogAgent~\cite{bao2026analogagent} & 2026 & S & Code & SPICE & Si & Corr & MA\\
AutoSizer~\cite{yu2026autosizer} & 2026 & S, K & Cfg & SPICE & Si & Succ & P\\
CircuitLM~\cite{rp260104505} & 2026 & S & Code & Off & St, LJ & Acc & MA, RAG\\
RKHS~\cite{wang2026energykernel} & 2026 & S & Code & EDA & Ex & Perf & RAG, P\\
Self-Evolved ABC~\cite{rp260415082} & 2026 & C & Code & EDA & Co, Ex & PPA & MA, Evo\\
\bottomrule
\end{tabularx}
\end{table}

\subsubsection{Kernel Engineering (A2)}
\label{sec:ker}

\textbf{Task and formulation.}
Kernel methods generate or optimize CUDA, Triton, HIP, CuTe, SYCL, AscendC,
and other device-level implementations, mostly of deep-learning operators.
Evaluation combines numerical comparison against a reference with
execution-time measurements. KernelBench defines $\mathrm{fast}_p$ as the
fraction of tasks with a correct candidate whose speedup exceeds threshold
$p$~\cite{kernelbench2025}. Its extension to a budget of $k$ candidates can be
written as
\begin{equation}
 \mathrm{fast}_p@k=\frac{1}{|\mathcal D|}
 \sum_{d\in\mathcal D}\ind\!\left[
 \exists i\leq k:\mathrm{Correct}_d(a_{d,i})\land
 \frac{T_{\mathrm{ref},d}}{T_d(a_{d,i})}>p\right].
 \label{eq:fastk}
\end{equation}
Here $\mathcal D$ is the evaluated task set, $|\mathcal D|$ its size, and $d$
indexes a task. Candidate $a_{d,i}$ is the $i$th proposal for task $d$,
$\mathrm{Correct}_d(a_{d,i})$ is its numerical-test verdict, and $T_d(a_{d,i})$
and $T_{\mathrm{ref},d}$ are the candidate and reference runtimes. The
indicator $\ind[\cdot]$ equals one when its condition holds and zero
otherwise. Thus $p$ denotes the speedup threshold and $k$ the candidate budget. For iterative refinement, candidates depend on earlier
feedback rather than being independent samples. A correctness test
establishes agreement only for the specified inputs and numerical tolerances;
broader numerical auditing remains necessary~\cite{sarkar2026correctness}.
A2 is among the most active categories, and many of its studies evaluate AI operators
or serving engines.

\textbf{Benchmark construction.}
KernelBench, TritonBench, and MultiKernelBench defined the first task suites
for CUDA, Triton, and multi-platform kernels,
respectively~\cite{kernelbench2025,tritonbench2025,wen2025multikernelbench}.
Later benchmarks address the failure modes that these suites revealed.
KernelBenchX~\cite{wang2026kernelbenchx} organizes Triton tasks into
categories to attribute correctness and efficiency to task structure rather
than to method, CUDABench~\cite{rp260302236} evaluates text-to-CUDA generation
with a roofline-based performance score, and
CUDAHercules~\cite{rp260508467} compares generated code with end-to-end expert
systems across three GPU generations. AgentKernelArena~\cite{rp260516819}
evaluates complete agent workflows rather than single calls and tests
generalization to input configurations not observed during optimization.
CUDABeaver~\cite{rp260508455} isolates debugging and identifies a failure mode
that correctness-only benchmarks reward, termed repair by degeneration, in
which a repair method passes the tests by replacing an optimized kernel with a
simpler and slower one. MobileKernelBench~\cite{rp260311935} extends
evaluation to the CPU backend of a mobile inference framework with on-device
verification. FlashInfer-Bench~\cite{xing2026flashinferbench} connects kernel
generation to deployment: it derives tasks from real serving traces and
substitutes generated kernels into SGLang~\cite{zheng2024sglang} and
vLLM~\cite{kwon2023vllm} through a single interface. Benchmark design has thus
progressed along three axes: from single calls to agents, from correctness to
performance against expert or production references, and from fixed inputs to
generalization and deployment.

\textbf{Prompting with execution feedback.}
The simplest feedback loop returns compilation, correctness, and timing
results to the model, as in the refinement experiments of
KernelBench~\cite{kernelbench2025}. GEAK~\cite{wu2025geak} applies
Reflexion-style feedback to Triton kernels on AMD GPUs.
CuTeGen~\cite{rp260401489} targets the CuTe abstraction and withholds
low-level profiling until the kernel's high-level structure has stabilized.
AutoKernel~\cite{rp260321331} profiles an entire model, ranks bottlenecks by
their Amdahl impact, and iterates Triton or CUDA implementations behind a
five-stage correctness harness. PIKE~\cite{nagaitsev2025pike} provides a
framework for comparing multi-agent organizations for PyTorch inference
optimization and studies how exploration, exploitation, and error-fixing agents
interact.

\textbf{Profiling- and structure-guided refinement.}
A third family constrains the space of edits available to the model.
PEAK~\cite{tariq2025peak} expresses optimizations as natural-language
transformations applied along human-selected sequences. Ave~\cite{rp260418616}
exposes hardware instructions through a tile-based domain-specific language
(DSL) and uses data-flow invariants checked by a satisfiability-modulo-theories
(SMT) solver as compile-time guardrails that return counterexamples instead of
sparse test failures. A two-stage tuner~\cite{wang2026fact} refactors
SGLang~\cite{zheng2024sglang} kernels into explicit parameter templates and then
tunes the parameters by constrained search, which reduces the randomness of
direct rewriting, and HTAM~\cite{zhang2026htam} organizes optimization
experience in a hierarchical memory of global directions and local
strategies. These systems trade generality for a more informative and safer
feedback signal.

\textbf{RL-trained kernel models.}
Reinforcement learning internalizes the feedback loop in the model.
CUDA-L1~\cite{li2026cudal1} trains with a contrastive RL algorithm that uses
measured speedups as rewards, and CUDA Agent~\cite{rp260224286} scales agentic
RL with a data-synthesis pipeline and a verified profiling environment.
Dr.~Kernel~\cite{rp260205885} studies the failure modes of this paradigm: its
KernelGYM environment detects reward hacking, its turn-level estimator removes
a bias in multi-turn group relative policy optimization
(GRPO)~\cite{shao2024deepseekmath}, and profiling-based rewards discourage
so-called lazy optimizations that remain correct but slow.
DICE~\cite{rp260211715} adapts supervised fine-tuning and staged RL to
diffusion language models, and the GPT-5-RL study~\cite{tehrani2026gpt5}
evaluates a fine-tuned frontier model and the MakoraGenerate agent built around
it. Collectively, these studies indicate that the reward must account for
performance and guard against reward hacking in addition to correctness.

\textbf{Evolutionary and agentic search.}
Search-based systems allocate inference-time computation to populations of
candidates. KernelEvolve~\cite{meta2025kernelevolve} combines agentic search
with retrieval to generate kernels for heterogeneous accelerators in
production at Meta. KernelFoundry~\cite{rp260312440} uses
MAP-Elites~\cite{mouret2015mapelites} quality-diversity search with co-evolved
meta-prompts and template-based parameter tuning, and
AdaExplore~\cite{rp260416625} converts recurring failures into a memory of
validity rules before a tree search that alternates local refinement with
structural regeneration. AlphaLab~\cite{rp260408590} applies a general
autonomous research harness to CUDA optimization among other domains.
CUCo~\cite{rp260302376} is the principal example in this family that targets
distributed training and serving: it co-designs computation and communication
in multi-GPU kernels, converting host-driven calls to the NVIDIA Collective
Communications Library (NCCL) into device-initiated communication on a
correctness-first fast path and then evolving overlap strategies on a slower
search path.

\textbf{Cross-platform transpilation and library-grounded generation.}
Accelerators with little public code require additional structure.
AscendCraft~\cite{rp260122760} generates kernels in a lightweight DSL and
lowers them to AscendC through constrained LLM passes, whereas
AscendKernelGen~\cite{rp260107160} trains a model for Ascend kernels with
chain-of-thought data, supervised fine-tuning, and execution-based RL.
Library-grounded systems reuse optimizations already encoded in expert
libraries. CuBridge~\cite{rp260505023} lifts expert attention kernels into an
executable IR and lowers adapted variants with reference guidance,
FACT~\cite{rp260426666} composes CUTLASS patterns matched to subgraphs of a
traced PyTorch module, and AgentCompile~\cite{rp260607665} combines
compiler-instantiated templates with LLM-generated decode kernels inside a
serving runtime with paged KV cache and continuous batching. GEAK and
MultiKernelBench extend evaluation beyond NVIDIA
devices~\cite{wu2025geak,wen2025multikernelbench}.

\textbf{Limitations.}
Functional validity and improvement over an optimized reference remain
separate objectives. A high pass rate does not imply that generated kernels
outperform TorchInductor~\cite{ansel2024pytorch2} or vendor libraries, and several benchmarks now show
that refinement can trade speed for correctness. Profiling feedback improves
some model--task combinations and reduces scores in others, so the reported
ablations do not support a uniformly beneficial effect. Portability requires
more than compilation on another platform: numerical agreement and performance
should be remeasured. Finally, most A2 results are measured on isolated
operators; only FlashInfer-Bench~\cite{xing2026flashinferbench}, AgentCompile~\cite{rp260607665}, and FACT~\cite{rp260426666} report gains inside
serving engines or model blocks. Table~\ref{tab:cmp-a2} and
Section~\ref{sec:exp-artifact} distinguish these outcomes.

\begin{table}[tp]
\centering
\caption{Kernel engineering (A2) systems under the task formulation of Eq.~(\ref{eq:problem}), grouped by method family. Abbreviations follow Table~\ref{tab:cmp-a1} and the preceding tables; new ones: $\mathcal{E}$: NPU NPU or TPU.}
\label{tab:cmp-a2}
\scriptsize
\setlength{\tabcolsep}{2.6pt}
\renewcommand{\arraystretch}{1.04}
\begin{tabularx}{\linewidth}{@{}>{\raggedright\arraybackslash}X c>{\raggedright\arraybackslash}p{0.75cm}>{\raggedright\arraybackslash}p{1.05cm}>{\raggedright\arraybackslash}p{1.25cm}>{\raggedright\arraybackslash}p{1.05cm}>{\raggedright\arraybackslash}p{1.25cm}>{\raggedright\arraybackslash}p{1.25cm}@{}}
\toprule
\textbf{System} & \textbf{Year} & $\boldsymbol{\mathcal{I}}$ & $\boldsymbol{\mathcal{A}}$ & $\boldsymbol{\mathcal{E}}$ & $\boldsymbol{F}$ & $\boldsymbol{J}$ & \textbf{Adapt.}\\
\midrule
\rowcolor{tabband}\multicolumn{8}{@{}l}{\textit{Benchmark construction}}\\
KernelBench$^{\star}$~\cite{kernelbench2025} & 2025 & C & Code & GPU & Co, Ex & Corr, Perf & P\\
MultiKernelBench$^{\star}$~\cite{wen2025multikernelbench} & 2025 & C & Code & GPU, NPU & --- & Corr, Perf & P\\
TritonBench$^{\star}$~\cite{tritonbench2025} & 2025 & C & Code & GPU & --- & Corr, Perf & P\\
AgentKernelArena$^{\star}$~\cite{rp260516819} & 2026 & C & Code & GPU & Co, Ex & Corr, Perf & MA\\
CUDABeaver~\cite{rp260508455} & 2026 & C, T & Code & GPU & Co, Ex & Corr, Perf & P\\
CUDABench~\cite{rp260302236} & 2026 & S & Code & GPU & --- & Corr, Perf & P\\
CUDAHercules~\cite{rp260508467} & 2026 & S, C & Code & GPU & Co, Ex & Corr, Perf & P\\
FlashInfer-Bench$^{\star}$~\cite{xing2026flashinferbench} & 2026 & C, T & Code & GPU & Co, Ex & Corr, Perf & P, MA\\
KernelBenchX$^{\star}$~\cite{wang2026kernelbenchx} & 2026 & C & Code & GPU & Co, Ex & Corr, Perf & P\\
MobileKernelBench$^{\star}$~\cite{rp260311935} & 2026 & C & Code & CPU & Co, Ex & Corr, Perf & MA\\
\addlinespace[1pt]
\rowcolor{tabband}\multicolumn{8}{@{}l}{\textit{Prompting with execution feedback}}\\
GEAK$^{\star}$~\cite{wu2025geak} & 2025 & S, C & Code & GPU & Co, Ex & Corr, Perf & P\\
PIKE$^{\star}$~\cite{nagaitsev2025pike} & 2025 & C & Code & GPU & Co, Ex & Perf & MA\\
AutoKernel$^{\star}$~\cite{rp260321331} & 2026 & C & Code & GPU & Ex & Corr, Perf & P\\
CuTeGen$^{\star}$~\cite{rp260401489} & 2026 & C & Code & GPU & Co, Ex & Corr, Perf & P\\
\addlinespace[1pt]
\rowcolor{tabband}\multicolumn{8}{@{}l}{\textit{Profiling- and structure-guided refinement}}\\
PEAK$^{\star}$~\cite{tariq2025peak} & 2025 & C & Code & GPU & Co, Ex, Hu & Perf & P\\
Ave$^{\star}$~\cite{rp260418616} & 2026 & C & Code & GPU & St, Ex & Perf & MA\\
HTAM$^{\star}$~\cite{zhang2026htam} & 2026 & C & Code & GPU & Co, Ex & Corr, Perf & Evo\\
Semantic+Search GPU Tuner$^{\star}$~\cite{wang2026fact} & 2026 & C & Code, Cfg & GPU & Ex & Perf & P, Evo\\
\addlinespace[1pt]
\rowcolor{tabband}\multicolumn{8}{@{}l}{\textit{RL-trained kernel models}}\\
CUDA-L1$^{\star}$~\cite{li2026cudal1} & 2025 & C & Code & GPU & Ex & Perf & RL\\
CUDA Agent$^{\star}$~\cite{rp260224286} & 2026 & C & Code & GPU & Co, Ex & Corr, Perf & RL\\
DICE$^{\star}$~\cite{rp260211715} & 2026 & C & Code & GPU & Co, Ex & Corr, Perf & SFT, RL\\
Dr.\ Kernel$^{\star}$~\cite{rp260205885} & 2026 & C & Code & GPU & Co, Ex & Corr, Perf & RL\\
\addlinespace[1pt]
\rowcolor{tabband}\multicolumn{8}{@{}l}{\textit{Evolutionary and agentic search}}\\
KernelEvolve$^{\star}$~\cite{meta2025kernelevolve} & 2025 & C & Code & GPU & Co, Ex & Corr, Perf & Evo, MA\\
AdaExplore$^{\star}$~\cite{rp260416625} & 2026 & C & Code & GPU & Co, Ex & Corr, Perf & Evo\\
AlphaLab$^{\star}$~\cite{rp260408590} & 2026 & S & Code & GPU & Ex & Perf & MA\\
CUCo$^{\star}$~\cite{rp260302376} & 2026 & C & Code & GPU & Co, Ex & Perf & Evo, MA\\
KernelFoundry$^{\star}$~\cite{rp260312440} & 2026 & C & Code & GPU & Ex & Perf & Evo\\
\addlinespace[1pt]
\rowcolor{tabband}\multicolumn{8}{@{}l}{\textit{Cross-platform transpilation and library-grounded generation}}\\
AgentCompile$^{\star}$~\cite{rp260607665} & 2026 & C & Code & GPU & Co, Ex & Perf & P\\
AscendCraft$^{\star}$~\cite{rp260122760} & 2026 & C & Code & NPU & Co, Ex & Corr, Perf & P\\
AscendKernelGen$^{\star}$~\cite{rp260107160} & 2026 & C & Code & NPU & Co, Ex & Corr & SFT, RL\\
CuBridge$^{\star}$~\cite{rp260505023} & 2026 & C & Code & GPU & Co, Ex & Corr, Perf & P\\
FACT$^{\star}$~\cite{rp260426666} & 2026 & C & Code & GPU & Co, Ex & Perf & MA, RAG\\
\bottomrule
\end{tabularx}
\end{table}

\subsubsection{HPC Software (A3)}
\label{sec:hpc}

\textbf{Task and formulation.}
HPC-software applications include compiler optimization, deep-learning
framework engineering, parallel program translation, and the detection or
repair of concurrency errors. GCC, LLVM, tensor compilers, parallel runtimes,
and test harnesses provide complementary observations. In terms of
Equation~\eqref{eq:problem}, $\mathrm{Feasible}$ typically requires that the
generated program or transformation preserve the semantics of the input
program, as checked by tests, and $J$ is runtime, instruction count, or code
size. For parallelization, Amdahl's
law~\cite{amdahl1967validity} gives an idealized bound for a fixed workload,
\begin{equation}
 S(n)=\frac{1}{s+(1-s)/n},
 \label{eq:amdahl}
\end{equation}
where $S(n)$ is speedup relative to serial execution, $s\in[0,1]$ is the
serial fraction, and $n\geq1$ is the processor count. Communication,
synchronization, and memory effects can reduce realized speedup. Improvements
may arise from reducing either serial or parallel work, improving memory
behavior, or changing communication; they are not restricted to reducing $s$.
Correctness assessment must account for concurrency and floating-point effects.

\textbf{Compiler optimization and heuristic synthesis.}
LLM Compiler~\cite{cummins2023compiler} is a model trained to predict compiler
pass sequences that reduce instruction counts. AwareCompiler~\cite{lin2025awarecompiler}
conditions pass generation on program features and structured compiler
knowledge, while Magellan~\cite{chen2026magellan} couples an LLM coding agent
with evolutionary search and autotuning to synthesize LLVM inlining and
register-allocation heuristics. CompilerGPT~\cite{pirkelbauer2025compilergpt}
returns compiler optimization reports to the LLM to guide iterative source-code
rewriting. MOSAIC~\cite{tang2026mosaic} reduces the cost of such search by
having heterogeneous LLMs share one MCTS tree, which records the model that
produced each improvement and favors smaller models. These methods operate on
different optimization targets: pass sequences, decision heuristics,
transformation schedules, and source programs.

\textbf{Deep-learning compilers, frameworks, and runtimes.}
The software stack of AI workloads is itself an engineering target.
PassNet~\cite{chen2026passnet} argues that LLMs should author compiler passes
rather than standalone kernels and provides a dataset, benchmark, and
fine-tuning recipe for pass generation on long-tail fusion subgraphs that
default TorchInductor~\cite{ansel2024pytorch2} compilation handles poorly. A
second line of work addresses testing. FuzzGPT~\cite{rp230402014} primes LLMs
with historical bug-triggering programs to generate unusual inputs for fuzzing
PyTorch and TensorFlow, and AlignGuard~\cite{rp260408720} applies LLM-based test
mutation to detect silent correctness bugs in torch.compile.
VibeTensor~\cite{rp260116238} represents the most autonomous setting: a
PyTorch-style tensor library, autograd engine, and CUDA runtime produced by
coding agents under high-level human guidance and validated mainly by builds
and tests. Its authors report that locally correct subsystems can compose into
globally inefficient software.

\textbf{Parallel translation and directive generation.}
UniPar~\cite{bitan2025unipar} evaluates LLM-based translation among serial,
OpenMP, and CUDA programs on its ParaTrans benchmark;
ACCeLLiuM~\cite{jhaveri2025accellium} fine-tunes models to generate OpenACC
directives for data-parallel loops; and ParEval-Repo~\cite{davis2025parevalrepo}
supplies repository-level translation tasks, for which build systems and
cross-file dependencies remain the main obstacles. GEPA-based prompt
evolution~\cite{rp260108884} iteratively optimizes the prompts used for OpenACC
pragma generation on PolyBench~\cite{pouchet2012polybench} programs without
post-training the model. Other work examines scaling rather than translation:
LLM-generated task-parallel code for OpenMP tasking, C++ standard parallelism,
and HPX is compared for correctness and scalability across
frameworks~\cite{rp260222240}, and agent-generated Julia programs are evaluated
at large core counts, where deadlocks, oversubscription, and memory exhaustion
emerge~\cite{rp260616534}.

\textbf{Performance-rewarded generation and cross-architecture evaluation.}
An online RL approach~\cite{rp260212049} executes generated
matrix-multiplication code on a supercomputer and feeds the measured GFLOPS back
as the reward, with a staged quality-diversity curriculum over optimization
techniques. CodegenBench~\cite{rp260604023} evaluates BLAS routines and
specialized kernels across x86, Sunway, and Kunpeng processors to measure how
generation quality varies with the target architecture.

\textbf{Scientific-software assistants, workflows, and defects.}
An MPI study~\cite{piersall2026mpi} compares prompting and retrieval
strategies for detecting and repairing errors in MPI programs. PETSc Agent
Evaluation~\cite{rp260315976} combines deterministic checks and LLM-based
assessment in a multi-evaluator pipeline to determine whether generated code
uses a production HPC library as an expert would. AstraAI~\cite{rp260327423}
grounds edits to an exascale framework in retrieved code and
abstract-syntax-tree context, CelloAI~\cite{rp260301051} defines repeatable
benchmarks for high-energy-physics and HPC coding assistance, and
LARA~\cite{rp260422571} validates generated atomistic simulation workflows
through dry runs before they consume cluster resources.

\textbf{Limitations.}
No single aggregate of these outcomes measures HPC engineering capability.
Textual agreement with a reference pragma is weaker evidence than executing a
parallel program successfully, and passing available tests is weaker than a
semantic-preservation proof. Performance results additionally depend on
compiler flags, inputs, processor counts, and hardware. The deep-learning
stack is the A3 area most directly related to AI infrastructure, but its
evaluations are still dominated by bug counts and isolated subgraphs rather
than end-to-end training or serving time. The synthesis retains these
distinctions and reports source-specific costs when available
(Tables~\ref{tab:cmp-a3} and~\ref{tab:a3-gaincost}).

\begin{table}[tp]
\centering
\caption{HPC software (A3) systems under the task formulation of Eq.~(\ref{eq:problem}), grouped by method family. Abbreviations follow Table~\ref{tab:cmp-a1} and the preceding tables; new ones: $\mathcal{A}$: Plan plan or schedule; $\mathcal{E}$: HPC HPC system or GPU cluster; $J$: Size code size, Bug defects found, Cost resource cost.}
\label{tab:cmp-a3}
\scriptsize
\setlength{\tabcolsep}{2.6pt}
\renewcommand{\arraystretch}{1.04}
\begin{tabularx}{\linewidth}{@{}>{\raggedright\arraybackslash}X c>{\raggedright\arraybackslash}p{0.75cm}>{\raggedright\arraybackslash}p{1.05cm}>{\raggedright\arraybackslash}p{1.25cm}>{\raggedright\arraybackslash}p{1.05cm}>{\raggedright\arraybackslash}p{1.25cm}>{\raggedright\arraybackslash}p{1.25cm}@{}}
\toprule
\textbf{System} & \textbf{Year} & $\boldsymbol{\mathcal{I}}$ & $\boldsymbol{\mathcal{A}}$ & $\boldsymbol{\mathcal{E}}$ & $\boldsymbol{F}$ & $\boldsymbol{J}$ & \textbf{Adapt.}\\
\midrule
\rowcolor{tabband}\multicolumn{8}{@{}l}{\textit{Compiler optimization and heuristic synthesis}}\\
LLM Compiler~\cite{cummins2023compiler} & 2023 & C & Cfg & CPU & --- & Size & SFT\\
AwareCompiler~\cite{lin2025awarecompiler} & 2025 & C & Cfg & CPU & Co & Size & SFT, RAG\\
CompilerGPT~\cite{pirkelbauer2025compilergpt} & 2025 & C, T & Code & CPU & Co, Ex & Perf & P\\
Magellan~\cite{chen2026magellan} & 2026 & C & Code & CPU & Co, Ex & Perf, Size & Evo\\
MOSAIC$^{\star}$~\cite{tang2026mosaic} & 2026 & C & Code & GPU, CPU & Ex & Perf, Cost & Evo\\
\addlinespace[1pt]
\rowcolor{tabband}\multicolumn{8}{@{}l}{\textit{Deep-learning compilers, frameworks, and runtimes}}\\
FuzzGPT$^{\star}$~\cite{rp230402014} & 2023 & C & Code & CPU, GPU & Ex & Bug & SFT, P\\
PassNet$^{\star}$~\cite{chen2026passnet} & 2026 & C & Code & GPU & Co, Ex & Perf & SFT\\
AlignGuard$^{\star}$~\cite{rp260408720} & 2026 & C & Code & GPU & Ex & Bug & P\\
VibeTensor$^{\star}$~\cite{rp260116238} & 2026 & S & Code & GPU & Co, Ex, Hu & Corr & MA\\
\addlinespace[1pt]
\rowcolor{tabband}\multicolumn{8}{@{}l}{\textit{Parallel translation and directive generation}}\\
ACCeLLiuM~\cite{jhaveri2025accellium} & 2025 & C & Code & Off & --- & Acc & SFT\\
ParEval-Repo~\cite{davis2025parevalrepo} & 2025 & C & Code & GPU & Co, Ex & Corr & P, MA\\
UniPar~\cite{bitan2025unipar} & 2025 & C & Code & CPU, GPU & Co, Ex & Corr, Perf & SFT, P\\
Agentic Julia at Scale~\cite{rp260616534} & 2026 & S & Code & HPC & Co, Ex & Corr, Perf & MA\\
GEPA OpenACC~\cite{rp260108884} & 2026 & C & Code & GPU & Co & Corr, Perf & Evo\\
Task-Based Parallel CodeGen~\cite{rp260222240} & 2026 & S, C & Code & CPU & Ex & Corr, Perf & P\\
\addlinespace[1pt]
\rowcolor{tabband}\multicolumn{8}{@{}l}{\textit{Performance-rewarded generation and cross-architecture evaluation}}\\
CodegenBench~\cite{rp260604023} & 2026 & S & Code & CPU & Co, Ex & Corr, Perf & P\\
HPC CodeGen RL~\cite{rp260212049} & 2026 & S & Code & HPC & Ex & Perf & RL\\
\addlinespace[1pt]
\rowcolor{tabband}\multicolumn{8}{@{}l}{\textit{Scientific-software assistants, workflows, and defects}}\\
AstraAI~\cite{rp260327423} & 2026 & S, C & Code & HPC & --- & --- & RAG\\
CelloAI Benchmarks~\cite{rp260301051} & 2026 & S, C & Code & HPC & Ex & Corr, Acc & P\\
LARA~\cite{rp260422571} & 2026 & S & Plan & HPC & Ex & Succ & MA, RAG\\
MPI Error Repair~\cite{piersall2026mpi} & 2026 & C & Diag, Code & Off & --- & Acc & RAG, P\\
PETSc Agent Evaluation~\cite{rp260315976} & 2026 & S & Code & HPC & Co, Ex, LJ & Corr & P\\
\bottomrule
\end{tabularx}
\end{table}

\subsection{Operational-Centric Control}
\label{sec:operation}\label{sec:operations}

Operational methods use target-environment descriptions or observations from
running systems. Actions include configurations, diagnoses, interventions, and
resource-allocation policies. Evaluation may use a live service, an isolated
environment, recorded traces, a replay system, or a simulator. Reproducibility
depends on the environment and task: operational evaluations can often be
replayed, but actions that modify production systems can introduce downtime
or other costs. Retrieval, structured observations,
simulation, and constrained execution can help manage these conditions, but
their effectiveness requires task-specific evidence. This group comprises
system integration, production operations, and workload orchestration.

\subsubsection{System Integration (O1)}
\label{sec:int}

\textbf{Task and formulation.}
System integration connects artifacts to execution environments through
dependency installation, deployment configuration, and service interfaces.
A syntactically valid manifest is insufficient if the resulting service cannot
execute its intended workload. Over a task set $\mathcal D$, deployment
success is
\begin{equation}
 \mathrm{DSR}=\frac{1}{|\mathcal D|}\sum_{d\in\mathcal D}
 \ind[\mathrm{Deploy}(d)=\mathrm{Success}],
 \label{eq:deploy-success}
\end{equation}
with success defined by the benchmark's verifier. This measure does not
establish security, sustained availability, or performance under load.

\textbf{Environment construction and artifact deployment.}
DeployBench~\cite{wang2026deploybench} evaluates agents on deploying research
artifacts in fresh virtual machines, verifying each deployment through a hidden
pipeline that executes the paper's designated experiment.
Multi-Docker-Eval~\cite{rp251206915} measures automated environment
construction for repositories across multiple programming languages, in terms
of both executable states and efficiency. SWE-World~\cite{rp260203419} replaces
containerized execution with a learned surrogate that predicts intermediate and
final test outcomes, which lowers the cost of training and evaluating
software-engineering agents but makes the surrogate's fidelity part of the
evaluation. For HPC software, SpackIt~\cite{rp251105626} generates
Spack~\cite{gamblin2015spack} build recipes by combining repository analysis,
retrieved examples, and diagnostic feedback from installation attempts.

\textbf{Kubernetes and network policy configuration.}
GenKubeSec~\cite{malul2024genkubesec} fine-tunes a locally deployable model to
detect, localize, explain, and remediate Kubernetes misconfigurations.
KubeGuard~\cite{rp250904191} derives least-privilege manifests from runtime logs
through modular prompt chaining, and Ueno and Uchiumi~\cite{ueno2024migrating}
benchmark the synthesis of Kubernetes manifests from Compose specifications.
An intent-based framework~\cite{rp260216100} translates natural-language
privacy requirements into Kubernetes placement rules and
software-defined-networking flow rules and validates them on a testbed.

\textbf{Agent--infrastructure interfaces.}
A Model Context Protocol server exposes quantum and HPC resources as tools, so
that an LLM agent can execute quantum workflows submitted in natural language
on a hybrid platform~\cite{rp260408318}. Such interfaces define the action
space available to integration agents rather than evaluating a particular
integration policy.

\textbf{Limitations.}
Integration evaluation benefits from documenting dependency state, runtime
versions, permissions, time limits, and success criteria. Tests of deployment,
security, and service quality address different requirements and should remain
separate when they are not jointly measured. None of the O1 systems is
evaluated on the deployment of a training or serving stack, although such
stacks are among the most dependency-intensive software that infrastructure
teams deploy. Table~\ref{tab:cmp-o1} summarizes these distinctions.

\begin{table}[tp]
\centering
\caption{System integration (O1) systems under the task formulation of Eq.~(\ref{eq:problem}), grouped by method family. Abbreviations follow Table~\ref{tab:cmp-a1} and the preceding tables; new ones: $\mathcal{E}$: VM virtual machine or container, K8s Kubernetes or cloud, Sur learned surrogate; $F$: Li live system.}
\label{tab:cmp-o1}
\scriptsize
\setlength{\tabcolsep}{2.6pt}
\renewcommand{\arraystretch}{1.04}
\begin{tabularx}{\linewidth}{@{}>{\raggedright\arraybackslash}X c>{\raggedright\arraybackslash}p{0.75cm}>{\raggedright\arraybackslash}p{1.05cm}>{\raggedright\arraybackslash}p{1.25cm}>{\raggedright\arraybackslash}p{1.05cm}>{\raggedright\arraybackslash}p{1.25cm}>{\raggedright\arraybackslash}p{1.25cm}@{}}
\toprule
\textbf{System} & \textbf{Year} & $\boldsymbol{\mathcal{I}}$ & $\boldsymbol{\mathcal{A}}$ & $\boldsymbol{\mathcal{E}}$ & $\boldsymbol{F}$ & $\boldsymbol{J}$ & \textbf{Adapt.}\\
\midrule
\rowcolor{tabband}\multicolumn{8}{@{}l}{\textit{Environment construction and artifact deployment}}\\
SpackIt~\cite{rp251105626} & 2025 & C & Cfg & HPC & Co, Ex & Succ & RAG, P\\
Multi-Docker-Eval~\cite{rp251206915} & 2025 & C & Cfg & VM & Ex & Succ, Cost & P, MA\\
DeployBench~\cite{wang2026deploybench} & 2026 & S, C & Plan, Cfg & VM & Ex, Li & Succ & P\\
SWE-World~\cite{rp260203419} & 2026 & C & Code & Sur & LJ & Succ & SFT, RL\\
\addlinespace[1pt]
\rowcolor{tabband}\multicolumn{8}{@{}l}{\textit{Kubernetes and network policy configuration}}\\
GenKubeSec~\cite{malul2024genkubesec} & 2024 & C & Diag, Cfg & Off & --- & Acc & SFT\\
K8s Migration~\cite{ueno2024migrating} & 2024 & C & Cfg & Off & --- & Acc & P\\
KubeGuard~\cite{rp250904191} & 2025 & C, T & Cfg & Off & --- & Acc & P\\
Privacy-Aware Cloud-Edge Orchestration~\cite{rp260216100} & 2026 & S & Cfg & K8s & Ex, Li & Succ & P\\
\addlinespace[1pt]
\rowcolor{tabband}\multicolumn{8}{@{}l}{\textit{Agent--infrastructure interfaces}}\\
Quantum-HPC MCP Server~\cite{rp260408318} & 2026 & S & Plan & HPC & Ex, Li & Succ & P\\
\bottomrule
\end{tabularx}
\end{table}

\subsubsection{Production Operations (O2)}
\label{sec:ops}

\textbf{Task and formulation.}
Production-operations methods interpret logs, metrics, traces, and incident
records, or propose actions such as restarts and configuration changes.
For a labeled incident set $\mathcal D$, root-cause classification accuracy is
\begin{equation}
 \mathrm{Acc}_{\mathrm{RCA}}=\frac{1}{|\mathcal D|}
 \sum_{i\in\mathcal D}\ind[\hat R_i=R_i^*].
 \label{eq:rca}
\end{equation}
Here $i$ indexes an incident, $\hat R_i$ is the predicted root-cause label,
and $R_i^*$ is its reference label. This definition must be distinguished from
language-model ranking scores, which assess relative report quality. Mean time
to recovery is relevant when an intervention is executed and its operational
effect is measured.

\textbf{Log parsing and operational analytics.}
A domain-adapted, instruction-following framework~\cite{karimi2026logs} uses
chain-of-thought reasoning to parse unstructured logs from leadership-class HPC
systems at production scale. EPIC~\cite{rp250916212} answers operational
analytics queries over multimodal data with a hierarchical multi-agent
architecture in which a top-level model orchestrates retrieval, descriptive,
and predictive agents.

\textbf{Root-cause analysis and support.}
MetaKube~\cite{sun2026metakube} accumulates diagnostic experience for
Kubernetes failures in an episodic pattern memory, routes problems between fast
and analytical pathways, and adapts a locally deployable model through
post-training. StateGraph~\cite{xiang2025stategraph} organizes system-state
relationships into a graph that guides root-cause analysis in Kubernetes
clusters, while retrieval-based approaches~\cite{tran2026rca} build root-cause
knowledge bases from support tickets through fine-tuning, retrieval, or a
combination of both. For HPC I/O, IOAgent~\cite{egersdoerfer2026ioagent}
diagnoses performance issues from Darshan traces with a module-based
pre-processor, retrieval-augmented domain knowledge, and a tree-based merger,
and compares its diagnoses with the rule-based Drishti tool~\cite{drishti} and
the LLM-based ION system~\cite{ion}. HPC-LLM~\cite{hpcllm2026} provides
retrieval-augmented, domain-adapted assistance for common HPC workflows such as
job scheduling and MPI execution.

\textbf{Diagnosis for AI training and serving clusters.}
Failures in AI clusters are dominated by hardware and user faults and are
diagnosed mainly from logs~\cite{jiang2025l4}; non-LLM systems such as L4 and
Flare automate parts of this diagnosis for training
jobs~\cite{jiang2025l4,cui2025flare}. ARGUS~\cite{zhou2026argus} provides
always-on, fine-grained tracing and fail-slow detection for training workloads
in production GPU clusters and describes the integration of LLM agents with its
traces as ongoing work; it is included as the observability substrate for such
agents rather than as evidence of LLM-driven diagnosis.
TSGuard~\cite{yang2026tsguard} provides the most direct evidence of LLM-driven
diagnosis: it mines historical on-call records into knowledge bases and
emulates expert diagnosis through structured reasoning and iterative
verification on production AI workload incidents. A cluster diagnosis
agent~\cite{shi2024clusterdiag} combines retrieval, diagram-of-thought
reasoning, and self-play with a benchmark for AI-cluster issues.

\textbf{Autonomous management.}
KubeIntellect~\cite{rp250902449} exposes the full Kubernetes API to a
supervisor and domain-specific agents that can synthesize new tools on demand.
An LLM-orchestrated big-data-as-a-service framework~\cite{rp260617915}
coordinates agents across ingestion, training, deployment, monitoring, and
drift detection. These systems execute actions, which makes their safety
controls part of the evaluated method.

\textbf{Limitations.}
These approaches differ in whether they parse evidence, produce a report, or
execute a remediation, and evaluation should preserve that distinction. An LLM
judge can rank explanations without establishing the correctness of an
intervention, and shared judge and model families introduce potential bias.
Transfer from one cluster, trace collection, or support corpus to another
remains an empirical question. For AI clusters in particular, diagnosis is now
evaluated on production incidents, but mitigation of the diagnosed straggler
or failure is not. Table~\ref{tab:cmp-o2} and Section~\ref{sec:exp-operational}
separate these forms of evidence.

\begin{table}[tp]
\centering
\caption{Production operations (O2) systems under the task formulation of Eq.~(\ref{eq:problem}), grouped by method family. Abbreviations follow Table~\ref{tab:cmp-a1} and the preceding tables; new ones: $\mathcal{E}$: DB database or data lake.}
\label{tab:cmp-o2}
\scriptsize
\setlength{\tabcolsep}{2.6pt}
\renewcommand{\arraystretch}{1.04}
\begin{tabularx}{\linewidth}{@{}>{\raggedright\arraybackslash}X c>{\raggedright\arraybackslash}p{0.75cm}>{\raggedright\arraybackslash}p{1.05cm}>{\raggedright\arraybackslash}p{1.25cm}>{\raggedright\arraybackslash}p{1.05cm}>{\raggedright\arraybackslash}p{1.25cm}>{\raggedright\arraybackslash}p{1.25cm}@{}}
\toprule
\textbf{System} & \textbf{Year} & $\boldsymbol{\mathcal{I}}$ & $\boldsymbol{\mathcal{A}}$ & $\boldsymbol{\mathcal{E}}$ & $\boldsymbol{F}$ & $\boldsymbol{J}$ & \textbf{Adapt.}\\
\midrule
\rowcolor{tabband}\multicolumn{8}{@{}l}{\textit{Log parsing and operational analytics}}\\
EPIC~\cite{rp250916212} & 2025 & T & Diag & HPC & Ex & Acc, Cost & MA, SFT\\
HPC Log Parsing~\cite{karimi2026logs} & 2026 & T & Diag & Off & --- & Acc & SFT\\
\addlinespace[1pt]
\rowcolor{tabband}\multicolumn{8}{@{}l}{\textit{Root-cause analysis and support}}\\
StateGraph~\cite{xiang2025stategraph} & 2025 & T & Diag & Prod & --- & Acc & RAG\\
HPC-LLM~\cite{hpcllm2026} & 2026 & S & Diag & Off & --- & Acc & SFT, RAG\\
IOAgent~\cite{egersdoerfer2026ioagent} & 2026 & T & Diag & Off & --- & Acc & RAG, P\\
LLM-Augmented RCA~\cite{tran2026rca} & 2026 & T & Diag & Off & --- & Acc & SFT, RAG\\
MetaKube~\cite{sun2026metakube} & 2026 & T & Diag & Off & --- & Acc & RAG, SFT\\
\addlinespace[1pt]
\rowcolor{tabband}\multicolumn{8}{@{}l}{\textit{Diagnosis for AI training and serving clusters}}\\
Cluster Diagnosis Agent$^{\star}$~\cite{shi2024clusterdiag} & 2024 & T & Diag, Cfg & HPC & Li & Acc & RAG, P\\
TSGuard$^{\star}$~\cite{yang2026tsguard} & 2025 & T & Diag & Prod & Li & Acc & MA, RAG\\
\addlinespace[1pt]
\rowcolor{tabband}\multicolumn{8}{@{}l}{\textit{Autonomous management}}\\
KubeIntellect~\cite{rp250902449} & 2025 & S & Plan, Code & K8s & Li, Hu & Succ & MA\\
LLM-Orchestrated Big-Data Service~\cite{rp260617915} & 2026 & S & Plan & DB & Ex, Hu & Succ & MA\\
\bottomrule
\end{tabularx}
\end{table}

\subsubsection{Workload Orchestration (O3)}
\label{sec:orch}

\textbf{Task and formulation.}
Workload orchestration decides how training and serving workloads occupy a
cluster: how a model is partitioned across devices, which requests are batched
and routed where, which cached state is evicted, where experts and models are
placed, and which job runs on which machine. The action is a plan or a policy
$\pi$, often an executable program, and the environment $\mathcal E$ is a
simulator, a projection model, a trace, or a live cluster. Two objectives
recur. For serving, SLO-constrained goodput counts the requests that meet both
a time-to-first-token (TTFT) and a time-per-output-token (TPOT) target over a
horizon $T$,
\begin{equation}
 \mathrm{Goodput}(\pi)=\frac{1}{T}\sum_{r\in\mathcal R}
 \ind\!\left[\mathrm{TTFT}_r(\pi)\le\tau_{\mathrm{TTFT}}\land
 \mathrm{TPOT}_r(\pi)\le\tau_{\mathrm{TPOT}}\right],
 \label{eq:goodput}
\end{equation}
where $\mathcal R$ is the set of requests in the trace and
$\tau_{\mathrm{TTFT}}$ and $\tau_{\mathrm{TPOT}}$ are the latency targets. For
training and cluster scheduling, the mean job completion time (JCT) is
$\overline{\mathrm{JCT}}(\pi)=|\mathcal J|^{-1}\sum_{j\in\mathcal J}(C_j(\pi)-s_j)$,
where $\mathcal J$ is the job set and $C_j$ and $s_j$ are the completion and
submission times of job $j$. A training configuration is instead scored by
throughput per device or model FLOPs utilization at a fixed global batch. In
all cases the policy is evaluated on a workload distribution, so the fidelity
of the simulator or projection and the representativeness of the trace limit
the conclusions that a result supports.

\textbf{Parallelism and sharding configuration.}
LLM agents have been applied to the configuration of distributed training.
ASAP~\cite{ding2025asap} combines coordinator, analyzer, and proposal agents
with profiling data, roofline analysis, and a knowledge base of past
optimizations to diagnose whether a TPU training job is compute-,
memory-bandwidth-, or communication-bound and to propose a sharding mesh. The
Primus Tuning Agent~\cite{raina2026primus} searches tensor, pipeline, expert,
context, and data parallelism together with micro-batch size, recomputation,
and kernel backends, using a single-node projection engine as a fast scoring
oracle and an analytic memory filter that rejects infeasible candidates before
any GPU time is spent. OptiCo~\cite{chen2026optico} coordinates reasoning
agents through a shared message pool and a failure-analysis agent to tune
Megatron-LM~\cite{shoeybi2019megatron} parallelization strategies.

\textbf{Serving and memory policies.}
A second family synthesizes the runtime policies of inference serving as
programs. Glia~\cite{hamadanian2025glia} organizes LLM agents into a
researcher--supervisor workflow that forms hypotheses, designs simulator
experiments, and analyzes telemetry, and applies it to request routing,
scheduling, and autoscaling in a GPU cluster that serves LLMs.
Autopoiesis~\cite{jiang2026autopoiesis} performs policy synthesis online: an
LLM-driven evolution loop rewrites the scheduling and rescheduling policies of a
serving system as workload and cluster conditions shift, and is evaluated
against DistServe~\cite{zhong2024distserve}, HexGen~\cite{jiang2024hexgen}, and
SpotServe~\cite{miao2024spotserve} on homogeneous, heterogeneous, and elastic
clusters. CacheCraft~\cite{poudel2026cachecraft} evolves KV-cache eviction
policies under a strict policy interface and a cascade evaluator, using
long-context benchmarks such as RULER~\cite{hsieh2024ruler} as its fitness
signal. The ADRS case
studies~\cite{cheng2025adrs} apply the open-source
OpenEvolve~\cite{sharma2025openevolve} framework to systems problems with
simulator-based verifiers, including expert-parallel load balancing (EPLB) for
MoE inference, model-to-GPU placement, and spot-instance scheduling, and a
follow-up study~\cite{cheng2025letbarbarians} extends the analysis to further
case studies and evolution frameworks.

\textbf{Cluster scheduling and placement.}
SchedMate~\cite{wang2025semantic} augments existing deep-learning schedulers
with three LLM components that read source code, runtime logs, and historical
jobs to estimate job durations, cancel poor packing decisions, and diagnose
failures. AlphaEvolve~\cite{novikov2025alphaevolve} evolves the
machine-priority heuristic of Google's Borg scheduler in a fleet simulator built
from historical workload snapshots and deploys the result fleet-wide; the same
framework also evolved a matrix-multiplication tiling heuristic used to train
Gemini and a FlashAttention~\cite{dao2022flashattention} configuration at the
compiler-IR level. An agentic scheduler~\cite{lu2026agenticsched} for
heterogeneous AI tool workloads pairs an LLM with an algorithmic runtime
monitor to assign tools to CPU or GPU execution under a memory budget. The HPC
studies in this family are smaller in scale: Incisor~\cite{laurenzano2026incisor}
combines program analysis with LLM reasoning to select cloud instances for HPC
jobs from submission artifacts alone, a comparative study~\cite{sharma2025workload}
evaluates many LLMs on a heterogeneous workload-mapping problem described in
natural language, and a ReAct-style~\cite{yao2023react}
scheduler~\cite{rp250602025} balances multiple objectives with a scratchpad
memory and a constraint-enforcement module, in comparison with
first-come-first-served (FCFS) and shortest-job-first (SJF) ordering and an
OR-Tools~\cite{perron2024ortools} solver.

\textbf{Context: optimizers without an LLM agent.}
Several serving systems search configuration spaces that overlap with O3
without using an LLM as the optimizer. AIConfigurator searches serving
configurations with performance models and a calibrated
database~\cite{xu2026aiconfigurator}; SLO-Guard autotunes eight vLLM~\cite{kwon2023vllm} knobs,
including quantization, batching limits, and prefix caching, with annealing
and tree-structured Parzen estimation~\cite{lysenstoen2026sloguard}; and
Scepsy searches GPU shares, tensor-parallel degrees, and replica counts for
multi-LLM agentic workflows with a throughput--latency
predictor~\cite{white2026scepsy}. They provide baselines and evaluators for
future LLM-driven methods but are not counted as evidence of LLM-driven
engineering. LLM agents have not yet been applied to checkpoint
frequency and placement, straggler mitigation, or the joint choice of
quantization and parallelism for a deployment.

\textbf{Limitations.}
Evidence in O3 is recent and relies largely on simulation. Glia~\cite{hamadanian2025glia}, ADRS~\cite{cheng2025adrs}, CacheCraft~\cite{poudel2026cachecraft}, and the
AlphaEvolve~\cite{novikov2025alphaevolve} scheduling study score candidates in simulators or evaluation
harnesses, and Primus~\cite{raina2026primus} scores them with a projection engine whose
approximation error must be accounted for; only some of them report real-cluster validation, and
only AlphaEvolve reports production deployment. Evolved policies can overfit
the evaluator, and several studies report reward hacking or plateaus that
required changing the policy interface. The baselines also differ in
strength, from least-loaded routing to published state-of-the-art systems and
expert-tuned configurations. Search cost is reported inconsistently: in hours
and dollars by ADRS, in simulations by Glia, and in minutes of single-node exploration by Primus.
Table~\ref{tab:cmp-o3} and Section~\ref{sec:exp-orch} compare these results
under their native conditions.

\begin{table}[tp]
\centering
\caption{Workload orchestration (O3) systems under the task formulation of Eq.~(\ref{eq:problem}), grouped by method family. Abbreviations follow Table~\ref{tab:cmp-a1} and the preceding tables; new ones: $\mathcal{A}$: Policy runtime policy; $\mathcal{E}$: Sim system simulator; $F$: Re trace replay.}
\label{tab:cmp-o3}
\scriptsize
\setlength{\tabcolsep}{2.6pt}
\renewcommand{\arraystretch}{1.04}
\begin{tabularx}{\linewidth}{@{}>{\raggedright\arraybackslash}X c>{\raggedright\arraybackslash}p{0.75cm}>{\raggedright\arraybackslash}p{1.05cm}>{\raggedright\arraybackslash}p{1.25cm}>{\raggedright\arraybackslash}p{1.05cm}>{\raggedright\arraybackslash}p{1.25cm}>{\raggedright\arraybackslash}p{1.25cm}@{}}
\toprule
\textbf{System} & \textbf{Year} & $\boldsymbol{\mathcal{I}}$ & $\boldsymbol{\mathcal{A}}$ & $\boldsymbol{\mathcal{E}}$ & $\boldsymbol{F}$ & $\boldsymbol{J}$ & \textbf{Adapt.}\\
\midrule
\rowcolor{tabband}\multicolumn{8}{@{}l}{\textit{Parallelism and sharding configuration}}\\
ASAP$^{\star}$~\cite{ding2025asap} & 2025 & T & Cfg & NPU & --- & Perf & MA, RAG\\
OptiCo$^{\star}$~\cite{chen2026optico} & 2026 & K, T & Cfg & GPU & Ex & Perf & MA\\
Primus Tuning Agent$^{\star}$~\cite{raina2026primus} & 2026 & K, T & Cfg & GPU, Sim & Ex, Si & Perf & MA\\
\addlinespace[1pt]
\rowcolor{tabband}\multicolumn{8}{@{}l}{\textit{Serving and memory policies}}\\
ADRS$^{\star}$~\cite{cheng2025adrs} & 2025 & C & Policy & Sim & Si & Perf, Cost & Evo\\
Glia$^{\star}$~\cite{hamadanian2025glia} & 2025 & S, T & Policy & Sim, GPU & Si, Ex & Perf & MA\\
Let the Barbarians In$^{\star}$~\cite{cheng2025letbarbarians} & 2025 & C & Policy & Sim & Si & Perf & Evo\\
Autopoiesis$^{\star}$~\cite{jiang2026autopoiesis} & 2026 & T & Policy & Sim, GPU & Si, Li & Perf & Evo\\
CacheCraft$^{\star}$~\cite{poudel2026cachecraft} & 2026 & C & Policy & GPU & Ex & Acc & Evo\\
\addlinespace[1pt]
\rowcolor{tabband}\multicolumn{8}{@{}l}{\textit{Cluster scheduling and placement}}\\
AlphaEvolve$^{\star}$~\cite{novikov2025alphaevolve} & 2025 & C & Policy & Sim, Prod & Si, Li & Perf & Evo\\
Multiobjective HPC Scheduling~\cite{rp250602025} & 2025 & S, K, T & Plan & Sim & Si & Perf & P\\
SchedMate$^{\star}$~\cite{wang2025semantic} & 2025 & C, T & Plan & HPC, Sim & Re, Li & Perf & RAG, P\\
Workload Mapping and Scheduling~\cite{sharma2025workload} & 2025 & S, K & Plan & Off & --- & Perf & P\\
Agentic CPU-GPU Scheduling$^{\star}$~\cite{lu2026agenticsched} & 2026 & T & Plan & GPU, CPU & Li & Perf & P\\
Incisor~\cite{laurenzano2026incisor} & 2026 & C & Cfg & K8s & St & Perf, Cost & P\\
\bottomrule
\end{tabularx}
\end{table}

\subsection{Resource-Centric Optimization}
\label{sec:resource}

Resource-oriented methods examine data-processing quality and cost, or energy
and power requirements. Their actions may still be code or configuration edits,
but the main evaluation concerns the data service or resource consumed.
Measurement boundaries require care: token counts, device energy, facility energy,
and carbon estimates represent different quantities. Quality and resource use
should be reported jointly when the method trades one against the other. This
group comprises data management and energy optimization.

\subsubsection{Data Management (R1)}
\label{sec:data}

\textbf{Task and formulation.}
Data-management applications synthesize pipelines, extraction functions,
queries, and access plans. Output quality may be assessed using reference
records, answer labels, or data constraints. Resource measures include bytes,
tokens, tool calls, storage footprint, and execution time. The task trades
quality against resource use,
\begin{equation}
 \max_{a\in\mathcal{A}}\;Q(a)
 \quad\text{subject to}\quad C(a)\le B,
 \label{eq:data-quality-cost}
\end{equation}
where $Q(a)$ is output quality against the reference, $C(a)$ is one of these
resource measures, and $B$ is a budget; studies that report only quality
correspond to an unconstrained budget. Text-to-SQL work supplies related methods for
natural-language data access~\cite{shi2025text2sql,hong2025text2sql}.

\textbf{Pipeline, query, and contract generation.}
Pipeline generation translates transformation specifications into executable
code~\cite{chen2025datapipeline,dataflow2025prep}. Among R1 systems, DataFlow is
most directly related to AI infrastructure: it prepares training data for LLMs
with nearly 200 reusable operators and an agent that turns natural-language
specifications into pipelines, and evaluates the pipelines by downstream model
quality~\cite{dataflow2025prep}. StreamLink generates Spark and Hadoop queries
with locally fine-tuned models and LLM-based syntax and security
checks~\cite{rp250521575}, and fine-tuned models generate data contracts that
formalize schema and quality expectations between producers and
consumers~\cite{bhoite2025contracts}.

\textbf{Discovery and catalogs.}
Data-discovery and catalog-oriented work supports identifying and describing
relevant sources: LEDD builds hierarchical semantic catalogs and table search
for data lakes~\cite{an2025ledd}, and LAKEGEN uses LLMs to generate
domain-specific tabular corpora with annotated joinability for evaluating
discovery methods~\cite{lakegen2025tabular}.

\textbf{Extraction and analysis agents.}
Evaporate compares direct extraction with synthesized extraction functions,
reporting both extraction quality and LLM resource use~\cite{arora2023evaporate}.
SANA examines planning, retrieval, and analysis bottlenecks using controlled
and idealized tool configurations~\cite{sana2026datalake}; its idealized
conditions estimate diagnostic headroom rather than deployable performance.
NotebookRAG retrieves and re-executes cells from existing analysis notebooks
to ground automated exploratory analysis~\cite{rp260217215}.

\textbf{Data-pipeline benchmarks.}
KramaBench~\cite{rp250606541} contains curated data-lake-to-insight challenges
that require the end-to-end orchestration of extraction, cleaning,
integration, and analysis. DW-Bench~\cite{rp260418964} poses graph-topology
questions over data-warehouse schemas that combine foreign-key and lineage
edges and compares tool-augmented with static methods.

\textbf{Limitations.}
Resource claims should specify the quantities being counted. Fewer retrieval calls need not
reduce total model computation when other tool calls increase, and token
reductions exclude some costs of generating, validating, and executing
extraction code. Except for DataFlow~\cite{dataflow2025prep}, these studies evaluate data services
that can support AI workflows rather than measuring training-data throughput
directly; the data loaders and input pipelines that keep accelerators
supplied during training have not yet been targeted by LLM-driven methods. Evaluation on a
fixed corpus with explicit quality measures and end-to-end resource
accounting would support stronger comparisons (Table~\ref{tab:cmp-r1} and
Section~\ref{sec:exp-resources}).

\begin{table}[tp]
\centering
\caption{Data management (R1) systems under the task formulation of Eq.~(\ref{eq:problem}), grouped by method family. Abbreviations follow Table~\ref{tab:cmp-a1} and the preceding tables; new ones: $\mathcal{A}$: Data data or catalog.}
\label{tab:cmp-r1}
\scriptsize
\setlength{\tabcolsep}{2.6pt}
\renewcommand{\arraystretch}{1.04}
\begin{tabularx}{\linewidth}{@{}>{\raggedright\arraybackslash}X c>{\raggedright\arraybackslash}p{0.75cm}>{\raggedright\arraybackslash}p{1.05cm}>{\raggedright\arraybackslash}p{1.25cm}>{\raggedright\arraybackslash}p{1.05cm}>{\raggedright\arraybackslash}p{1.25cm}>{\raggedright\arraybackslash}p{1.25cm}@{}}
\toprule
\textbf{System} & \textbf{Year} & $\boldsymbol{\mathcal{I}}$ & $\boldsymbol{\mathcal{A}}$ & $\boldsymbol{\mathcal{E}}$ & $\boldsymbol{F}$ & $\boldsymbol{J}$ & \textbf{Adapt.}\\
\midrule
\rowcolor{tabband}\multicolumn{8}{@{}l}{\textit{Pipeline, query, and contract generation}}\\
Data-Pipeline Interface~\cite{chen2025datapipeline} & 2024 & S & Plan & --- & --- & --- & P\\
AI Data Contracts~\cite{bhoite2025contracts} & 2025 & C & Cfg & DB & St & Acc & SFT\\
DataFlow$^{\star}$~\cite{dataflow2025prep} & 2025 & S & Code & DB & Ex & Acc & MA\\
StreamLink~\cite{rp250521575} & 2025 & S & Code & DB & St & Corr & SFT\\
\addlinespace[1pt]
\rowcolor{tabband}\multicolumn{8}{@{}l}{\textit{Discovery and catalogs}}\\
LAKEGEN~\cite{lakegen2025tabular} & 2025 & S & Data & DB & --- & Acc & P\\
LEDD~\cite{an2025ledd} & 2025 & S & Data & DB & --- & --- & P\\
\addlinespace[1pt]
\rowcolor{tabband}\multicolumn{8}{@{}l}{\textit{Extraction and analysis agents}}\\
Evaporate~\cite{arora2023evaporate} & 2023 & C & Code & Off & Ex & Acc, Cost & P\\
NotebookRAG~\cite{rp260217215} & 2026 & S & Code & DB & Ex, Hu & Acc & RAG, MA\\
SANA~\cite{sana2026datalake} & 2026 & S & Plan, Code & DB & Ex & Acc & MA\\
\addlinespace[1pt]
\rowcolor{tabband}\multicolumn{8}{@{}l}{\textit{Data-pipeline benchmarks}}\\
KramaBench~\cite{rp250606541} & 2025 & S & Plan, Code & DB & Ex & Acc & P, MA\\
DW-Bench~\cite{rp260418964} & 2026 & S & Diag & DB & Ex & Acc & P\\
\bottomrule
\end{tabularx}
\end{table}

\subsubsection{Energy Optimization (R2)}
\label{sec:energy}

\textbf{Task and formulation.}
Energy-oriented applications generate code, estimate facility demand, or
propose power-management actions. Relevant outcomes include joules, average
power, runtime, and forecast error. An energy-delay
product~\cite{gonzalez1996energy} can summarize one trade-off,
\begin{equation}
 \mathrm{EDP}=E\,T=P_{\mathrm{avg}}T^2,
 \label{eq:edp}
\end{equation}
where $E$ is measured energy, $T$ is elapsed execution time, and
$P_{\mathrm{avg}}=E/T$ is average power over the same measurement interval.
This optional summary is distinct from expected energy reduction and does not
replace application-specific performance constraints.

\textbf{Energy-aware code generation.}
Prompting and measurement studies examine the energy consumption of generated
code on multiple platforms~\cite{cappendijk2024energyprompt,peng2024energycode,apsan2025energy},
including comparisons with code written by developers and green-software
experts~\cite{apsan2025energy}; their conclusions depend on device, workload,
and measurement harness. LASSI-EE~\cite{dearing2026lassiee} refactors parallel
scientific code through a multi-stage loop that combines runtime power
profiling, energy-aware prompting, self-correction, and an LLM judge.

\textbf{Power-oriented hardware optimization.}
Design-oriented systems connect energy objectives to artifact optimization.
POET~\cite{ping2026poet} evolves RTL with power-first Pareto ranking and uses
differential testing against the original design as a functional oracle.
AUTOGATE~\cite{rp260617461} distills long toggle traces into clustered
representations that guide LLM-based clock-gating rewrites across hierarchical
designs, including the NVDLA deep-learning accelerator. A co-evolutionary
method~\cite{rp260613089} evolves 8-bit approximate multipliers together with
the prompts that steer the LLM. Their category follows the main objective
rather than the language of the generated artifact.

\textbf{Facility forecasting.}
HyperLoad uses an LLM-based architecture to forecast data-center cooling
load~\cite{jiang2025hyperload}; lower prediction error does not itself
demonstrate reduced facility energy.

\textbf{Limitations.}
A rigorous energy evaluation reports the device, meter, workload, repetition
scheme, measurement boundary, and reference implementation. Search and model
inference costs should be included when assessing net savings, and carbon
estimates additionally depend on the carbon-intensity trace and the time of
execution. LLMs
have not yet been applied to power capping or energy-aware scheduling of
training and serving jobs, even though these workloads dominate the power
budget of AI clusters. The available studies motivate multi-objective
evaluation while leaving substantial uncertainty about transfer across
platforms and scales (Table~\ref{tab:cmp-r2}).

\begin{table}[tp]
\centering
\caption{Energy optimization (R2) systems under the task formulation of Eq.~(\ref{eq:problem}), grouped by method family. Abbreviations follow Table~\ref{tab:cmp-a1} and the preceding tables; new ones: $\mathcal{A}$: Pred forecast; $J$: Energy energy or power.}
\label{tab:cmp-r2}
\scriptsize
\setlength{\tabcolsep}{2.6pt}
\renewcommand{\arraystretch}{1.04}
\begin{tabularx}{\linewidth}{@{}>{\raggedright\arraybackslash}X c>{\raggedright\arraybackslash}p{0.75cm}>{\raggedright\arraybackslash}p{1.05cm}>{\raggedright\arraybackslash}p{1.25cm}>{\raggedright\arraybackslash}p{1.05cm}>{\raggedright\arraybackslash}p{1.25cm}>{\raggedright\arraybackslash}p{1.25cm}@{}}
\toprule
\textbf{System} & \textbf{Year} & $\boldsymbol{\mathcal{I}}$ & $\boldsymbol{\mathcal{A}}$ & $\boldsymbol{\mathcal{E}}$ & $\boldsymbol{F}$ & $\boldsymbol{J}$ & \textbf{Adapt.}\\
\midrule
\rowcolor{tabband}\multicolumn{8}{@{}l}{\textit{Energy-aware code generation}}\\
Energy-Efficient Code Directions~\cite{peng2024energycode} & 2024 & C & Code & CPU & Ex & Energy & P\\
Energy-Efficient Code Prompting~\cite{cappendijk2024energyprompt} & 2024 & S & Code & CPU & --- & Energy & P\\
Energy-Efficient Code Study~\cite{apsan2025energy} & 2025 & S & Code & CPU & --- & Energy & P\\
LASSI-EE~\cite{dearing2026lassiee} & 2025 & C & Code & GPU & Ex, LJ & Energy & P\\
\addlinespace[1pt]
\rowcolor{tabband}\multicolumn{8}{@{}l}{\textit{Power-oriented hardware optimization}}\\
AUTOGATE$^{\star}$~\cite{rp260617461} & 2026 & C, T & Code & RTL, EDA & Si, Co & PPA & MA\\
Coevolutionary Circuit Approximation~\cite{rp260613089} & 2026 & S & Code & EDA & Si & PPA, Acc & Evo\\
POET~\cite{ping2026poet} & 2026 & C & Code & RTL, EDA & Si, Co & PPA & Evo\\
\addlinespace[1pt]
\rowcolor{tabband}\multicolumn{8}{@{}l}{\textit{Facility forecasting}}\\
HyperLoad~\cite{jiang2025hyperload} & 2025 & T & Pred & Off & --- & Acc & SFT\\
\bottomrule
\end{tabularx}
\end{table}

\section{Benchmarks and Quantitative Synthesis}
\label{sec:experiments}

\subsection{Benchmarks, Metrics, and Evidence Extraction}
\label{sec:exp-setup}

This section synthesizes the evaluations reported by the cited studies; it does
not re-execute their systems. Numerical values, metric definitions, task sets,
and experimental conditions were checked against the primary publications.
Values are transcribed from tables, text, or printed figure labels, and derived
quantities are computed from the reported values or underlying counts and
identified as such; no value is read from an unlabeled graphical coordinate.
Appendix~\ref{app:ledgers} lists the source locations, and
Appendix~\ref{app:tiers} describes the evaluation protocols.

Hardware studies commonly use VerilogEval~\cite{liu2023verilogeval} and RTLLM~\cite{lu2023rtllm}, whereas kernel studies use
KernelBench~\cite{kernelbench2025}, TritonBench~\cite{tritonbench2025}, and MultiKernelBench~\cite{wen2025multikernelbench}. HPC-software evaluations range
from compiler test collections to parallel translation and defect-analysis
corpora. DeployBench~\cite{wang2026deploybench} and TraceBench~\cite{egersdoerfer2026ioagent} evaluate deployment and I/O diagnosis,
respectively. Workload-orchestration studies use serving simulators such as
vidur~\cite{agrawal2024vidur}, projection engines calibrated by single-node benchmarks, production
cluster traces, and real training and serving clusters. Data and energy studies
use task-specific corpora and measurement harnesses. A shared benchmark name
does not imply identical revisions, task subsets, or execution conditions.

Each result retains its metric, baseline, model, hardware, and budget where
these are reported. Within-study comparisons provide the more controlled
contrast, although simultaneous changes in several components still limit
causal attribution. Cross-study comparisons characterize reported outcomes under
their respective conditions; they do not estimate the effect of transferring a
method to another study's environment. The interpretation rules and the
category-specific protocol differences are given in Appendix~\ref{app:tiers}.

\subsection{Artifact-Centric Evaluation (A1--A3)}
\label{sec:exp-artifact}\label{sec:exp-hardware}
\label{sec:exp-kernels}\label{sec:exp-hpc}

\begin{figure}[t]
\centering
\begin{subfigure}[t]{0.32502\textwidth}
\centering
\includegraphics[width=\linewidth]{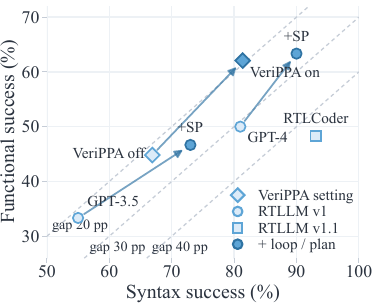}
\caption{Syntax and functional success on RTLLM~\cite{lu2023rtllm}.}
\label{fig:a1-syntax}
\end{subfigure}\hfill
\begin{subfigure}[t]{0.31388\textwidth}
\centering
\includegraphics[width=\linewidth]{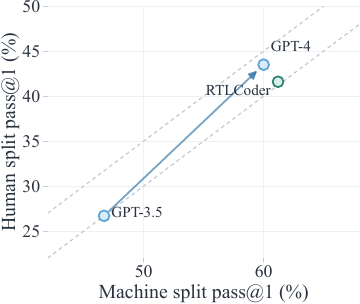}
\caption{Machine and Human splits of VerilogEval~\cite{liu2023verilogeval}.}
\label{fig:a1-splits}
\end{subfigure}\hfill
\begin{subfigure}[t]{0.32110\textwidth}
\centering
\includegraphics[width=\linewidth]{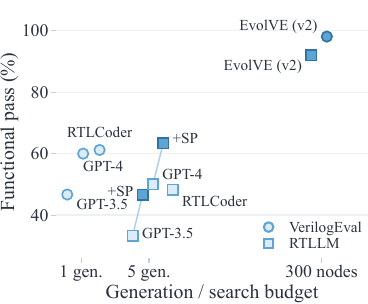}
\caption{Functional success and generation/search budget.}
\label{fig:a1-budget}
\end{subfigure}
\caption{RTL-generation performance across syntax, functionality, evaluation splits, and search
budgets. Results follow the protocols of RTLLM, VerilogEval, RTLCoder, VeriPPA, and
EvolVE~\cite{lu2023rtllm,liu2023verilogeval,liu2023rtlcoder,thorat2025verippa,hsin2026evolve}.
Values are success rates in percent; higher is better. Arrows denote within-study
changes, and dashed lines denote fixed inter-metric gaps. Panel~(c) uses a logarithmic budget axis. Protocol differences are detailed in Appendix~\ref{app:tiers}.}
\Description{Three scatter plots compare syntax with functionality, Machine with Human pass rates, and functional success with budget.}
\label{fig:a1-plane}
\end{figure}

\noindent\textbf{Hardware design (A1).}
The A1 results distinguish two requirements of RTL automation: generating
plausible candidates and validating their intended behavior.
In Figure~\ref{fig:a1-plane}, self-planning~\cite{lu2023rtllm} and VeriPPA~\cite{thorat2025verippa} improve functional
outcomes within their respective studies, while RTLCoder's high syntax rate
coexists with substantially lower functional success~\cite{liu2023rtlcoder}. Language
specialization and syntactic validity therefore provide an incomplete basis for
selecting a hardware-generation method, and functional feedback should remain an
explicit optimization target rather than a final compilation
check~\cite{lu2023rtllm,liu2023rtlcoder,thorat2025verippa}. Syntax and
functionality use different denominators in these sources
(Appendix~\ref{app:tiers}), so their plotted separation is not a common failure
probability.

Table~\ref{tab:a1-crosspaper} further shows that domain adaptation improves
both the Machine and Human VerilogEval~\cite{liu2023verilogeval} results but does not close the
gap between the two specification sources. Models should therefore be evaluated
on human-authored requirements in addition to automatically generated
specifications, since effectiveness on one form of task description does not
establish effectiveness on the other.

Method selection also depends on the verification budget. BetterV~\cite{rp240203375}
achieves the highest Machine pass@1 in Table~\ref{tab:a1-crosspaper}, whereas
RTLCoder-DeepSeek~\cite{liu2023rtlcoder} achieves the highest pass@10. This
reversal separates first-attempt success from success obtained through
additional candidates, so a single attempt budget cannot characterize both use
cases. EvolVE extends the design space with structured search and reports
higher success on the v2 revisions of both benchmarks~\cite{hsin2026evolve}; the
revision change, however, prevents attributing the cross-study difference to
search alone. The evidence therefore favors evaluating the model, search
strategy, and validator jointly under the intended engineering budget.

\begin{table}[t]
\centering
\caption{RTL-generation systems under the RTLCoder evaluation protocol~\cite{liu2023rtlcoder}.
Values are transcribed from Table~II of that study; for starred systems, the VerilogEval
values are quoted from their original publications. Pass@$k$, syntax success, and functional
success are percentages; higher is better. M$-$H is the Machine-minus-Human pass@1 gap, and the
$\Delta$ row gives the change in percentage points after RTL-specific fine-tuning. Bold marks
column maxima, excluding M$-$H; a dash denotes an unreported value.}
\label{tab:a1-crosspaper}
\footnotesize
\setlength{\tabcolsep}{4.6pt}
\renewcommand{\arraystretch}{1.06}
\begin{fittable}\begin{tabular}{@{}llrrrrrrcrr@{}}
\toprule
& & \multicolumn{3}{c}{\textbf{VerilogEval-Machine}}
  & \multicolumn{3}{c}{\textbf{VerilogEval-Human}}
  & & \multicolumn{2}{c}{\textbf{RTLLM v1.1 (pass@5)}}\\
\cmidrule(lr){3-5}\cmidrule(lr){6-8}\cmidrule(lr){10-11}
\textbf{System} & \textbf{Params}
& $k{=}1$ & $k{=}5$ & $k{=}10$ & $k{=}1$ & $k{=}5$ & $k{=}10$
& \textbf{M$-$H} & \textbf{Syntax} & \textbf{Func.}\\
\midrule
\rowcolor{tabband}\multicolumn{11}{@{}l}{\textit{Closed-source general models}}\\
GPT-3.5$^{\star}$ & --- & 46.7 & 69.1 & 74.1 & 26.7 & 45.8 & 51.7 & 20.0 & 89.7 & 37.9\\
GPT-4$^{\star}$ & --- & 60.0 & 70.6 & 73.5 & \textbf{43.5} & \textbf{55.8} & \textbf{58.9}
      & 16.5 & \textbf{100.0} & \textbf{65.5}\\
\midrule
\rowcolor{tabband}\multicolumn{11}{@{}l}{\textit{Models adapted to hardware description languages}}\\
ChipNeMo$^{\star}$~\cite{chipnemo2023} & 13B & 43.4 & \na & \na & 22.4 & \na & \na & 21.0 & \na & \na\\
VerilogEval-SFT$^{\star}$~\cite{liu2023verilogeval} & 16B & 46.2 & 67.3 & 73.7 & 28.8 & 45.9 & 52.3 & 17.4 & \na & \na\\
BetterV$^{\star}$~\cite{rp240203375} & 7B & \textbf{64.2} & 75.4 & 79.1 & 40.9 & 50.0 & 53.3 & 23.3 & \na & \na\\
Thakur et al.~\cite{rp221211140} & 16B & 44.0 & 52.6 & 59.2 & 30.3 & 43.9 & 49.6 & 13.7 & 86.2 & 24.1\\
\midrule
\rowcolor{tabband}\multicolumn{11}{@{}l}{\textit{General open code models}}\\
CodeGen2 & 16B & 5.0 & 9.0 & 13.9 & 0.9 & 4.1 & 7.3 & 4.1 & 72.4 & 6.9\\
StarCoder & 15B & 46.8 & 54.5 & 59.6 & 18.1 & 26.1 & 30.4 & 28.7 & 93.1 & 27.6\\
Mistral-7B-v0.1 & 7B & 36.9 & 48.8 & 57.4 & 4.5 & 12.6 & 18.6 & 32.4 & 72.4 & 20.7\\
DeepSeek-Coder-6.7b & 6.7B & 54.1 & 63.8 & 67.5 & 30.2 & 42.2 & 46.2 & 23.9 & 89.6 & 34.5\\
\midrule
\rowcolor{tabband}\multicolumn{11}{@{}l}{\textit{Open code models after RTL-specific fine-tuning}}\\
RTLCoder-Mistral & 7B & 62.5 & 72.2 & 76.6 & 36.7 & 45.5 & 49.2 & 25.8 & 96.6 & 48.3\\
RTLCoder-DeepSeek & 6.7B & 61.2 & \textbf{76.5} & \textbf{81.8} & 41.6 & 50.1 & 53.4 & 19.6 & 93.1 & 48.3\\
\midrule
\rowcolor{tabband}
\multicolumn{2}{@{}l}{$\Delta$ fine-tuning (DeepSeek-Coder $\rightarrow$ RTLCoder)}
& \gain{7.1} & \gain{12.7} & \gain{14.3} & \gain{11.4} & \gain{7.9} & \gain{7.2}
& $-4.3$ & \gain{3.5} & \gain{13.8}\\
\bottomrule
\end{tabular}\end{fittable}
\end{table}

\begin{figure}[t]
\centering
\begin{subfigure}[t]{0.29307\textwidth}
\centering
\includegraphics[width=\linewidth]{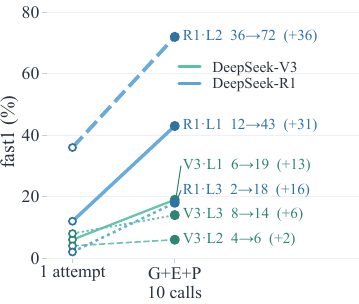}
\caption{$\mathrm{fast}_1$ before and after feedback refinement.}
\label{fig:a2-refinement}
\end{subfigure}\hfill
\begin{subfigure}[t]{0.29981\textwidth}
\centering
\includegraphics[width=\linewidth]{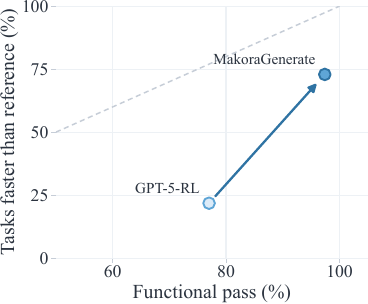}
\caption{Correctness and speed under a shared reference.}
\label{fig:a2-correctness}
\end{subfigure}\hfill
\begin{subfigure}[t]{0.36712\textwidth}
\centering
\includegraphics[width=\linewidth]{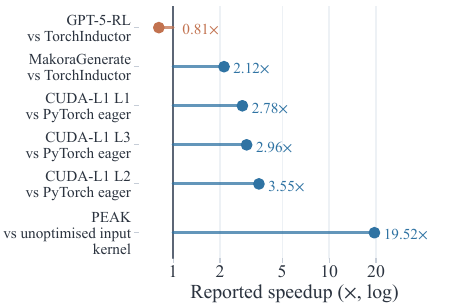}
\caption{Speedup against each study's reference.}
\label{fig:a2-speedup}
\end{subfigure}
\caption{Kernel-generation performance under feedback refinement and study-specific execution
references. Panel~(a) uses KernelBench on L40S~\cite{kernelbench2025}; panel~(b) uses
the TorchInductor~\cite{ansel2024pytorch2} protocol~\cite{tehrani2026gpt5}; panel~(c) includes different
baselines and aggregation rules~\cite{tehrani2026gpt5,li2026cudal1,tariq2025peak}.
$\mathrm{fast}_1$ requires both correctness and faster execution. Both coordinates in~(b) are task
percentages; speedups in~(c) are factors relative to the named baseline. Higher is
better, and a speedup below $1\times$ denotes a slowdown.}
\Description{A slopegraph shows six refinement gains; a scatter plot compares GPT-5-RL and MakoraGenerate; a logarithmic dot plot shows reported speedup factors.}
\label{fig:a2-loop}
\end{figure}

\noindent\textbf{Kernel engineering (A2).}
The kernel results indicate that the value of feedback depends on how the model
interacts with the evaluation environment. In Figure~\ref{fig:a2-loop}(a), the
same refinement budget yields substantially different gains across models and
task levels. Moreover, the fixed-budget ablation in
Table~\ref{tab:a2-fast1matrix} shows that adding profiling information improves
some configurations and degrades others~\cite{kernelbench2025}, so additional
context is not an unconditional improvement. Feedback selection and
presentation should be evaluated jointly with the model; adaptive selection
between execution diagnostics and profiling is a research direction suggested by
these results rather than an established explanation of the differences.

The results further indicate that reliable generation and competitive
optimization require separate evaluation targets. GPT-5-RL exhibits a large gap
between functional success and performance success, whereas MakoraGenerate
improves both and narrows the gap~\cite{tehrani2026gpt5}. Evaluation should
therefore assess whether an agent uses its search budget to discover faster
valid implementations, not only to repair invalid ones. KernelEvolve reports
functional success on all 250 KernelBench tasks but no faster-than-reference
result~\cite{meta2025kernelevolve}, so it cannot be placed on the same
performance frontier. Table~\ref{tab:a2-correct-speed} preserves
this distinction.

Finally, a reported speedup reflects the optimization headroom of its reference.
PEAK's speedup over unoptimized input MatMul kernels and CUDA-L1's speedup over
the KernelBench reference implementations address different deployment
decisions from comparisons against TorchInductor~\cite{tariq2025peak,li2026cudal1,ansel2024pytorch2}.
Figure~\ref{fig:a2-loop}(c) therefore characterizes optimization headroom rather
than relative method quality: existing implementations may leave substantial
removable overhead, whereas surpassing an optimizing compiler is a different
performance target. For practical adoption, the relevant reference is the
implementation that would otherwise execute on the target system, and the
largest reported factor alone is insufficient to select a method.

\begin{table}[t]
\centering
\caption{$\mathrm{fast}_1$ of test-time methods on KernelBench~\cite{kernelbench2025}. Levels~1, 2,
and~3 contain 100, 100, and 50 tasks; all runs use an NVIDIA L40S GPU and the PyTorch eager
reference. G, E, and P return the previous generation, the execution result, and the timing
profile to the model. $\mathrm{fast}_1$ is the percentage of tasks that are both correct and faster
than the reference; higher is better. The $\Delta$ rows give percentage-point changes. Bold marks
column maxima; a dash denotes that repeated sampling is unavailable for the fixed-temperature
DeepSeek-R1 endpoint.}
\label{tab:a2-fast1matrix}
\footnotesize
\setlength{\tabcolsep}{5.4pt}
\renewcommand{\arraystretch}{1.06}
\begin{fittable}\begin{tabular}{@{}lcrrrrrrrrr@{}}
\toprule
& & \multicolumn{3}{c}{\textbf{Level 1}}
  & \multicolumn{3}{c}{\textbf{Level 2}}
  & \multicolumn{3}{c}{\textbf{Level 3}}\\
\cmidrule(lr){3-5}\cmidrule(lr){6-8}\cmidrule(lr){9-11}
\textbf{Test-time method} & \textbf{Calls}
& Llama-70B & V3 & R1 & Llama-70B & V3 & R1 & Llama-70B & V3 & R1\\
\midrule
Single attempt                     & 1  & 3 & 6 & 12 & 0 & 4 & 36 & 0 & 8 & 2\\
Repeated sampling                  & 10 & 5 & 11 & \na & 3 & \textbf{14} & \na & 1 & 14 & \na\\
Iterative refinement, G            & 10 & \textbf{9} & 9 & 18 & 0 & 7 & 44 & 0 & 14 & 4\\
Iterative refinement, G$+$E        & 10 & 5 & 13 & 41 & \textbf{5} & 5 & 62 & \textbf{8} & \textbf{22} & 12\\
Iterative refinement, G$+$E$+$P    & 10 & 7 & \textbf{19} & \textbf{43} & 4 & 6 & \textbf{72} & 2 & 14 & \textbf{18}\\
\midrule
\rowcolor{tabband}
\multicolumn{2}{@{}l}{$\Delta$ best $-$ single attempt}
& \gain{6} & \gain{13} & \gain{31} & \gain{5} & \gain{10} & \gain{36} & \gain{8} & \gain{14} & \gain{16}\\
\rowcolor{tabband}
\multicolumn{2}{@{}l}{$\Delta$ profiler (G$+$E $\rightarrow$ G$+$E$+$P)}
& \gain{2} & \gain{6} & \gain{2} & $-1$ & \gain{1} & \gain{10} & $-6$ & $-8$ & \gain{6}\\
\bottomrule
\end{tabular}\end{fittable}
\end{table}

\begin{table}[t]
\centering
\caption{Correctness and speed of GPU-kernel generation systems. The GPT-5-RL study evaluates
264 KernelBench tasks against TorchInductor~\cite{tehrani2026gpt5,ansel2024pytorch2}. The CUDA-L1
study evaluates the 250 KernelBench tasks on an A100 GPU against PyTorch eager and counts a task as
faster only above $1.01\times$~\cite{li2026cudal1}. KernelEvolve reports functional success on the
250 KernelBench tasks without a speed reference or device~\cite{meta2025kernelevolve}, and PEAK
reports speedups over input MatMul kernels on three devices~\cite{tariq2025peak}. Functional and
Faster are task percentages, Gap is their difference in percentage points, and Speedup is a geometric
(g), arithmetic (a), or device--precision (d) mean factor. Bold marks the best value within each
block; n/r means not reported and a dash means not applicable.}
\label{tab:a2-correct-speed}
\footnotesize
\setlength{\tabcolsep}{6pt}
\renewcommand{\arraystretch}{1.06}
\begin{fittable}\begin{tabular}{@{}llrrrr@{}}
\toprule
\textbf{System} & \textbf{Approach} & \textbf{Functional} &
\textbf{Faster} & \textbf{Gap} & \textbf{Speedup}\\
\midrule
\rowcolor{tabband}\multicolumn{6}{@{}l}{\textit{GPT-5-RL study}~\cite{tehrani2026gpt5}}\\
MakoraGenerate    & agent           & \textbf{97.4} & \textbf{72.9} & 24.5 & \textbf{2.12}$\times^{\mathrm{g}}$\\
GPT-5-RL          & single attempt  & 77.0 & 21.8 & 55.2 & 0.81$\times^{\mathrm{g}}$\\
Claude Opus 4.5   & single attempt  & 59.8 & 13.2 & 46.6 & 0.55$\times^{\mathrm{g}}$\\
Claude 4.5 Sonnet & single attempt  & 47.2 & 12.8 & 34.4 & 0.69$\times^{\mathrm{g}}$\\
GPT-5             & single attempt  & 43.7 & 14.8 & 28.9 & 0.73$\times^{\mathrm{g}}$\\
Gemini 2.5 Pro    & single attempt  & 37.7 &  8.7 & 29.0 & 0.61$\times^{\mathrm{g}}$\\
Grok 4            & single attempt  & 24.4 & 10.5 & 13.9 & 0.51$\times^{\mathrm{g}}$\\
Grok Code Fast 1  & single attempt  & 15.8 &  3.8 & 12.0 & 0.39$\times^{\mathrm{g}}$\\
\midrule
\rowcolor{tabband}\multicolumn{6}{@{}l}{\textit{CUDA-L1 study}~\cite{li2026cudal1}}\\
Llama-3.1-405B & prompting   & 27.2 & 2.0 & 25.2 & 0.23$\times^{\mathrm{a}}$\\
DeepSeek-V3    & prompting   & 39.6 & 3.6 & 36.0 & 0.34$\times^{\mathrm{a}}$\\
OpenAI-o1      & prompting   & 56.4 & 5.6 & 50.8 & 0.73$\times^{\mathrm{a}}$\\
DeepSeek-R1    & prompting   & 71.6 & 7.2 & 64.4 & 0.88$\times^{\mathrm{a}}$\\
CUDA-L1        & RL-trained  & \textbf{99.6} & \textbf{90.4} & \phantom{0}9.2 & \textbf{3.12}$\times^{\mathrm{a}}$\\
\midrule
\rowcolor{tabband}\multicolumn{6}{@{}l}{\textit{Other studies}}\\
KernelEvolve~\cite{meta2025kernelevolve} & agentic search & 100.0 & \nr & \na & \nr\\
PEAK~\cite{tariq2025peak} & guided transformations & \nr & \nr & \na & 19.52$\times^{\mathrm{d}}$\\
\bottomrule
\end{tabular}\end{fittable}
\end{table}

\begin{figure}[t]
\centering
\begin{subfigure}[t]{0.48581\textwidth}
\centering
\includegraphics[width=\linewidth]{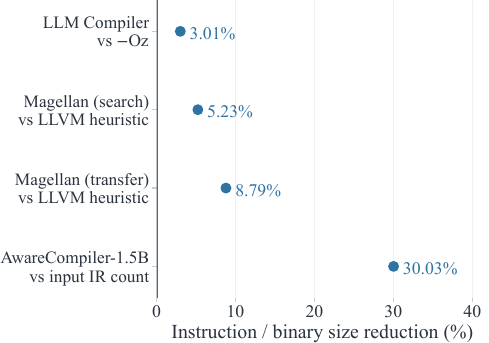}
\caption{Improvement over study-specific baselines.}
\label{fig:a3-gain}
\end{subfigure}\hfill
\begin{subfigure}[t]{0.47419\textwidth}
\centering
\includegraphics[width=\linewidth]{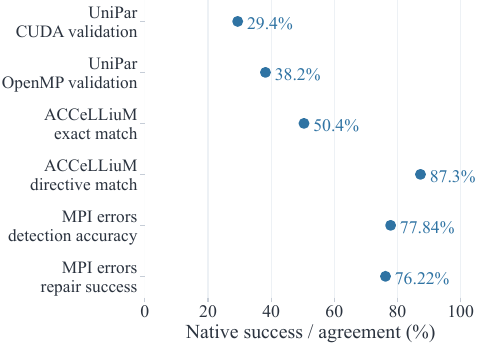}
\caption{Validation, agreement, and defect-analysis outcomes.}
\label{fig:a3-correctness}
\end{subfigure}
\caption{Source-reported outcomes in HPC-software studies. Panel~(a) shows instruction-count
or binary-size reduction under each study's reference. Panel~(b) shows validation,
textual-agreement, defect-detection, and repair rates, with the metric named per row.
Values are percentages; higher is better within the stated metric. The rows retain
native definitions and do not constitute a common correctness benchmark.}
\Description{Two plots show reported improvement percentages and correctness rates, with each metric and reference identified.}
\label{fig:a3-cost}
\end{figure}

\noindent\textbf{HPC software (A3).}
In the HPC studies, the strength of the reference is central to interpreting
progress. AwareCompiler's average IR-count reduction is substantial relative to
the input programs, but it differs from the corresponding \texttt{-Oz} average
by only 0.05 percentage points~\cite{lin2025awarecompiler}.
Figure~\ref{fig:a3-cost}(a) therefore indicates the remaining optimization
opportunities rather than a ranking by headline reduction, and workload-level
results are needed to determine where a method adds value beyond established
compiler heuristics.

The studies also differ in how optimization cost can be amortized.
LLM Compiler predicts pass sequences for individual inputs; Magellan searches
for reusable compiler heuristics; and CompilerGPT rewrites selected programs
using compiler reports~\cite{cummins2023compiler,chen2026magellan,pirkelbauer2025compilergpt}.
These outputs have different scopes of reuse: a costly heuristic search can
benefit many future compilations, whereas a source rewrite must recover its cost
through subsequent executions of a single workload. Studies should therefore
report both the optimization cost and the workload population over which the
resulting artifact is reused; the cost measurements available in
Table~\ref{tab:a3-gaincost} do not yet support a common break-even estimate.

Figure~\ref{fig:a3-cost}(b) shows a separate gap in validation depth.
ACCeLLiuM's directive agreement, UniPar's executable validation, and MPI
repair success measure different stages of establishing a usable parallel
program~\cite{jhaveri2025accellium,bitan2025unipar,piersall2026mpi}.
High agreement with a reference directive cannot substitute for execution under
concurrency, and successful repair on a defect corpus does not establish
performance portability. An evaluation that connects artifact-level agreement,
execution correctness, and performance at the intended parallel scale would make
improvements in code generation directly interpretable as improvements in HPC
engineering.

\subsection{Operational-Centric Evaluation (O1--O3)}
\label{sec:exp-operational}\label{sec:exp-integration}\label{sec:exp-operations}

\noindent\textbf{System integration (O1).}
DeployBench~\cite{wang2026deploybench} exposes the gap between producing installation actions and
establishing a working research artifact. With a common scaffold and hidden
verifier, the strongest evaluated agent completes 51.0\% of the tasks, and the
source's failure analysis attributes more than half of the failures to agents
that stop before the deployment has succeeded~\cite{wang2026deploybench}.
Completion judgment is therefore an engineering target in its own right: an
agent requires evidence that the intended execution path works, not only a
plausible sequence of setup steps. Figure~\ref{fig:o1-integration} accordingly
motivates execution-grounded stopping criteria and explicit verification of the
deployed artifact.

GenKubeSec addresses a narrower, complementary validation layer, the
identification of security misconfigurations~\cite{malul2024genkubesec}. Its high
detection scores do not establish the successful deployment or safe remediation
of an entire service. Deployment readiness, policy compliance, and service
quality thus require distinct checks that should be connected within one
integration workflow; Appendix~\ref{app:tiers} describes the corresponding
endpoints.

\noindent\textbf{Production operations (O2).}
TraceBench~\cite{egersdoerfer2026ioagent} shows that the dimensions of a useful diagnosis need not
improve together. Under the same IOAgent workflow, the two backing models obtain
nearly identical utility scores, whereas their accuracy and interpretability
differ substantially~\cite{egersdoerfer2026ioagent}. Table~\ref{tab:o2-axes} and
Figure~\ref{fig:o2-diagnosis} therefore support retaining multiple diagnostic
criteria, since a single average would obscure which aspect of the report
changes with the model. Model selection for operations should accordingly
evaluate the capability required by the intended user, such as localizing a
cause or providing an actionable explanation, rather than a single aggregate
diagnosis score.

Both IOAgent~\cite{egersdoerfer2026ioagent} variants also outperform the rule-based Drishti~\cite{drishti} and the LLM-based ION~\cite{ion}
on all reported criteria. The use of an LLM alone does not account for this
ordering: the compared systems also differ in how they structure and process
diagnostic evidence, and the experiment motivates a closer study of these workflow
components without isolating their individual effects. The scores remain
judge-assigned rankings, and the next validation step is to relate report
quality to correct interventions and observed recovery.
HPC-LLM's answer-similarity evaluation represents a related support capability,
but does not supply that operational endpoint~\cite{hpcllm2026}.

\begin{table}[t]
\centering
\caption{TraceBench normalized ranking scores and paired differences. A fixed GPT-4o ranker evaluates
every system under the same protocol~\cite{egersdoerfer2026ioagent}. Accuracy, utility,
and interpretability lie in $[0,1]$, with higher values indicating a more favorable judge-assigned rank.
Mean is their arithmetic average. The $\Delta$ rows subtract the named comparator;
positive differences favor the first term.}
\label{tab:o2-axes}
\footnotesize
\setlength{\tabcolsep}{4pt}
\renewcommand{\arraystretch}{1.14}
\begin{fittable}\begin{tabular}{@{}llllll@{}}
\toprule
\textbf{System} & \textbf{Backing} & \textbf{Accuracy} & \textbf{Utility} &
\textbf{Interp.} & \textbf{Mean}\\
\midrule
IOAgent  & GPT-4o        & 0.641 & 0.609 & 0.645 & 0.632\\
IOAgent  & Llama-3.1-70B & 0.513 & 0.607 & 0.530 & 0.550\\
Drishti~\cite{drishti} & rule-based & 0.459 & 0.436 & 0.447 & 0.447\\
ION~\cite{ion} & GPT-4o & 0.380 & 0.385 & 0.385 & 0.383\\
\midrule
\rowcolor{tabband}
$\Delta$ model swap & GPT-4o $-$ Llama & \gain{0.128} & \gain{0.002} & \gain{0.115} & \gain{0.082}\\
\rowcolor{tabband}
$\Delta$ vs.\ Drishti & IOAgent-GPT-4o & \gain{0.182} & \gain{0.173} & \gain{0.198} & \gain{0.185}\\
\rowcolor{tabband}
$\Delta$ vs.\ Drishti & IOAgent-Llama & \gain{0.054} & \gain{0.171} & \gain{0.083} & \gain{0.103}\\
\bottomrule
\end{tabular}\end{fittable}
\end{table}

\begin{figure}[t]
\centering
\begin{subfigure}[t]{0.588\textwidth}
\centering
\includegraphics[width=\linewidth]{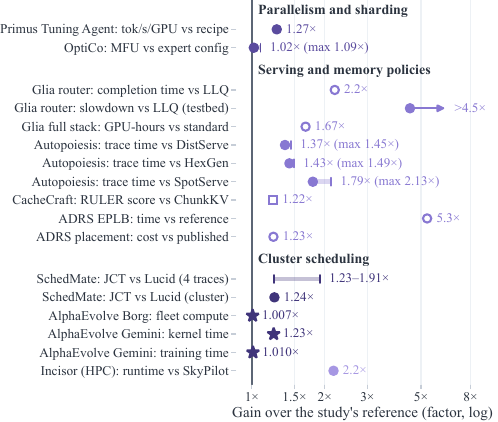}
\caption{Reported gain over each study's reference.}
\label{fig:o3-gain}
\end{subfigure}\hfill
\begin{subfigure}[t]{0.383\textwidth}
\centering
\includegraphics[width=\linewidth]{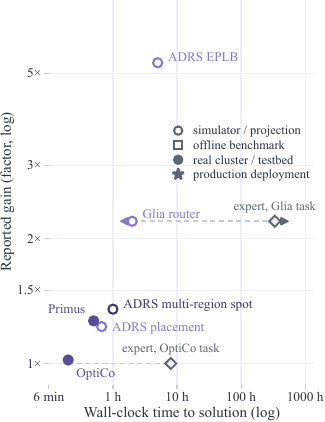}
\caption{Time to the reported solution.}
\label{fig:o3-search}
\end{subfigure}
\caption{Workload-orchestration (O3) outcomes across studies. Panel~(a) expresses each reported
result as a factor relative to the reference named in the row, computed from the source's values
(reference/method for time and cost, method/reference for throughput, MFU, and score, and
$1/(1-r)$ for a reported reduction $r$); bars show reported ranges and arrows lower bounds. Marker
shape, explained in the legend of~(b), gives the evaluation environment and color the method family. Panel~(b) plots the reported
wall-clock time to the solution against the same factor; diamonds mark the human-expert effort
that the Glia~\cite{hamadanian2025glia} and OptiCo~\cite{chen2026optico} studies report for a comparable result. Rows retain their own metrics,
workloads, and baselines (Table~\ref{tab:o3-evidence} in the appendix) and are not a ranking of methods.}
\Description{Panel a is a dot plot on a logarithmic axis of reported gain factors for O3 systems, grouped into parallelism, serving, and scheduling, with marker shapes for simulator, offline benchmark, real cluster, and production evaluations. Panel b is a log-log scatter of wall-clock time to solution against gain, with dashed lines connecting systems to the human-expert time reported for a comparable result.}
\label{fig:o3-evidence}
\end{figure}

\noindent\textbf{Workload orchestration (O3).}\label{sec:exp-orch}
The O3 studies are the first in this synthesis to measure LLM-derived decisions
on training and serving workloads at cluster scale, and
Figure~\ref{fig:o3-evidence}(a) shows that the size of a reported gain depends
strongly on where it is measured. The largest factors are obtained in
simulators and offline harnesses, for example a rebalancing algorithm for
mixture-of-experts inference whose runtime falls from 19.6 to 3.7\,ms and a
2.2$\times$ lower mean request completion time for Glia's
router~\cite{cheng2025adrs,hamadanian2025glia}. Most real-cluster results lie
between 1.02$\times$ and 1.79$\times$, from OptiCo's MFU relative to the best
expert configuration to Autopoiesis's trace time relative to
SpotServe~\cite{chen2026optico,jiang2026autopoiesis,miao2024spotserve}. The
exceptions are Glia's slowdown on a four-GPU vLLM~\cite{kwon2023vllm} testbed,
which falls by more than 4.5$\times$, and Incisor's runtime reduction for HPC
jobs~\cite{laurenzano2026incisor}. In production, AlphaEvolve's Borg heuristic
recovers 0.7\% of fleet-wide compute~\cite{novikov2025alphaevolve}. These
differences cannot be attributed to simulator optimism alone. SchedMate's
simulator deviates from its physical-cluster measurements by less than 3.7\%;
on the physical cluster, SchedMate reduces mean job completion time by a factor
of 1.24, and in simulation its factors of 1.23--1.91 span four production
traces~\cite{wang2025semantic}. Glia likewise reports a larger reduction in
slowdown on its testbed than in completion time in its simulator, although the
two metrics differ. The
environments differ instead in workload, scale, and the strength of the
reference that the method replaces. A small factor in a production fleet can
therefore represent a larger absolute saving than a large factor on a four-GPU
testbed, and simulated and deployed results should be reported separately rather
than ranked on one axis.

The strength of the reference determines what a factor means, and the
within-study contrasts in Table~\ref{tab:o3-evidence} make this explicit. Glia's
router reduces mean completion time by 2.2$\times$ relative to least-loaded
routing and by 1.6$\times$ relative to the policy found by
OpenEvolve~\cite{sharma2025openevolve} with the same simulator, and it matches a
design that took a human expert two weeks~\cite{hamadanian2025glia}. OptiCo
achieves on average 101.84\% of the MFU of the best expert-tuned
Megatron-LM~\cite{shoeybi2019megatron} configuration, a relative improvement of
1.84\%~\cite{chen2026optico}. ASAP is evaluated by
agreement with the expert sharding mesh, which it matches in all three cases, so
the speedups it reports belong to the expert configurations~\cite{ding2025asap}.
Primus~\cite{raina2026primus} provides the most explicit decomposition of an
orchestration gain. Its 27\% measured improvement over the published reference
includes a change from BF16 to FP8 precision. In its ablation, the
benchmark-calibrated scoring oracle raises projected throughput by 36\% over an
analytic oracle with the same agent, and the agent framework adds 10\% over a
general coding agent with the same oracle. With the less accurate analytic
oracle, an exhaustive non-LLM search over 530 configurations outperforms the LLM
agent, with projected gains of 8.6\% and 2.3\%, respectively~\cite{raina2026primus}.
This evidence supports LLM agents as proposers of cross-axis configurations
under an accurate evaluator, not as substitutes for one.

AlphaEvolve's Gemini results show how gains attenuate from the decision to the
job~\cite{novikov2025alphaevolve}: an average kernel speedup of 23\% reduced
overall training time by 1\%. By Amdahl's law
(Equation~\eqref{eq:amdahl}) with $n=1.23$ and $1-s$ the share of training time
spent in the tuned kernels, this corresponds to a share of roughly 5\%. Orchestration decisions act on the whole job, which is one reason why
their absolute savings can exceed those of larger kernel speedups.
Figure~\ref{fig:o3-evidence}(b) adds the search cost. The reported solutions are
found within minutes to five hours, at LLM costs of at most \$20 per ADRS case
study and \$0.26 for an OptiCo configuration~\cite{cheng2025adrs,chen2026optico},
whereas the human effort reported for comparable outcomes is about eight hours
of expert tuning and more than two weeks of expert design. OptiCo estimates
that, for GPT-3~7B on 16 A100 GPUs, its 2.0\% throughput gain (4{,}164 to
4{,}246 tokens/s/GPU) saves about 1{,}280 A100 GPU-hours over a
one-trillion-token training run, so even small orchestration gains can exceed
their search cost by a wide margin. These costs exclude the construction of the
simulators, projection engines, and evaluators on which every result depends,
which ADRS notes are difficult to design~\cite{cheng2025adrs}. Establishing
deployable value will require reporting evaluator fidelity alongside the gain,
comparing against strong non-LLM optimizers that use the same evaluator, and
validating policies on production traces.

\subsection{Resource-Centric Evaluation (R1--R2)}
\label{sec:exp-resources}

\noindent\textbf{Data management (R1).}
The studies in Figure~\ref{fig:r1-quality-cost} indicate different routes to
improving a data service. SANA~\cite{sana2026datalake} separates an implementable tool configuration
from a diagnostic condition with privileged information. On LakeQA~\cite{wang2026lakeqa}, moving
from Naive to Standard changes quality far less than moving to
Ideal~\cite{sana2026datalake}, so additional planning and retrieval machinery
does not by itself realize the headroom exposed by idealized tools. The
methodological implication is that planning, source selection, and analysis
should be diagnosed separately before workflow complexity is increased.
Retrieval calls also decrease while analysis calls increase from Naive to Ideal,
so reducing one tool counter need not reduce total work.

Evaporate illustrates a complementary approach: replacing repeated
per-document inference with reusable extraction code improves the average
reported quality while reducing LLM token use~\cite{arora2023evaporate}.
Resource efficiency can therefore depend on where model computation is placed in
a pipeline, not only on which model is used. The quality regression on Enron
nevertheless shows that reusable code requires source-specific validation, which
a lower aggregate cost does not remove; Table~\ref{tab:r1-quality-cost} retains
the per-source outcomes. Together, SANA~\cite{sana2026datalake} and Evaporate
motivate evaluating data workflows through both bottleneck diagnosis and
computation reuse, with end-to-end accounting of quality and resource
consumption. Their differing evaluation coverage and aggregation conventions
are described in Appendix~\ref{app:tiers}.

\begin{table}[t]
\centering
\caption{Quality and resource use in data-management studies. (a)~Evaporate compares direct
prompting (Direct) with synthesized extraction code (Code$+$) on five data-lake sources; Direct is
evaluated on sampled documents and Code$+$ on the full datasets~\cite{arora2023evaporate}. Quality is
OpenIE pair F1 (\%); tokens and cost are per 10K documents, and Reduction is the per-source token
ratio, averaged over sources. (b)~SANA evaluates the same corpus and model under three levels of
tool support, where Ideal is a diagnostic condition with privileged tools~\cite{sana2026datalake}.
LakeQA and KramaBench-conv.\ contain 135 and 83 tasks; quality is semantic match (\%), and the
last column gives retrieval calls per task from Naive to Ideal. Higher quality and lower resource use are better.
$\Delta$ values are percentage points; those in~(b) are as reported by the source.}
\label{tab:r1-quality-cost}
\footnotesize
\setlength{\tabcolsep}{4.6pt}
\renewcommand{\arraystretch}{1.06}

\begin{fittable}\begin{tabular}{@{}lrrcrrcrr@{}}
\multicolumn{9}{@{}l}{\textit{(a) Evaporate}~\cite{arora2023evaporate}}\\
\toprule
& \multicolumn{3}{c}{\textbf{OpenIE Pair F1}}
& \multicolumn{3}{c}{\textbf{LLM tokens (M)}}
& \multicolumn{2}{c}{\textbf{Cost (\$)}}\\
\cmidrule(lr){2-4}\cmidrule(lr){5-7}\cmidrule(lr){8-9}
\textbf{Source (format)} & Direct & Code$+$ & $\Delta$ & Direct & Code$+$ & Reduction & Direct & Code$+$\\
\midrule
FDA (TXT)            & 45.5 & 62.8 & \gain{17.3} & 145.6 & 1.9 & 77$\times$  & 2\,900 & 38\\
Enron Emails (TXT)   & 93.8 & 86.9 & $-6.9$      & \phantom{0}21.2 & 0.6 & 35$\times$  & \phantom{0,}425 & 12\\
Wiki NBA (HTML)      & 44.8 & 68.2 & \gain{23.4} & 650.1 & 3.0 & 217$\times$ & 13\,000 & 60\\
SWDE Movie (HTML)    & 45.2 & 56.8 & \gain{11.6} & 282.9 & 2.3 & 123$\times$ & 5\,660 & 46\\
SWDE University (HTML)& 53.8 & 59.0 & \gain{5.2} & 190.1 & 1.9 & 100$\times$ & 3\,800 & 38\\
\midrule
\rowcolor{tabband}
Average     & 56.6 & 66.7 & \gain{10.1} & 258.0 & 1.9 & 110$\times$ & 5\,157 & 39\\
\bottomrule
\end{tabular}\end{fittable}

\vspace{1.6ex}

\begin{fittable}\begin{tabular}{@{}llrrrrrc@{}}
\multicolumn{8}{@{}l}{\textit{(b) SANA}~\cite{sana2026datalake}}\\
\toprule
\textbf{Benchmark} & \textbf{Agent} & \textbf{Naive} & \textbf{Standard} &
\textbf{Ideal} & \textbf{$\Delta$ Standard} & \textbf{$\Delta$ Ideal} & \textbf{Retrieval calls}\\
\midrule
LakeQA          & gpt-5.4-nano & 20.7 & 19.3 & 37.0 & $-1.5$ & \gain{16.3} & 5.0 $\rightarrow$ 4.1\\
LakeQA          & gpt-5-mini   & 56.3 & 57.8 & 76.3 & \gain{1.5} & \gain{20.0} & 7.2 $\rightarrow$ 5.9\\
KramaBench-conv.  & gpt-5.4-nano & 44.6 & 57.8 & 87.9 & \gain{13.3} & \gain{43.4} & 3.2 $\rightarrow$ 1.6\\
KramaBench-conv.  & gpt-5-mini   & 62.6 & 66.3 & 90.4 & \gain{3.6} & \gain{27.7} & 3.8 $\rightarrow$ 1.2\\
\bottomrule
\end{tabular}\end{fittable}
\end{table}

\noindent\textbf{Energy optimization (R2).}
The energy results in Table~\ref{tab:r2-device} show that portability must be
assessed per application and device. LASSI-EE's cross-application averages are
similar on MI100 and A100, yet individual applications differ much more,
including an 81.9-point ER@5 difference for
\texttt{entropy}~\cite{dearing2026lassiee}. Agreement between device averages
therefore does not establish that a particular transformation transfers. The
negative margins over the vanilla baseline for \texttt{XSBench}~\cite{tramm2014xsbench} on MI100
($-2.4$ and $-4.9$ percentage points at $k{=}3$ and $k{=}5$; Table~III of the
source) further
motivate validating and selecting transformations on the deployment device
rather than treating an energy-aware workflow as a universally beneficial
replacement.

Budget selection introduces a further trade-off. The LASSI-EE~\cite{dearing2026lassiee} averages
show smaller additional savings from later candidates than from the initial
budget increase, so a deployment decision should relate this incremental
benefit to the energy and latency of further search, which the execution-energy
outcome does not include. In the study of Apsan et al.\ on EvoEval~\cite{xia2024evoeval}
problems, the ordering of human and LLM solutions reverses across platforms,
which reinforces the need for measured, device-specific
decisions~\cite{apsan2025energy}.

HyperLoad occupies a different position in the resource-management loop: it
improves a forecast that could inform a downstream
controller~\cite{jiang2025hyperload}, and its effect on energy use depends on how
that controller acts on the improved information. Table~\ref{tab:r2-forecast}
therefore records progress in a supporting capability, whereas
Table~\ref{tab:r2-device} records direct program-energy outcomes. Relating
forecast quality to executed control and measured consumption is required to
establish end-to-end resource savings.

\begin{table}[t]
\centering
\caption{Device-resolved energy outcomes of generated programs. (a)~Apsan et al.\ measure 363
solutions to nine EvoEval problems~\cite{apsan2025energy,xia2024evoeval}, using EnergiBridge on the
server and PC and a Monsoon monitor on the Raspberry Pi; values are mean energy in kJ, lower is
better, and positive differences indicate higher consumption by the LLM solutions. The LLM column
uses the Functional prompting condition. (b)~LASSI-EE reports the expected energy reduction ER@$k$
(\%) over $k$ sampled candidates for 22 applications~\cite{dearing2026lassiee}; higher is better.
Parenthesized values are margins over the vanilla LLM baseline in percentage points, and the last
column is the absolute difference between the two GPUs at $k{=}5$, with the largest in bold.}
\label{tab:r2-device}
\footnotesize
\setlength{\tabcolsep}{5pt}
\renewcommand{\arraystretch}{1.06}

\begin{fittable}\begin{tabular}{@{}lrrrr@{}}
\multicolumn{5}{@{}l}{\textit{(a) Apsan et al.}~\cite{apsan2025energy}}\\
\toprule
\textbf{Device} & \textbf{Human} & \textbf{LLM} &
\textbf{LLM $-$ human} & \textbf{Relative}\\
\midrule
Server (Xeon Silver 4208)  & \phantom{0}5.131 & \phantom{0}6.082 & \gain{0.951} & \gain{18.5}\%\\
PC (Core i9)               & 11.345 & \phantom{0}8.588 & $-2.757$ & $-24.3$\%\\
Raspberry Pi (Cortex-A72)  & \phantom{0}0.442 & \phantom{0}0.452 & \gain{0.010} & \gain{2.3}\%\\
\bottomrule
\end{tabular}\end{fittable}

\vspace{1.6ex}

\begin{fittable}\begin{tabular}{@{}lrrrrrrr@{}}
\multicolumn{8}{@{}l}{\textit{(b) LASSI-EE}~\cite{dearing2026lassiee}}\\
\toprule
& \multicolumn{3}{c}{\textbf{AMD MI100}} & \multicolumn{3}{c}{\textbf{NVIDIA A100}} & \\
\cmidrule(lr){2-4}\cmidrule(lr){5-7}
\textbf{Application} & $k{=}1$ & $k{=}3$ & $k{=}5$ & $k{=}1$ & $k{=}3$ & $k{=}5$
& \textbf{$|$MI100$-$A100$|$}\\
\midrule
\rowcolor{tabband}
Average (22 applications) & 29.6 & 49.5 & 55.2 & 28.6 & 46.9 & 54.1 & \phantom{0}1.1\\
\quad {\scriptsize $\Delta$ vs.\ vanilla} & {\scriptsize (\gain{19.1})} & {\scriptsize (\gain{22.5})} & {\scriptsize (\gain{19.2})} & {\scriptsize (\gain{18.0})} & {\scriptsize (\gain{21.4})} & {\scriptsize (\gain{20.9})} & \\
\midrule
segment-reduce      & 28.5 & 57.4 & 68.7 & 48.1 & 66.1 & 74.0 & \phantom{0}5.3\\
miniMDock           & 20.8 & 44.1 & 51.1 & 29.6 & 50.8 & 57.9 & \phantom{0}6.8\\
XSBench             & \phantom{0}3.0 & 10.6 & 16.3 & \phantom{0}5.3 & 15.8 & 19.4 & \phantom{0}3.1\\
jacobi              & 22.7 & 46.5 & 52.5 & 44.8 & 66.4 & 75.2 & 22.7\\
lid-driven-cavity   & \phantom{0}7.8 & 16.4 & 22.9 & 16.6 & 41.3 & 51.0 & 28.1\\
keogh               & 13.7 & 46.4 & 59.1 & \phantom{0}1.0 & \phantom{0}5.9 & 11.7 & 47.4\\
entropy             & 63.9 & 88.0 & 90.7 & \phantom{0}3.2 & \phantom{0}6.5 & \phantom{0}8.8 & \textbf{81.9}\\
\bottomrule
\end{tabular}\end{fittable}
\end{table}

\subsection{Cross-Category Synthesis}
\label{sec:exp-cross}

The category comparisons yield three principles for translating reported gains
into infrastructure value. First, feedback is a design component whose utility
depends on the model and the task. RTL planning, kernel refinement, and
energy-oriented candidate selection improve the evaluated workflows, whereas
the kernel ablations show that an additional feedback channel can be
counterproductive. The proposal and validation processes should therefore be
designed jointly, and the feedback that improves the target outcome should be
identified under a fixed budget. Figure~\ref{fig:loop-closure} summarizes the
within-study contrasts that support this conclusion.

Second, the most useful evaluation endpoint follows the role of the generated
output. A computational artifact must satisfy its functional contract and
improve the relevant execution objective; a diagnosis must support a correct
operational decision; an orchestration policy must improve a workload-level
objective on a representative trace or cluster; and a resource-oriented action
must preserve service quality while reducing the measured resource. The O3
evidence adds a requirement that applies across categories: the fidelity of the
evaluator bounds the claim. Gains measured on isolated kernels shrink in
end-to-end training time, and gains measured in simulators and projection
engines are not directly comparable with those measured on real clusters, so the
evaluation environment belongs in the result alongside the metric and the
reference. The taxonomy makes these
distinctions explicit. It also exposes missing connections between stages,
such as report quality without recovery measurements or forecasting accuracy
without observed facility savings. These connections are priorities for
benchmark design rather than reasons to discard the existing results.

Third, optimization cost should be assessed over the scope of reuse.
Candidate-level search, reusable compiler heuristics, and synthesized extraction
functions incur costs at different points in the lifecycle, and their practical
value depends on how often the validated result is subsequently used.
Table~\ref{tab:budget-ladder} characterizes the reported budget dependence,
while the benchmark map in Appendix~\ref{sec:appendix:evidence} identifies
where source studies share evaluation families. Together, these analyses
support comparisons that preserve setting differences and indicate which
additional measurements would establish deployment value beyond a local score.

\section{Related Work}
\label{sec:related}

Several surveys examine parts of the literature reviewed here, each organized
around a particular layer of infrastructure or class of tasks.

\noindent\textbf{Artifact generation and optimization.}
Pan et al.~\cite{pan2025llmeda} survey LLMs for electronic design automation,
and Yang et al.~\cite{yang2026verilog} focus on Verilog code generation; both
provide domain-specific accounts of hardware generation and validation.
Yu et al.~\cite{yu2026kernel} review automated GPU kernel generation through
model adaptation and agentic optimization. Ljaljevi\'{c}
et al.~\cite{ljaljevic2026hpc} survey LLMs in HPC programming, including
parallel programming and code translation. Jiang et al.~\cite{jiang2024codellm}
survey LLMs for code generation more broadly, organizing data, models,
evaluation, and software-development tasks. These specialized surveys provide a
technical depth that a broader synthesis cannot reproduce in every subfield.

\noindent\textbf{Deployment and operations.}
Pahl et al.~\cite{pahl2026iac} organize AI applications to
infrastructure-as-code across development and operational activities, and
Zhang et al.~\cite{zhang2025aiops} examine LLMs for AIOps and failure
management. Together, they motivate the distinction between generating a
deployment configuration and interpreting evidence from a running system,
although configuration, monitoring, and remediation may form successive stages
of one operational workflow.

\noindent\textbf{Systems for AI training and serving.}
Surveys of distributed training~\cite{duan2024trainsurvey} and efficient LLM
serving~\cite{miao2023servingsurvey} catalogue the parallelism, communication,
reliability, batching, and memory-management techniques that define the
workloads of Section~\ref{sec:prelim:infra}. They review systems built for AI
workloads rather than LLMs that engineer those systems, and they provide the
vocabulary used here for the workload-orchestration category.

\noindent\textbf{Data and energy.}
Hong et al.~\cite{hong2025text2sql} survey LLM-based Text-to-SQL, with
particular emphasis on query correctness in natural-language interfaces to data
systems; data management in the present taxonomy additionally considers
extraction quality, data access, and processing cost. Ali
et al.~\cite{masthanali2026sustainable} review sustainable code generation,
which overlaps the software component of energy optimization, whereas
power-oriented hardware design and facility forecasting introduce further
measurements and constraints.

\noindent\textbf{Distinctive perspective.}
This survey complements these works through its three contributions. The
cross-lifecycle survey relates work on hardware, kernels, compilers,
deployment, operations, orchestration, data, and energy to the training and
serving of AI models, rather than treating each layer separately. The
lifecycle-aligned taxonomy analyzes what the LLM manages, the feedback
available to it, and the corresponding validation requirements jointly across
these areas; a primary category organizes the presentation, while secondary
attributes preserve the overlap with the domain-specific taxonomies above. The
quantitative analysis uses primary-source outcomes to examine where LLM-driven
methods improve reported results and where differing benchmarks, references,
or evaluation environments restrict stronger conclusions.

\section{Conclusion}
\label{sec:conclusion}

This survey reviewed LLMs as engineering agents across the lifecycle of AI
infrastructure, from accelerator design and kernel development through
deployment, operation, and workload orchestration to data supply and power
management. By organizing the field into a lifecycle-aligned taxonomy,
characterizing each method under a common task formulation, and synthesizing
the reported evidence under its native metrics and settings, it shows that LLM-driven
methods now reach every stage of the lifecycle but cover few of its training- and
serving-specific concerns, that their value is
established most clearly where reliable feedback is available, and that
reported gains must be interpreted in relation to their reference and
evaluation environment. These results offer a common frame in which work from
hardware, systems, HPC, and machine learning can be positioned, compared, and
built upon. As evaluators become more faithful and validation extends from
individual stages to the hand-offs between them, LLM-driven methods can progress from optimizing isolated artifacts to
co-designing the hardware, software, and operational policies of AI
infrastructure, so that the systems used to train and serve AI models are
increasingly improved with the help of AI itself.

\bibliographystyle{ACM-Reference-Format}
\bibliography{references}

@misc{yu2026kernel,
  author    = {Yang Yu and Peiyu Zang and Chi Hsu Tsai and Haiming Wu and
               Yixin Shen and Jialing Zhang and Haoyu Wang and Zhiyou Xiao and
               Jingze Shi and Yuyu Luo and Wentao Zhang and Chunlei Men and
               Guang Liu and Yonghua Lin},
  title     = {Towards Automated Kernel Generation in the Era of {LLMs}},
  note      = {arXiv preprint arXiv:2601.15727},
  year      = {2026},
}

@inproceedings{vaswani2017attention,
  author    = {Ashish Vaswani and Noam Shazeer and Niki Parmar and
               Jakob Uszkoreit and Llion Jones and Aidan N. Gomez and
               {\L}ukasz Kaiser and Illia Polosukhin},
  title     = {Attention Is All You Need},
  booktitle = {Advances in Neural Information Processing Systems (NeurIPS)},
  pages     = {5998--6008},
  publisher = {Curran Associates, Inc.},
  address   = {Red Hook, NY, USA},
  year      = {2017},
}

@misc{kaplan2020scaling,
  author    = {Jared Kaplan and Sam McCandlish and Tom Henighan and
               Tom B. Brown and Benjamin Chess and Rewon Child and
               Scott Gray and Alec Radford and Jeffrey Wu and Dario Amodei},
  title     = {Scaling Laws for Neural Language Models},
  note      = {arXiv preprint arXiv:2001.08361},
  year      = {2020},
}

@inproceedings{dao2022flashattention,
  author    = {Tri Dao and Dan Fu and Stefano Ermon and Atri Rudra and
               Christopher R\'{e}},
  title     = {{FlashAttention}: Fast and Memory-Efficient Exact Attention
               with {IO}-Awareness},
  booktitle = {Advances in Neural Information Processing Systems (NeurIPS)},
  pages     = {16344--16359},
  publisher = {Curran Associates, Inc.},
  address   = {Red Hook, NY, USA},
  year      = {2022},
}

@misc{liu2023verilogeval,
  author    = {Mingjie Liu and Nathaniel Pinckney and Brucek Khailany and Haoxing Ren},
  title     = {{VerilogEval}: Evaluating Large Language Models for {Verilog}
               Code Generation},
  year      = {2023},
  note      = {arXiv preprint arXiv:2309.07544},
}

@misc{lu2023rtllm,
  author    = {Yao Lu and Shang Liu and Qijun Zhang and Zhiyao Xie},
  title     = {{RTLLM}: An Open-Source Benchmark for Design {RTL} Generation
               with Large Language Model},
  year      = {2023},
  note      = {arXiv preprint arXiv:2308.05345},
}

@misc{liu2023rtlcoder,
  author    = {Shang Liu and Wenji Fang and Yao Lu and Qijun Zhang and
               Hongce Zhang and Zhiyao Xie},
  title     = {{RTLCoder}: Outperforming {GPT-3.5} in Design {RTL} Generation
               with Our Open-Source Dataset and Lightweight Solution},
  year      = {2023},
  note      = {arXiv preprint arXiv:2312.08617},
}

@misc{chang2023chipgpt,
  author    = {Kaiyan Chang and Ying Wang and Haimeng Ren and Mengdi Wang and
               Shengwen Liang and Yinhe Han and Huawei Li and Xiaowei Li},
  title     = {{ChipGPT}: How Far Are We from Natural Language Hardware Design},
  year      = {2023},
  note      = {arXiv preprint arXiv:2305.14019},
}

@misc{pinckney2024revisiting,
  author    = {Nathaniel Pinckney and Christopher Batten and Mingjie Liu and
               Haoxing Ren and Brucek Khailany},
  title     = {Revisiting {VerilogEval}: A Year of Improvements in
               Large-Language Models for Hardware Code Generation},
  year      = {2024},
  note      = {arXiv preprint arXiv:2408.11053},
}

@misc{thorat2025verippa,
  author    = {Kiran Thorat and Jiahui Zhao and Yaotian Liu and Amit Hasan and
               Hongwu Peng and Xi Xie and Bin Lei and Caiwen Ding},
  title     = {{LLM-VeriPPA}: Power, Performance, and Area Optimization Aware
               {Verilog} Code Generation with Large Language Models},
  year      = {2025},
  note      = {arXiv preprint arXiv:2510.15899},
}

@misc{hsin2026evolve,
  author    = {Wei-Po Hsin and Ren-Hao Deng and Yao-Ting Hsieh and
               En-Ming Huang and Shih-Hao Hung},
  title     = {{EvolVE}: Evolutionary Search for {LLM}-Based {Verilog}
               Generation and Optimization},
  year      = {2026},
  note      = {arXiv preprint arXiv:2601.18067},
}

@misc{kernelbench2025,
  author    = {Anne Ouyang and Simon Guo and Simran Arora and
               Alex L. Zhang and William Hu and Christopher R\'{e} and
               Azalia Mirhoseini},
  title     = {{KernelBench}: Can {LLMs} Write Efficient {GPU} Kernels?},
  year      = {2025},
  note      = {arXiv preprint arXiv:2502.10517},
}

@misc{tritonbench2025,
  author    = {Jianling Li and Shangzhan Li and Zhenye Gao and Qi Shi and
               Yuxuan Li and Zefan Wang and Jiacheng Huang and Haojie Wang and
               Jianrong Wang and Xu Han and Zhiyuan Liu and Maosong Sun},
  title     = {{TritonBench}: Benchmarking Large Language Model Capabilities
               for Generating {Triton} Operators},
  year      = {2025},
  note      = {arXiv preprint arXiv:2502.14752},
}

@misc{wen2025multikernelbench,
  author    = {Zhongzhen Wen and Yinghui Zhang and Zhong Li and
               Zhongxin Liu and Linna Xie and Tian Zhang},
  title     = {{MultiKernelBench}: A Multi-Platform Benchmark for Kernel
               Generation},
  year      = {2025},
  note      = {arXiv preprint arXiv:2507.17773},
}

@misc{wang2026kernelbenchx,
  author    = {Han Wang and Jintao Zhang and Kai Jiang and Haoxu Wang and
               Jianfei Chen and Jun Zhu},
  title     = {{KernelBenchX}: A Comprehensive Benchmark for Evaluating
               {LLM}-Generated {GPU} Kernels},
  year      = {2026},
  note      = {arXiv preprint arXiv:2605.04956},
}

@misc{tehrani2026gpt5,
  author    = {Ali Tehrani and Yahya Emara and Essam Wissam and
               Wojciech Paluch and Waleed Atallah and
               {\L}ukasz Dudziak and Mohamed S. Abdelfattah},
  title     = {Fine-Tuning {GPT-5} for {GPU} Kernel Generation},
  year      = {2026},
  note      = {arXiv preprint arXiv:2602.11000},
}

@misc{tariq2025peak,
  author    = {Muhammad Usman Tariq and Abhinav Jangda and
               Angelica Moreira and Madan Musuvathi and Tyler Sorensen},
  title     = {{PEAK}: A Performance Engineering {AI}-Assistant for {GPU}
               Kernels Powered by Natural Language Transformations},
  year      = {2025},
  note      = {arXiv preprint arXiv:2512.19018},
}

@misc{sarkar2026correctness,
  author    = {Dipankar Sarkar},
  title     = {The Correctness Illusion in {LLM}-Generated {GPU} Kernels},
  year      = {2026},
  note      = {arXiv preprint arXiv:2606.20128},
}

@misc{ping2026poet,
  author    = {Heng Ping and Peiyu Zhang and Zhenkun Wang and Shixuan Li and
               Anzhe Cheng and Wei Yang and Paul Bogdan and Shahin Nazarian},
  title     = {{POET}: Power-Oriented Evolutionary Tuning for {LLM}-Based
               {RTL} {PPA} Optimization},
  year      = {2026},
  note      = {arXiv preprint arXiv:2603.19333},
}

@misc{cummins2023compiler,
  author    = {Chris Cummins and Volker Seeker and Dejan Grubisic and
               Mostafa Elhoushi and Youwei Liang and Baptiste Roziere and
               Jonas Gehring and Fabian Gloeckle and Kim Hazelwood and
               Gabriel Synnaeve and Hugh Leather},
  title     = {Large Language Models for Compiler Optimization},
  year      = {2023},
  note      = {arXiv preprint arXiv:2309.07062},
}

@misc{lin2025awarecompiler,
  author    = {Hongyu Lin and Haolin Pan and Haoran Luo and Yuchen Li and
               Kaichun Yao and Libo Zhang and Mingjie Xing and Yanjun Wu},
  title     = {{AwareCompiler}: Agentic Context-Aware Compiler Optimization
               via a Synergistic Knowledge-Data Driven Framework},
  year      = {2025},
  note      = {arXiv preprint arXiv:2510.11759},
}

@misc{chen2026magellan,
  author    = {Hongzheng Chen and Alexander Novikov and Ng{\^a}n V{\~u} and
               Hanna Alam and Zhiru Zhang and Aiden Grossman and
               Mircea Trofin and Amir Yazdanbakhsh},
  title     = {{Magellan}: Autonomous Discovery of Novel Compiler Optimization
               Heuristics with {AlphaEvolve}},
  year      = {2026},
  note      = {arXiv preprint arXiv:2601.21096},
}

@misc{pirkelbauer2025compilergpt,
  author    = {Peter Pirkelbauer and Chunhua Liao},
  title     = {{CompilerGPT}: Leveraging Large Language Models for Analyzing
               and Acting on Compiler Optimization Reports},
  year      = {2025},
  note      = {arXiv preprint arXiv:2506.06227},
}

@misc{bitan2025unipar,
  author    = {Tomer Bitan and Tal Kadosh and Erel Kaplan and Shira Meiri and
               Le Chen and Peter Morales and Niranjan Hasabnis and Gal Oren},
  title     = {{UniPar}: A Unified {LLM}-Based Framework for Parallel and
               Accelerated Code Translation in {HPC}},
  year      = {2025},
  note      = {arXiv preprint arXiv:2509.12136},
}

@misc{jhaveri2025accellium,
  author    = {Samyak Jhaveri and Vanessa Klotzmann and Crista Lopes},
  title     = {{ACCeLLiuM}: Supervised Fine-Tuning for Automated {OpenACC}
               Pragma Generation},
  year      = {2025},
  note      = {arXiv preprint arXiv:2509.20380},
}

@misc{davis2025parevalrepo,
  author    = {Joshua H. Davis and Daniel Nichols and Ishan Khillan and
               Abhinav Bhatele},
  title     = {{ParEval-Repo}: A Benchmark Suite for Evaluating {LLMs} with
               Repository-Level {HPC} Translation Tasks},
  year      = {2025},
  note      = {arXiv preprint arXiv:2506.20938},
}

@misc{piersall2026mpi,
  author    = {Scott Piersall and Yang Gao and Shenyang Liu and Liqiang Wang},
  title     = {Improving {MPI} Error Detection and Repair with Large Language
               Models and Bug References},
  year      = {2026},
  note      = {arXiv preprint arXiv:2604.02398},
}

@misc{wang2026deploybench,
  author    = {Yuanli Wang and Yaoyao Qian and Yue Zhang and Hanhan Zhou and
               Jindan Huang and Tianfu Fu and Qiuyang Mang and Huanzhi Mao and
               Wenhao Chai and Wendong Fan and Liqiang Jing},
  title     = {{DeployBench}: Benchmarking {LLM} Agents for Research Artifact
               Deployment},
  year      = {2026},
  note      = {arXiv preprint arXiv:2606.05238},
}

@misc{xu2026aiconfigurator,
  author    = {Tianhao Xu and Yiming Liu and Xianglong Lu and Yijia Zhao and
               Xuting Zhou and Aichen Feng and Yiyi Chen and Yi Shen and
               Qin Zhou and Xumeng Chen and Ilya Sherstyuk and Haorui Li and
               Rishi Thakkar and Ben Hamm and Yuanzhe Li and Xue Huang and
               Wenpeng Wu and Anish Shanbhag and Harry Kim and Chuan Chen and
               Junjie Lai},
  title     = {{AIConfigurator}: Lightning-Fast Configuration Optimization for
               Multi-Framework {LLM} Serving},
  year      = {2026},
  note      = {arXiv preprint arXiv:2601.06288},
}

@misc{ueno2024migrating,
  author    = {Masaru Ueno and Tetsuya Uchiumi},
  title     = {Migrating Existing Container Workload to {Kubernetes}:
               {LLM}-Based Approach and Evaluation},
  year      = {2024},
  note      = {arXiv preprint arXiv:2408.11428},
}

@misc{malul2024genkubesec,
  author    = {Ehud Malul and Yair Meidan and Dudu Mimran and
               Yuval Elovici and Asaf Shabtai},
  title     = {{GenKubeSec}: {LLM}-Based {Kubernetes} Misconfiguration
               Detection, Localization, Reasoning, and Remediation},
  year      = {2024},
  note      = {arXiv preprint arXiv:2405.19954},
}

@misc{zhou2026argus,
  author    = {Jiasheng Zhou and Longbin Zeng and Clavis Chen and
               Ruiming Lu and Qinwei Yang and Leyi Ye and Ray Ying and
               Key Zhang},
  title     = {{ARGUS}: Production-Scale Tracing and Performance Diagnosis
               for over 10,000-{GPU} Clusters},
  year      = {2026},
  note      = {arXiv preprint arXiv:2606.20374},
}

@misc{sun2026metakube,
  author    = {Wei Sun and Ting Wang and Xinran Tian and Wanshun Lan and
               Xuhan Feng and Haoyue Li and Fangxin Wang},
  title     = {{MetaKube}: An Experience-Aware {LLM} Framework for
               {Kubernetes} Failure Diagnosis},
  year      = {2026},
  note      = {arXiv preprint arXiv:2603.23580},
}

@misc{xiang2025stategraph,
  author    = {Yong Xiang and Charley Peter Chen and Liyi Zeng and Wei Yin and
               Xin Liu and Hu Li and Wei Xu},
  title     = {Simplifying Root Cause Analysis in {Kubernetes} with
               {StateGraph} and {LLM}},
  year      = {2025},
  note      = {arXiv preprint arXiv:2506.02490},
}

@misc{tran2026rca,
  author    = {Nguyen Phuc Tran and Brigitte Jaumard and Oscar Delgado and
               Tristan Glatard and Karthikeyan Premkumar and Kun Ni},
  title     = {{LLM}-Augmented Knowledge Base Construction for Root Cause
               Analysis},
  year      = {2026},
  note      = {arXiv preprint arXiv:2604.06171},
}

@misc{egersdoerfer2026ioagent,
  author    = {Chris Egersdoerfer and Arnav Sareen and Jean Luca Bez and
               Suren Byna and Dongkuan Xu and Dong Dai},
  title     = {{IOAgent}: Democratizing Trustworthy {HPC} {I/O} Performance
               Diagnosis Capability via {LLMs}},
  year      = {2026},
  note      = {arXiv preprint arXiv:2602.22017},
}

@misc{karimi2026logs,
  author    = {Ahmad Maroof Karimi and Jong Youl Choi and Charles Qing Cao and
               Awais Khan},
  title     = {Instruction-Tuned {LLMs} for Parsing and Mining Unstructured
               Logs on Leadership {HPC} Systems},
  year      = {2026},
  note      = {arXiv preprint arXiv:2604.05168},
}

@misc{wang2025semantic,
  author    = {Zerui Wang and Qinghao Hu and Ana Klimovic and
               Tianwei Zhang and Yonggang Wen and Peng Sun and Dahua Lin},
  title     = {Semantic-Aware Scheduling for {GPU} Clusters with Large
               Language Models},
  year      = {2025},
  note      = {arXiv preprint arXiv:2510.03334},
}

@misc{sharma2025workload,
  author    = {Aasish Kumar Sharma and Julian Kunkel},
  title     = {Evaluating Large Language Models for Workload Mapping and
               Scheduling in Heterogeneous {HPC} Systems},
  year      = {2025},
  note      = {arXiv preprint arXiv:2511.11612},
}

@misc{laurenzano2026incisor,
  author    = {Michael A. Laurenzano and Shihan Cheng and David A. B. Hyde},
  title     = {{Incisor}: Ex Ante Cloud Instance Selection for {HPC} Jobs},
  year      = {2026},
  note      = {arXiv preprint arXiv:2604.24464},
}

@misc{apsan2025energy,
  author    = {Radu Apsan and Vincenzo Stoico and Michel Albonico and
               Rudra Dhar and Karthik Vaidhyanathan and Ivano Malavolta},
  title     = {Generating Energy-Efficient Code via Large-Language Models --
               Where Are We Now?},
  year      = {2026},
  note      = {Proceedings of the 48th International Conference on Software
               Engineering (ICSE), arXiv:2509.10099},
}

@misc{jiang2025hyperload,
  author    = {Haoyu Jiang and Boan Qu and Junjie Zhu and Fanjie Zeng and
               Xiaojie Lin and Wei Zhong},
  title     = {{HyperLoad}: A Cross-Modality Enhanced Large Language
               Model-Based Framework for Green Data Center Cooling Load
               Prediction},
  year      = {2025},
  note      = {arXiv preprint arXiv:2512.19114},
}

@article{arora2023evaporate,
  author    = {Simran Arora and Brandon Yang and Sabri Eyuboglu and
               Avanika Narayan and Andrew Hojel and Immanuel Trummer and
               Christopher R\'{e}},
  title = {{Language Models Enable Simple Systems for Generating Structured Views of Heterogeneous Data Lakes}},
  journal   = {Proceedings of the VLDB Endowment},
  volume = {17},
  number = {2},
  pages = {92--105},
  year      = {2023},
  doi       = {10.14778/3626292.3626294},
  url       = {https://www.vldb.org/pvldb/vol17/p92-arora.pdf},
  note      = {arXiv:2304.09433},
}

@misc{dearing2026lassiee,
  author    = {Matthew T. Dearing and Yiheng Tao and Zhiling Lan and
               Xingfu Wu and Valerie Taylor},
  title     = {Leveraging {LLMs} to Automate Energy-Aware Refactoring of
               Parallel Scientific Codes},
  year      = {2025},
  note      = {arXiv preprint arXiv:2505.02184},
}

@misc{cappendijk2024energyprompt,
  author    = {Tom Cappendijk and Pepijn de Reus and Ana Oprescu},
  title     = {{Generating Energy-efficient code with LLMs}},
  year      = {2024},
  note      = {arXiv preprint arXiv:2411.10599},
}

@misc{peng2024energycode,
  author    = {Huiyun Peng and Arjun Gupte and Nicholas John Eliopoulos and
               Chien Chou Ho and Rishi Mantri and Leo Deng and Wenxin Jiang and
               Yung-Hsiang Lu and Konstantin L\"{a}ufer and George K. Thiruvathukal
               and James C. Davis},
  title     = {Large Language Models for Energy-Efficient Code: Emerging
               Results and Future Directions},
  year      = {2024},
  note      = {arXiv preprint arXiv:2410.09241},
}

@misc{an2025ledd,
  author    = {Qi An and Chihua Ying and Yuqing Zhu and Yihao Xu and
               Manwei Zhang and Jianmin Wang},
  title     = {{LEDD}: Large Language Model-Empowered Data Discovery in Data Lakes},
  year      = {2025},
  eprint    = {2502.15182},
  archivePrefix = {arXiv},
  primaryClass = {cs.DB},
  doi       = {10.48550/arXiv.2502.15182},
  url       = {https://arxiv.org/abs/2502.15182},
}

@misc{bhoite2025contracts,
  author    = {Harshraj Bhoite},
  title     = {{AI}-Driven Generation of Data Contracts in Modern Data
               Engineering Systems},
  year      = {2025},
  note      = {arXiv preprint arXiv:2507.21056},
}

@misc{jiang2024codellm,
  author    = {Juyong Jiang and Fan Wang and Jiasi Shen and
               Sungju Kim and Sunghun Kim},
  title     = {A Survey on Large Language Models for Code Generation},
  year      = {2024},
  note      = {arXiv preprint arXiv:2406.00515},
}

@misc{pan2025llmeda,
  author    = {Jingyu Pan and Guanglei Zhou and Chen-Chia Chang and
               Isaac Jacobson and Jiang Hu and Yiran Chen},
  title     = {A Survey of Research in Large Language Models for Electronic
               Design Automation},
  year      = {2025},
  note      = {arXiv preprint arXiv:2501.09655},
}

@misc{sana2026datalake,
  author    = {Austin Senna Wijaya and Jiaxiang Liu and Haonan Wang and Eugene Wu},
  title     = {{SANA}: What Matters for {QA} Agents over Massive Data Lakes?},
  year      = {2026},
  note      = {arXiv preprint arXiv:2606.13904},
}

@misc{dataflow2025prep,
  author    = {Hao Liang and Xiaochen Ma and Zhou Liu and Zhen Hao Wong and
               Zhengyang Zhao and Zimo Meng and others},
  title     = {{DataFlow}: An {LLM}-Driven Framework for Unified Data
               Preparation and Workflow Automation in the Era of Data-Centric {AI}},
  year      = {2025},
  note      = {arXiv preprint arXiv:2512.16676},
}

@misc{lakegen2025tabular,
  author    = {Zhenwei Dai and Chuan Lei and Asterios Katsifodimos and others},
  title     = {{LAKEGEN}: A {LLM}-based Tabular Corpus Generator for Evaluating
               Dataset Discovery in Data Lakes},
  year      = {2025},
  note      = {arXiv preprint arXiv:2507.04687},
}

@misc{meta2025kernelevolve,
  author    = {Liao, Gang and Qin, Hongsen and Wang, Ying and Golden, Alicia and Kuchnik, Michael and Yetim, Yavuz and Ang, Jia Jiunn and Fu, Chunli and He, Yihan and Hsia, Samuel and Jiang, Zewei and Li, Dianshi and Pashkevich, Uladzimir and Puvvada, Varna and Shi, Feng and Steiner, Matt and Xiao, Ruichao and Li, Liyuan and Yan, Nathan and Yu, Xiayu and Fang, Zhou and Levenstein, Roman and Ho, Kunming and Zhu, Haishan and Hammond, Alec and Li, Richard and Mathews, Ajit and Gondkar, Kaustubh and Zainul-Abedin, Abdul and Singh, Ketan and Yu, Hongtao and Chi, Wenyuan and Huang, Barney and Zhang, Sean and Weller, Noah and Marine, Zach and Cook, Wyatt and Wu, Carole-Jean and Liu, Gaoxiang},
  title     = {{KernelEvolve}: Scaling Agentic Kernel Coding for Heterogeneous
               {AI} Accelerators at Meta},
  year      = {2025},
  note      = {arXiv preprint arXiv:2512.23236},
}

@misc{li2026cudal1,
  author    = {Xiaoya Li and Albert Wang and Guoyin Wang and Jiwei Li and Chris Shum},
  title     = {{CUDA-L1}: Improving {CUDA} Optimization via Contrastive
               Reinforcement Learning},
  year      = {2025},
  note      = {arXiv preprint arXiv:2507.14111},
}

@misc{wang2026fact,
  author    = {Qiuyi Qu and Yicheng Sui and Yufei Sun and others},
  title     = {A Two-Stage {GPU} Kernel Tuner Combining Semantic Refactoring
               and Search-Based Optimization},
  year      = {2026},
  note      = {arXiv preprint arXiv:2601.12698},
}

@misc{bao2026analogagent,
  author    = {Zhixuan Bao and Zhuoyi Lin and Jiageng Wang and others},
  title     = {{AnalogAgent}: Self-Improving Analog Circuit Design Automation
               with {LLM} Agents},
  year      = {2026},
  note      = {arXiv preprint arXiv:2603.23910},
}

@misc{deng2026acertl,
  author    = {Chenhui Deng and Zhongzhi Yu and Guan-Ting Liu and others},
  title     = {{ACE-RTL}: When Agentic Context Evolution Meets {RTL}-Specialized {LLMs}},
  year      = {2026},
  note      = {arXiv preprint arXiv:2602.10218},
}

@misc{eryilmaz2026chipcraftbrain,
  author    = {Cagri Eryilmaz},
  title     = {{ChipCraftBrain}: Validation-First {RTL} Generation via
               Multi-Agent Orchestration},
  year      = {2026},
  note      = {arXiv preprint arXiv:2604.19856},
}

@misc{nallathambi2026a3d,
  author    = {Abinand Nallathambi and Christopher Knight and Shantanu Ganguly and others},
  title     = {{A3D}: Agentic {AI} flow for autonomous Accelerator Design},
  year      = {2026},
  note      = {arXiv preprint arXiv:2605.15237},
}

@misc{luo2026clover,
  author    = {Zizhang Luo and Yansong Xu and Runlin Guo and others},
  title     = {{Clover}: A Neural-Symbolic Agentic Harness with Stochastic
               Tree-of-Thoughts for Verified {RTL} Repair},
  year      = {2026},
  note      = {arXiv preprint arXiv:2604.17288},
}

@misc{ping2026coevo,
  author    = {Heng Ping and Peiyu Zhang and Shixuan Li and others},
  title     = {{COEVO}: Co-Evolutionary Framework for Joint Functional Correctness
               and {PPA} Optimization in {LLM}-Based {RTL} Generation},
  year      = {2026},
  note      = {arXiv preprint arXiv:2604.15001},
}

@misc{li2026autoppa,
  author    = {Chongxiao Li and Pengwei Jin and Di Huang and others},
  title     = {{AutoPPA}: Automated Circuit {PPA} Optimization via Contrastive
               Code-based Rule Library Learning},
  year      = {2026},
  note      = {arXiv preprint arXiv:2604.18445},
}

@misc{contracts2025ai,
  author    = {Jiale Liu and Taiyu Zhou and Tianqi Jiang},
  title     = {{Veri-Sure: Multi-Agent RTL Code Generation with Temporal Tracing, Slicing and Formal Verification}},
  year      = {2026},
  note      = {arXiv preprint arXiv:2601.19747},
}

@misc{wu2025geak,
  author    = {Jianghui Wang and Vinay Joshi and Saptarshi Majumder and others},
  title     = {{Geak}: Introducing Triton Kernel {AI} Agent \& Evaluation Benchmarks},
  year      = {2025},
  note      = {arXiv preprint arXiv:2507.23194},
}

@misc{chen2025datapipeline,
  author    = {Sylvio Barbon Junior and Paolo Ceravolo and Sven Groppe and others},
  title     = {Are Large Language Models the New Interface for Data Pipelines?},
  year      = {2024},
  note      = {arXiv preprint arXiv:2406.06596},
}

@misc{wang2026energykernel,
  author    = {Shiva Ahir and Alex Doboli},
  title     = {{RAG}-Enhanced Kernel-Based Heuristic Synthesis ({RKHS}): A Structured
               Methodology Using Large Language Models for Hardware Design},
  year      = {2026},
  note      = {arXiv preprint arXiv:2604.26153},
}

@misc{yu2026autosizer,
  author    = {Xi Yu and Dmitrii Torbunov and Soumyajit Mandal and Yihui Ren},
  title     = {{AutoSizer}: Automatic Sizing of Analog and Mixed-Signal Circuits
               via Large Language Model ({LLM}) Agents},
  year      = {2026},
  note      = {arXiv preprint arXiv:2602.02849},
}

@misc{zhang2026htam,
  author    = {Yining Zhang and Mingyang Yi and Chen Wang and others},
  title     = {{HTAM}: Hierarchical Transition-Attended Memory for Operator Optimization},
  year      = {2026},
  note      = {arXiv preprint arXiv:2605.29734},
}

@misc{hpcllm2026,
  author    = {Nourin Shahin and Izzat Alsmadi},
  title     = {{HPC-LLM}: Practical Domain Adaptation and Retrieval-Augmented
               Generation for {HPC} Support},
  year      = {2026},
  note      = {arXiv preprint arXiv:2605.16347},
}

@misc{chen2026passnet,
  author    = {Yiqun Liu and Yingsheng Wu and Ruqi Yang and others},
  title     = {{PassNet}: Scaling Large Language Models for Graph Compiler Pass Generation},
  year      = {2026},
  note      = {arXiv preprint arXiv:2605.29357},
}

@inproceedings{drishti,
  author = {Jean Luca Bez and Hammad Ather and Suren Byna},
  title = {{Drishti: Guiding End-Users in the I/O Optimization Journey}},
  booktitle = {2022 IEEE/ACM International Parallel Data Systems Workshop (PDSW)},
  year = {2022},
  pages = {1--6},
  doi = {10.1109/PDSW56643.2022.00006},
  url = {https://cdux.cs.uoregon.edu/pubs/BezDrishti.pdf},
}

@inproceedings{ion,
  author = {Chris Egersdoerfer and Arnav Sareen and Jean Luca Bez and Suren Byna and Dong Dai},
  title = {{ION: Navigating the HPC I/O Optimization Journey using Large Language Models}},
  booktitle = {Proceedings of the 16th ACM Workshop on Hot Topics in Storage and File Systems},
  year = {2024},
  pages = {86--92},
  doi = {10.1145/3655038.3665950},
  url = {https://escholarship.org/uc/item/4d4097dx},
}

@misc{rp260101765,
  author        = {Yao Lu and Shang Liu and Hangan Zhou and Wenji Fang and Qijun Zhang and Zhiyao Xie},
  title         = {{A New Benchmark for the Appropriate Evaluation of RTL Code Optimization}},
  year          = {2026},
  eprint        = {2601.01765},
  archivePrefix = {arXiv},
  primaryClass  = {cs.AI},
  doi           = {10.48550/arXiv.2601.01765},
  url           = {https://arxiv.org/abs/2601.01765},
}

@misc{rp260415082,
  author        = {Cunxi Yu and Haoxing Ren},
  title         = {{Autonomous Evolution of EDA Tools: Multi-Agent Self-Evolved ABC}},
  year          = {2026},
  eprint        = {2604.15082},
  archivePrefix = {arXiv},
  primaryClass  = {cs.AR},
  doi           = {10.48550/arXiv.2604.15082},
  url           = {https://arxiv.org/abs/2604.15082},
}

@misc{rp260311489,
  author        = {Yan Tan and Xiangchen Meng and Zijun Jiang and Yangdi Lyu},
  title         = {{AutoVeriFix+: High-Correctness RTL Generation via Trace-Aware Causal Fix and Semantic Redundancy Pruning}},
  year          = {2026},
  eprint        = {2603.11489},
  archivePrefix = {arXiv},
  primaryClass  = {cs.PL},
  doi           = {10.48550/arXiv.2603.11489},
  url           = {https://arxiv.org/abs/2603.11489},
}

@misc{rp260506936,
  author        = {Pengju Liu and Nuo Xu and Jinwei Tang and Yu Cao and Caiwen Ding},
  title         = {{Bridging the Last Mile of Circuit Design: PostEDA-Bench, a Hierarchical Benchmark for PPA Convergence and DRC Fixing}},
  year          = {2026},
  eprint        = {2605.06936},
  archivePrefix = {arXiv},
  primaryClass  = {cs.AR},
  doi           = {10.48550/arXiv.2605.06936},
  url           = {https://arxiv.org/abs/2605.06936},
}

@misc{rp260605680,
  author        = {Mohammad Akyash and Nowfel Mashnoor and Kimia Azar and Hadi Kamali},
  title         = {{CASS-RTL: Correctness-Aware Subspace Steering for RTL Generation with LLMs}},
  year          = {2026},
  eprint        = {2606.05680},
  archivePrefix = {arXiv},
  primaryClass  = {cs.PL},
  doi           = {10.48550/arXiv.2606.05680},
  url           = {https://arxiv.org/abs/2606.05680},
}

@misc{rp260104505,
  author        = {Khandakar Shakib Al Hasan and Syed Rifat Raiyan and Hasin Mahtab Alvee and Wahid Sadik},
  title         = {{CircuitLM: A Multi-Agent LLM-Aided Design Framework for Generating Circuit Schematics from Natural Language Prompts}},
  year          = {2026},
  eprint        = {2601.04505},
  archivePrefix = {arXiv},
  primaryClass  = {cs.AI},
  doi           = {10.48550/arXiv.2601.04505},
  url           = {https://arxiv.org/abs/2601.04505},
}

@misc{rp260507782,
  author        = {Jing Xiong and Qi Han and Chenchen Ding and He Xiao and Zunhai Su and Chaofan Tao and Ngai Wong},
  title         = {{CktFormalizer: Autoformalization of Natural Language into Circuit Representations}},
  year          = {2026},
  eprint        = {2605.07782},
  archivePrefix = {arXiv},
  primaryClass  = {cs.CL},
  doi           = {10.48550/arXiv.2605.07782},
  url           = {https://arxiv.org/abs/2605.07782},
}

@misc{rp260317204,
  author        = {Che-Ming Chang and Prashanth Vijayaraghavan and Ashutosh Jadhav and Charles Mackin and Vandana Mukherjee and Hsinyu Tsai and Ehsan Degan},
  title         = {{CODMAS: A Dialectic Multi-Agent Collaborative Framework for Structured RTL Optimization}},
  year          = {2026},
  eprint        = {2603.17204},
  archivePrefix = {arXiv},
  primaryClass  = {cs.CL},
  doi           = {10.48550/arXiv.2603.17204},
  url           = {https://arxiv.org/abs/2603.17204},
}

@misc{rp260417102,
  author        = {Minghao Shao and Zeng Wang and Weimin Fu and Xiaolong Guo and Johann Knechtel and Ozgur Sinanoglu and Ramesh Karri and Muhammad Shafique},
  title         = {{Configuration Over Selection: Hyperparameter Sensitivity Exceeds Model Differences in Open-Source LLMs for RTL Generation}},
  year          = {2026},
  eprint        = {2604.17102},
  archivePrefix = {arXiv},
  primaryClass  = {cs.AR},
  doi           = {10.48550/arXiv.2604.17102},
  url           = {https://arxiv.org/abs/2604.17102},
}

@misc{rp260517892,
  author        = {Shuo Yin and Yihe Wang and Lancheng Zou and Xufeng Yao and Tinghuan Chen and Chen Bai and Zhengrong Wang and Tsung-Yi Ho and Bei Yu},
  title         = {{CPPL: A Circuit Prompt Programming Language}},
  year          = {2026},
  eprint        = {2605.17892},
  archivePrefix = {arXiv},
  primaryClass  = {cs.AR},
  doi           = {10.48550/arXiv.2605.17892},
  url           = {https://arxiv.org/abs/2605.17892},
}

@misc{rp260505170,
  author        = {The Verkor Team and Ravi Krishna and Suresh Krishna and David Chin},
  title         = {{Design Conductor 2.0: An agent builds a TurboQuant inference accelerator in 80 hours}},
  year          = {2026},
  eprint        = {2605.05170},
  archivePrefix = {arXiv},
  primaryClass  = {cs.AR},
  doi           = {10.48550/arXiv.2605.05170},
  url           = {https://arxiv.org/abs/2605.05170},
}

@misc{rp260414989,
  author        = {Wenji Fang and Yao Lu and Shang Liu and Jing Wang and Ziyan Guo and Junxian He and Fengbin Tu and Zhiyao Xie},
  title         = {{Dr. RTL: Autonomous Agentic RTL Optimization through Tool-Grounded Self-Improvement}},
  year          = {2026},
  eprint        = {2604.14989},
  archivePrefix = {arXiv},
  primaryClass  = {cs.AI},
  doi           = {10.48550/arXiv.2604.14989},
  url           = {https://arxiv.org/abs/2604.14989},
}

@misc{rp260315672,
  author        = {Kyle Dumont and Nicholas Herbert and Hayder Tirmazi and Shrikanth Upadhayaya},
  title         = {{DRCY: Agentic Hardware Design Reviews}},
  year          = {2026},
  eprint        = {2603.15672},
  archivePrefix = {arXiv},
  primaryClass  = {cs.AR},
  doi           = {10.48550/arXiv.2603.15672},
  url           = {https://arxiv.org/abs/2603.15672},
}

@misc{rp260609867,
  author        = {Qi Xiong and Renzhi Chen and Bowei Wang and Yuqing Xiong and Libo Huang and Lei Wang},
  title         = {{EstRTL: Functional Estimation Guided RTL Code Generation}},
  year          = {2026},
  eprint        = {2606.09867},
  archivePrefix = {arXiv},
  primaryClass  = {cs.AR},
  doi           = {10.48550/arXiv.2606.09867},
  url           = {https://arxiv.org/abs/2606.09867},
}

@misc{rp260415388,
  author        = {Mu-Chi Chen and Po-Hsuan Huang and Yu-Hung Kao and Yen-Fu Liu and Yu-Kai Hung and Cheng Liang and Shao-Chun Ho and Chia-Heng Tu and Shih-Hao Hung},
  title         = {{Exploring LLM-based Verilog Code Generation with Data-Efficient Fine-Tuning and Testbench Automation}},
  year          = {2026},
  eprint        = {2604.15388},
  archivePrefix = {arXiv},
  primaryClass  = {cs.AR},
  doi           = {10.48550/arXiv.2604.15388},
  url           = {https://arxiv.org/abs/2604.15388},
}

@misc{rp260319347,
  author        = {Patrick Yubeaton and Siddharth Garg and Chinmay Hegde},
  title         = {{Exploring the Agentic Frontier of Verilog Code Generation}},
  year          = {2026},
  eprint        = {2603.19347},
  archivePrefix = {arXiv},
  primaryClass  = {cs.AR},
  doi           = {10.48550/arXiv.2603.19347},
  url           = {https://arxiv.org/abs/2603.19347},
}

@misc{rp240203375,
  author        = {Zehua Pei and Hui-Ling Zhen and Mingxuan Yuan and Yu Huang and Bei Yu},
  title         = {{BetterV: Controlled Verilog Generation with Discriminative Guidance}},
  year          = {2024},
  eprint        = {2402.03375},
  archivePrefix = {arXiv},
  primaryClass  = {cs.AI},
  doi           = {10.48550/arXiv.2402.03375},
  url           = {https://arxiv.org/abs/2402.03375},
}

@misc{rp231116543,
  author        = {Yun-Da Tsai and Mingjie Liu and Haoxing Ren},
  title         = {{RTLFixer: Automatically Fixing RTL Syntax Errors with Large Language Models}},
  year          = {2023},
  eprint        = {2311.16543},
  archivePrefix = {arXiv},
  primaryClass  = {cs.AR},
  doi           = {10.48550/arXiv.2311.16543},
  url           = {https://arxiv.org/abs/2311.16543},
}

@misc{rp221211140,
  author        = {Shailja Thakur and Baleegh Ahmad and Zhenxing Fan and Hammond Pearce and Benjamin Tan and Ramesh Karri and Brendan Dolan-Gavitt and Siddharth Garg},
  title         = {{Benchmarking Large Language Models for Automated Verilog RTL Code Generation}},
  year          = {2022},
  eprint        = {2212.11140},
  archivePrefix = {arXiv},
  primaryClass  = {cs.PL},
  doi           = {10.48550/arXiv.2212.11140},
  url           = {https://arxiv.org/abs/2212.11140},
}

@misc{rp260416625,
  author        = {Weihua Du and Jingming Zhuo and Yixin Dong and Andre Wang He and Weiwei Sun and Zeyu Zheng and Manupa Karunaratne and Ivan Fox and Tim Dettmers and Tianqi Chen and Yiming Yang and Sean Welleck},
  title         = {{AdaExplore: Failure-Driven Adaptation and Diversity-Preserving Search for Efficient Kernel Generation}},
  year          = {2026},
  eprint        = {2604.16625},
  archivePrefix = {arXiv},
  primaryClass  = {cs.CL},
  doi           = {10.48550/arXiv.2604.16625},
  url           = {https://arxiv.org/abs/2604.16625},
}

@misc{rp260607665,
  author        = {Xuanzhe Li and Ziyan Weng and Zhiyu Zhu and Junhui Hou},
  title         = {{AgentCompile: An LLM-Guided Compiler for Direct CUDA Inference}},
  year          = {2026},
  eprint        = {2606.07665},
  archivePrefix = {arXiv},
  primaryClass  = {cs.PL},
  doi           = {10.48550/arXiv.2606.07665},
  url           = {https://arxiv.org/abs/2606.07665},
}

@misc{rp260516819,
  author        = {Sharareh Younesian and Wenwen Ouyang and Sina Rafati and Mehdi Rezagholizadeh and Sharon Zhou and Ji Liu and Yue Liu and Yuchen Yang and Hao Li and Ziqiong Liu and Dong Li and Vikram Appia and Zhenyu Gu and Emad Barsoum},
  title         = {{AgentKernelArena: Generalization-Aware Benchmarking of GPU Kernel Optimization Agents}},
  year          = {2026},
  eprint        = {2605.16819},
  archivePrefix = {arXiv},
  primaryClass  = {cs.CL},
  doi           = {10.48550/arXiv.2605.16819},
  url           = {https://arxiv.org/abs/2605.16819},
}

@misc{rp260408590,
  author        = {Brendan R. Hogan and Xiwen Chen and James T. Wilson and Kashif Rasul and Adel Boyarsky and Thomas Kamei and Anderson Schneider and Yuriy Nevmyvaka},
  title         = {{AlphaLab: Autonomous Multi-Agent Research Across Optimization Domains with Frontier LLMs}},
  year          = {2026},
  eprint        = {2604.08590},
  archivePrefix = {arXiv},
  primaryClass  = {cs.LG},
  doi           = {10.48550/arXiv.2604.08590},
  url           = {https://arxiv.org/abs/2604.08590},
}

@misc{rp260418616,
  author        = {Haohui Mai and Xiaoyan Guo and Xiangyun Ding and Daifeng Li and Qiuchu Yu and Chenzhun Guo and Cong Wang and Jiacheng Zhao and Christos Kozyrakis and Binhang Yuan},
  title         = {{Ave: Guiding Agentic GPU Optimization Using Data-Flow Invariants}},
  year          = {2026},
  eprint        = {2604.18616},
  archivePrefix = {arXiv},
  primaryClass  = {cs.DC},
  doi           = {10.48550/arXiv.2604.18616},
  url           = {https://arxiv.org/abs/2604.18616},
}

@misc{rp260122760,
  author        = {Zhongzhen Wen and Shudi Shao and Zhong Li and Yu Ge and Tongtong Xu and Yuanyi Lin and Tian Zhang},
  title         = {{AscendCraft: Automatic Ascend NPU Kernel Generation via DSL-Guided Transcompilation}},
  year          = {2026},
  eprint        = {2601.22760},
  archivePrefix = {arXiv},
  primaryClass  = {cs.DC},
  doi           = {10.48550/arXiv.2601.22760},
  url           = {https://arxiv.org/abs/2601.22760},
}

@misc{rp260107160,
  author        = {Xinzi Cao and Jianyang Zhai and Pengfei Li and Zhiheng Hu and Cen Yan and Bingxu Mu and Guanghuan Fang and Bin She and Jiayu Li and Yihan Su and Dongyang Tao and Xiansong Huang and Fan Xu and Feidiao Yang and Yao Lu and Chang-Dong Wang and Yutong Lu and Weicheng Xue and Bin Zhou and Yonghong Tian},
  title         = {{AscendKernelGen: A Systematic Study of LLM-Based Kernel Generation for Neural Processing Units}},
  year          = {2026},
  eprint        = {2601.07160},
  archivePrefix = {arXiv},
  primaryClass  = {cs.AI},
  doi           = {10.48550/arXiv.2601.07160},
  url           = {https://arxiv.org/abs/2601.07160},
}

@misc{rp260321331,
  author        = {Jaber Jaber and Osama Jaber},
  title         = {{AutoKernel: Autonomous GPU Kernel Optimization via Iterative Agent-Driven Search}},
  year          = {2026},
  eprint        = {2603.21331},
  archivePrefix = {arXiv},
  primaryClass  = {cs.LG},
  doi           = {10.48550/arXiv.2603.21331},
  url           = {https://arxiv.org/abs/2603.21331},
}

@misc{rp260505023,
  author        = {Xing Ma and Yangjie Zhou and Wu Sun and Zihan Liu and Jingwen Leng and Yun Lin and Shixuan Sun and Minyi Guo and Jin Song Dong},
  title         = {{CuBridge: An LLM-Based Framework for Understanding and Reconstructing High-Performance Attention Kernels}},
  year          = {2026},
  eprint        = {2605.05023},
  archivePrefix = {arXiv},
  primaryClass  = {cs.LG},
  doi           = {10.48550/arXiv.2605.05023},
  url           = {https://arxiv.org/abs/2605.05023},
}

@misc{rp260302376,
  author        = {Yoga Sri Varshan Varadharajan and Bodun Hu and Saurabh Agarwal and Aditya Akella},
  title         = {{CUCo: An Agentic Framework for Compute and Communication Co-design}},
  year          = {2026},
  eprint        = {2603.02376},
  archivePrefix = {arXiv},
  primaryClass  = {cs.DC},
  doi           = {10.48550/arXiv.2603.02376},
  url           = {https://arxiv.org/abs/2603.02376},
}

@misc{rp260224286,
  author        = {Weinan Dai and Hanlin Wu and Qiying Yu and Huan-ang Gao and Jiahao Li and Chengquan Jiang and Weiqiang Lou and Yufan Song and Hongli Yu and Jiaze Chen and Wei-Ying Ma and Ya-Qin Zhang and Jingjing Liu and Mingxuan Wang and Xin Liu and Hao Zhou},
  title         = {{CUDA Agent: Large-Scale Agentic RL for High-Performance CUDA Kernel Generation}},
  year          = {2026},
  eprint        = {2602.24286},
  archivePrefix = {arXiv},
  primaryClass  = {cs.LG},
  doi           = {10.48550/arXiv.2602.24286},
  url           = {https://arxiv.org/abs/2602.24286},
}

@misc{rp260508455,
  author        = {Shiyang Li and Haoyang Chen and Mattia Fazzini and Caiwen Ding},
  title         = {{CUDABeaver: Benchmarking LLM-Based Automated CUDA Debugging}},
  year          = {2026},
  eprint        = {2605.08455},
  archivePrefix = {arXiv},
  primaryClass  = {cs.LG},
  doi           = {10.48550/arXiv.2605.08455},
  url           = {https://arxiv.org/abs/2605.08455},
}

@misc{rp260302236,
  author        = {Jiace Zhu and Wentao Chen and Qi Fan and Zhixing Ren and Junying Wu and Xing Zhe Chai and Chotiwit Rungrueangwutthinon and Yehan Ma and An Zou},
  title         = {{CUDABench: Benchmarking LLMs for Text-to-CUDA Generation}},
  year          = {2026},
  eprint        = {2603.02236},
  archivePrefix = {arXiv},
  primaryClass  = {cs.LG},
  doi           = {10.48550/arXiv.2603.02236},
  url           = {https://arxiv.org/abs/2603.02236},
}

@misc{rp260508467,
  author        = {Shiyang Li and Zijian Zhang and Guangyan Sun and Yuebo Luo and Winson Chen and Yanzhi Wang and Mingyi Hong and Caiwen Ding},
  title         = {{CUDAHercules: Benchmarking Hardware-Aware Expert-level CUDA Optimization for LLMs}},
  year          = {2026},
  eprint        = {2605.08467},
  archivePrefix = {arXiv},
  primaryClass  = {cs.LG},
  doi           = {10.48550/arXiv.2605.08467},
  url           = {https://arxiv.org/abs/2605.08467},
}

@misc{rp260401489,
  author        = {Tara Saba and Zhiyang Chen and Jikai Jason Li and Anne Ouyang and Xujie Si and Fan Long},
  title         = {{CuTeGen: An LLM-Based Agentic Framework for Generation and Optimization of High-Performance GPU Kernels using CuTe}},
  year          = {2026},
  eprint        = {2604.01489},
  archivePrefix = {arXiv},
  primaryClass  = {cs.LG},
  doi           = {10.48550/arXiv.2604.01489},
  url           = {https://arxiv.org/abs/2604.01489},
}

@misc{rp260211715,
  author        = {Haolei Bai and Lingcheng Kong and Xueyi Chen and Jianmian Wang and Zhiqiang Tao and Huan Wang},
  title         = {{DICE: Diffusion Large Language Models Excel at Generating CUDA Kernels}},
  year          = {2026},
  eprint        = {2602.11715},
  archivePrefix = {arXiv},
  primaryClass  = {cs.LG},
  doi           = {10.48550/arXiv.2602.11715},
  url           = {https://arxiv.org/abs/2602.11715},
}

@misc{rp260205885,
  author        = {Wei Liu and Jiawei Xu and Yingru Li and Longtao Zheng and Tianjian Li and Qian Liu and Junxian He},
  title         = {{Dr. Kernel: Reinforcement Learning Done Right for Triton Kernel Generations}},
  year          = {2026},
  eprint        = {2602.05885},
  archivePrefix = {arXiv},
  primaryClass  = {cs.LG},
  doi           = {10.48550/arXiv.2602.05885},
  url           = {https://arxiv.org/abs/2602.05885},
}

@misc{rp260426666,
  author        = {Sina Heidari and Dimitrios S. Nikolopoulos},
  title         = {{FACT: Compositional Kernel Synthesis with a Three-Stage Agentic Workflow}},
  year          = {2026},
  eprint        = {2604.26666},
  archivePrefix = {arXiv},
  primaryClass  = {cs.DC},
  doi           = {10.48550/arXiv.2604.26666},
  url           = {https://arxiv.org/abs/2604.26666},
}

@misc{rp260312440,
  author        = {Nina Wiedemann and Quentin Leboutet and Michael Paulitsch and Diana Wofk and Benjamin Ummenhofer},
  title         = {{KernelFoundry: Hardware-aware evolutionary GPU kernel optimization}},
  year          = {2026},
  eprint        = {2603.12440},
  archivePrefix = {arXiv},
  primaryClass  = {cs.DC},
  doi           = {10.48550/arXiv.2603.12440},
  url           = {https://arxiv.org/abs/2603.12440},
}

@misc{rp260311935,
  author        = {Xingze Zou and Jing Wang and Yuhua Zheng and Xueyi Chen and Haolei Bai and Lingcheng Kong and Syed A. R. Abu-Bakar and Zhaode Wang and Chengfei Lv and Haoji Hu and Huan Wang},
  title         = {{MobileKernelBench: Can LLMs Write Efficient Kernels for Mobile Devices?}},
  year          = {2026},
  eprint        = {2603.11935},
  archivePrefix = {arXiv},
  primaryClass  = {cs.LG},
  doi           = {10.48550/arXiv.2603.11935},
  url           = {https://arxiv.org/abs/2603.11935},
}

@misc{rp260315976,
  author        = {Hong Zhang and Barry Smith and Satish Balay and Le Chen and Murat Keceli and Lois Curfman McInnes and Junchao Zhang},
  title         = {{An Agentic Evaluation Framework for AI-Generated Scientific Code in PETSc}},
  year          = {2026},
  eprint        = {2603.15976},
  archivePrefix = {arXiv},
  primaryClass  = {cs.AI},
  doi           = {10.48550/arXiv.2603.15976},
  url           = {https://arxiv.org/abs/2603.15976},
}

@misc{rp260327423,
  author        = {Mahesh Natarajan and Xiaoye Li and Weiqun Zhang},
  title         = {{AstraAI: LLMs, Retrieval, and AST-Guided Assistance for HPC Codebases}},
  year          = {2026},
  eprint        = {2603.27423},
  archivePrefix = {arXiv},
  primaryClass  = {cs.AI},
  doi           = {10.48550/arXiv.2603.27423},
  url           = {https://arxiv.org/abs/2603.27423},
}

@misc{rp260108884,
  author        = {Samyak Jhaveri and Cristina V. Lopes},
  title         = {{Bridging the Gap: Empowering Small Models in Reliable OpenACC-based Parallelization via GEPA-Optimized Prompting}},
  year          = {2026},
  eprint        = {2601.08884},
  archivePrefix = {arXiv},
  primaryClass  = {cs.SE},
  doi           = {10.48550/arXiv.2601.08884},
  url           = {https://arxiv.org/abs/2601.08884},
}

@misc{rp260301051,
  author        = {Mohammad Atif and Kriti Chopra and Fang-Ying Tsai and Ozgur O. Kilic and Tianle Wang and Zhihua Dong and Douglas Benjamin and Charles Leggett and Meifeng Lin and Paolo Calafiura and Salman Habib},
  title         = {{CelloAI Benchmarks: Toward Repeatable Evaluation of AI Assistants}},
  year          = {2026},
  eprint        = {2603.01051},
  archivePrefix = {arXiv},
  primaryClass  = {hep-ex},
  doi           = {10.48550/arXiv.2603.01051},
  url           = {https://arxiv.org/abs/2603.01051},
}

@misc{rp260604023,
  author        = {Jie Li and Wenzhao Wu and Junqi Hu and Qinrui Zheng and Bowen Wu and Juepeng Zheng and Yutong Lu and Haohuan Fu},
  title         = {{CodegenBench: Can LLMs Write Efficient Code Across Architectures?}},
  year          = {2026},
  eprint        = {2606.04023},
  archivePrefix = {arXiv},
  primaryClass  = {cs.SE},
  doi           = {10.48550/arXiv.2606.04023},
  url           = {https://arxiv.org/abs/2606.04023},
}

@misc{rp260408720,
  author        = {Meiziniu Li and Dongze Li and Jianmeng Liu and Shing-Chi Cheung},
  title         = {{Demystifying the Silence of Correctness Bugs in PyTorch Compiler}},
  year          = {2026},
  eprint        = {2604.08720},
  archivePrefix = {arXiv},
  primaryClass  = {cs.SE},
  doi           = {10.48550/arXiv.2604.08720},
  url           = {https://arxiv.org/abs/2604.08720},
}

@misc{rp260222240,
  author        = {Linus Bantel and Moritz Strack and Alexander Strack and Dirk Pflüger},
  title         = {{From Prompts to Performance: Evaluating LLMs for Task-based Parallel Code Generation}},
  year          = {2026},
  eprint        = {2602.22240},
  archivePrefix = {arXiv},
  primaryClass  = {cs.PL},
  doi           = {10.48550/arXiv.2602.22240},
  url           = {https://arxiv.org/abs/2602.22240},
}

@misc{rp260616534,
  author        = {Linus Bantel and Anna-Lena Roth and Jonas Posner and Dirk Pflüger},
  title         = {{Generated, Parallel, Scalable? A Study of Agentic AI-Generated Julia Code on Supercomputers}},
  year          = {2026},
  eprint        = {2606.16534},
  archivePrefix = {arXiv},
  primaryClass  = {cs.DC},
  doi           = {10.48550/arXiv.2606.16534},
  url           = {https://arxiv.org/abs/2606.16534},
}

@misc{rp260212049,
  author        = {Ryo Mikasa and Shun-ichiro Hayashi and Daichi Mukunoki and Tetsuya Hoshino and Takahiro Katagiri},
  title         = {{Improving HPC Code Generation Capability of LLMs via Online Reinforcement Learning with Real-Machine Benchmark Rewards}},
  year          = {2026},
  eprint        = {2602.12049},
  archivePrefix = {arXiv},
  primaryClass  = {cs.LG},
  doi           = {10.48550/arXiv.2602.12049},
  url           = {https://arxiv.org/abs/2602.12049},
}

@misc{rp260422571,
  author        = {William Dawson and Louis Beal and Yoann Curé and Giuseppe Fisicaro and Dorian Rolland and Luigi Genovese},
  title         = {{LARA: Validation-Driven Agentic Supercomputer Workflows for Atomistic Modeling}},
  year          = {2026},
  eprint        = {2604.22571},
  archivePrefix = {arXiv},
  primaryClass  = {physics.comp-ph},
  doi           = {10.48550/arXiv.2604.22571},
  url           = {https://arxiv.org/abs/2604.22571},
}

@misc{rp260408318,
  author        = {Masaki Shiraishi and Ikko Hamamura and Tatsuya Ishigaki and Tadashi Kadowaki},
  title         = {{A Model Context Protocol Server for Quantum Execution in Hybrid Quantum-HPC Environments}},
  year          = {2026},
  eprint        = {2604.08318},
  archivePrefix = {arXiv},
  primaryClass  = {quant-ph},
  doi           = {10.48550/arXiv.2604.08318},
  url           = {https://arxiv.org/abs/2604.08318},
}

@misc{rp260216100,
  author        = {Zijie Su and Muhammed Tawfiqul Islam and Mohammad Goudarzi and Adel N. Toosi},
  title         = {{LLM-Driven Intent-Based Privacy-Aware Orchestration Across the Cloud-Edge Continuum}},
  year          = {2026},
  eprint        = {2602.16100},
  archivePrefix = {arXiv},
  primaryClass  = {cs.DC},
  doi           = {10.48550/arXiv.2602.16100},
  url           = {https://arxiv.org/abs/2602.16100},
}

@misc{rp260203419,
  author        = {Shuang Sun and Huatong Song and Lisheng Huang and Jinhao Jiang and Ran Le and Zhihao Lv and Zongchao Chen and Yiwen Hu and Wenyang Luo and Wayne Xin Zhao and Yang Song and Hongteng Xu and Tao Zhang and Ji-Rong Wen},
  title         = {{SWE-World: Building Software Engineering Agents in Docker-Free Environments}},
  year          = {2026},
  eprint        = {2602.03419},
  archivePrefix = {arXiv},
  primaryClass  = {cs.SE},
  doi           = {10.48550/arXiv.2602.03419},
  url           = {https://arxiv.org/abs/2602.03419},
}

@misc{rp260116238,
  author        = {Bing Xu and Terry Chen and Fengzhe Zhou and Tianqi Chen and Yangqing Jia and Vinod Grover and Haicheng Wu and Wei Liu and Craig Wittenbrink and Wen-mei Hwu and Roger Bringmann and Ming-Yu Liu and Luis Ceze and Michael Lightstone and Humphrey Shi},
  title         = {{VibeTensor: System Software for Deep Learning, Fully Generated by AI Agents}},
  year          = {2026},
  eprint        = {2601.16238},
  archivePrefix = {arXiv},
  primaryClass  = {cs.SE},
  doi           = {10.48550/arXiv.2601.16238},
  url           = {https://arxiv.org/abs/2601.16238},
}

@misc{rp250904191,
  author        = {Omri Sgan Cohen and Ehud Malul and Yair Meidan and Dudu Mimran and Yuval Elovici and Asaf Shabtai},
  title         = {{KubeGuard: LLM-Assisted Kubernetes Hardening via Configuration Files and Runtime Logs Analysis}},
  year          = {2025},
  eprint        = {2509.04191},
  archivePrefix = {arXiv},
  primaryClass  = {cs.CR},
  doi           = {10.48550/arXiv.2509.04191},
  url           = {https://arxiv.org/abs/2509.04191},
}

@misc{rp251105626,
  author        = {Caetano Melone and Daniel Nichols and Konstantinos Parasyris and Todd Gamblin and Harshitha Menon},
  title         = {{LLMs as Packagers of HPC Software}},
  year          = {2025},
  eprint        = {2511.05626},
  archivePrefix = {arXiv},
  primaryClass  = {cs.SE},
  doi           = {10.48550/arXiv.2511.05626},
  url           = {https://arxiv.org/abs/2511.05626},
}

@misc{rp251206915,
  author        = {Kelin Fu and Tianyu Liu and Zeyu Shang and Yingwei Ma and Jian Yang and Jiaheng Liu and Kaigui Bian},
  title         = {{Multi-Docker-Eval: A `Shovel of the Gold Rush' Benchmark on Automatic Environment Building for Software Engineering}},
  year          = {2025},
  eprint        = {2512.06915},
  archivePrefix = {arXiv},
  primaryClass  = {cs.SE},
  doi           = {10.48550/arXiv.2512.06915},
  url           = {https://arxiv.org/abs/2512.06915},
}

@misc{rp260617915,
  author        = {Aueaphum Aueawatthanaphisut and Badri Raj Lamichhane},
  title         = {{Trustworthy Self-Composable Big-Data-as-a-Service: An LLM-Orchestrated Multi-Agent Framework for Automated Data Engineering, AutoML, MLOps Deployment, and Drift-Aware Lifecycle Optimization}},
  year          = {2026},
  eprint        = {2606.17915},
  archivePrefix = {arXiv},
  primaryClass  = {cs.MA},
  doi           = {10.48550/arXiv.2606.17915},
  url           = {https://arxiv.org/abs/2606.17915},
}

@misc{rp250916212,
  author        = {Ahmad Maroof Karimi and Woong Shin and Jesse Hines and Tirthankar Ghosal and Naw Safrin Sattar and Feiyi Wang},
  title         = {{EPIC: Generative AI Platform for Accelerating HPC Operational Data Analytics}},
  year          = {2025},
  eprint        = {2509.16212},
  archivePrefix = {arXiv},
  primaryClass  = {cs.DB},
  doi           = {10.48550/arXiv.2509.16212},
  url           = {https://arxiv.org/abs/2509.16212},
}

@misc{rp250602025,
  author        = {Prachi Jadhav and Hongwei Jin and Ewa Deelman and Prasanna Balaprakash},
  title         = {{Evaluating the Efficacy of LLM-Based Reasoning for Multiobjective HPC Job Scheduling}},
  year          = {2025},
  eprint        = {2506.02025},
  archivePrefix = {arXiv},
  primaryClass  = {cs.DC},
  doi           = {10.48550/arXiv.2506.02025},
  url           = {https://arxiv.org/abs/2506.02025},
}

@misc{rp250902449,
  author        = {Mohsen Seyedkazemi Ardebili and Andrea Bartolini},
  title         = {{KubeIntellect: A Modular LLM-Orchestrated Agent Framework for End-to-End Kubernetes Management}},
  year          = {2025},
  eprint        = {2509.02449},
  archivePrefix = {arXiv},
  primaryClass  = {cs.DC},
  doi           = {10.48550/arXiv.2509.02449},
  url           = {https://arxiv.org/abs/2509.02449},
}

@misc{rp230402014,
  author        = {Yinlin Deng and Chunqiu Steven Xia and Chenyuan Yang and Shizhuo Dylan Zhang and Shujing Yang and Lingming Zhang},
  title         = {{Large Language Models are Edge-Case Fuzzers: Testing Deep Learning Libraries via FuzzGPT}},
  year          = {2023},
  eprint        = {2304.02014},
  archivePrefix = {arXiv},
  primaryClass  = {cs.SE},
  doi           = {10.48550/arXiv.2304.02014},
  url           = {https://arxiv.org/abs/2304.02014},
}

@misc{rp260418964,
  author        = {Ahmed G. A. H Ahmed and C. Okan Sakar},
  title         = {{DW-Bench: Benchmarking LLMs on Data Warehouse Graph Topology Reasoning}},
  year          = {2026},
  eprint        = {2604.18964},
  archivePrefix = {arXiv},
  primaryClass  = {cs.AI},
  doi           = {10.48550/arXiv.2604.18964},
  url           = {https://arxiv.org/abs/2604.18964},
}

@misc{rp260217215,
  author        = {Yi Shan and Yixuan He and Zekai Shao and Kai Xu and Siming Chen},
  title         = {{NotebookRAG: Retrieving Multiple Notebooks to Augment the Generation of EDA Notebooks for Crowd-Wisdom}},
  year          = {2026},
  eprint        = {2602.17215},
  archivePrefix = {arXiv},
  primaryClass  = {cs.HC},
  doi           = {10.48550/arXiv.2602.17215},
  url           = {https://arxiv.org/abs/2602.17215},
}

@misc{rp250606541,
  author        = {Eugenie Lai and Gerardo Vitagliano and Ziyu Zhang and Om Chabra and Sivaprasad Sudhir and Anna Zeng and Anton A. Zabreyko and Chenning Li and Ferdi Kossmann and Jialin Ding and Jun Chen and Markos Markakis and Matthew Russo and Weiyang Wang and Ziniu Wu and Michael J. Cafarella and Lei Cao and Samuel Madden and Tim Kraska},
  title         = {{KramaBench: A Benchmark for AI Systems on Data-to-Insight Pipelines over Data Lakes}},
  year          = {2025},
  eprint        = {2506.06541},
  archivePrefix = {arXiv},
  primaryClass  = {cs.DB},
  doi           = {10.48550/arXiv.2506.06541},
  url           = {https://arxiv.org/abs/2506.06541},
}

@misc{rp250521575,
  author        = {Dawei Feng and Di Mei and Huiri Tan and Lei Ren and Xianying Lou and Zhangxi Tan},
  title         = {{StreamLink: Large-Language-Model Driven Distributed Data Engineering System}},
  year          = {2025},
  eprint        = {2505.21575},
  archivePrefix = {arXiv},
  primaryClass  = {cs.DB},
  doi           = {10.48550/arXiv.2505.21575},
  url           = {https://arxiv.org/abs/2505.21575},
}

@misc{rp260617461,
  author        = {Yiting Wang and Chenhui Deng and Chia-Tung Ho and Yanqing Zhang and Zhuo Feng and Cunxi Yu and Ang Li and Gang Qu and Brucek Khailany},
  title         = {{AUTOGATE: Automated Clock Gating via Toggling-Aware LLM-based RTL Rewriting}},
  year          = {2026},
  eprint        = {2606.17461},
  archivePrefix = {arXiv},
  primaryClass  = {cs.AR},
  doi           = {10.48550/arXiv.2606.17461},
  url           = {https://arxiv.org/abs/2606.17461},
}

@misc{rp260613089,
  author        = {Martin Tomasovic and Lukas Sekanina},
  title         = {{Multi-Objective Coevolution of Prompts and Templates for Circuit Approximation}},
  year          = {2026},
  eprint        = {2606.13089},
  archivePrefix = {arXiv},
  primaryClass  = {cs.NE},
  doi           = {10.48550/arXiv.2606.13089},
  url           = {https://arxiv.org/abs/2606.13089},
}

@article{zhang2025aiops,
  author  = {Lingzhe Zhang and Tong Jia and Mengxi Jia and Yifan Wu and
             Aiwei Liu and Yong Yang and Zhonghai Wu and Xuming Hu and
             Philip S. Yu and Ying Li},
  title = {{A Survey of AIOps in the Era of Large Language Models}},
  journal = {ACM Computing Surveys},
  year    = {2025},
  doi     = {10.1145/3746635},
  note    = {arXiv:2507.12472},
  volume = {58},
  number = {2},
  pages = {1--35},
}

@article{shi2025text2sql,
  author  = {Liang Shi and Zhengju Tang and Nan Zhang and Xiaotong Zhang and Zhi Yang},
  title = {{A Survey on Employing Large Language Models for Text-to-SQL Tasks}},
  journal = {ACM Computing Surveys},
  year    = {2025},
  doi     = {10.1145/3737873},
  note    = {arXiv:2407.15186},
  volume = {58},
  number = {2},
  pages = {1--37},
}

@article{hong2025text2sql,
  author  = {Zijin Hong and Zheng Yuan and Qinggang Zhang and Hao Chen and
             Junnan Dong and Feiran Huang and Xiao Huang},
  title = {{Next-Generation Database Interfaces: A Survey of LLM-Based Text-to-SQL}},
  journal = {IEEE Transactions on Knowledge and Data Engineering},
  year    = {2025},
  doi     = {10.1109/TKDE.2025.3609486},
  note    = {arXiv:2406.08426},
}

@article{yang2026verilog,
  author = {Guang Yang and Wei Zheng and Dong Liang and Peng Hu and Yukui Yang and Shaohua Peng and Zhenghan Li and Jiahui Feng and Xiao Wei and Kexin Sun and Deyuan Ma and Haotian Cheng and Yiheng Shen and Xiang Chen and Xing Hu and Terry Yue Zhuo and David Lo},
  title = {{Large Language Model for Verilog Code Generation: Literature Review and the Road Ahead}},
  journal = {ACM Computing Surveys},
  year    = {2026},
  doi     = {10.1145/3841636},
  note    = {arXiv:2512.00020},
  volume = {58},
  number = {16},
  pages = {1--35},
}

@article{ljaljevic2026hpc,
  author  = {Strahinja Ljaljevi{\'c} and Josep Jorba and Sergio Iserte},
  title = {{Exploring the role of Large Language Models in High-Performance Computing programming: A survey}},
  journal = {Future Generation Computer Systems},
  volume = {184},
  pages = {108618},
  year    = {2026},
  doi     = {10.1016/j.future.2026.108618},
  note    = {arXiv:2608.26110},
}

@article{pahl2026iac,
  author  = {Claus Pahl and {\"O}vg{\"u}m Can Sezen and Florian Hofer},
  title = {{Artificial Intelligence for Infrastructure-as-Code---A Systematic Literature Review}},
  journal = {Electronics},
  volume = {15},
  number = {4},
  pages = {755},
  year    = {2026},
  doi     = {10.3390/electronics15040755},
}

@misc{masthanali2026sustainable,
  author  = {Sabiya Banu Masthan Ali and Oussema Kirmani and Aroosa Hameed and
             Syed Muhammad Danish and Gautam Srivastava},
  title   = {{Energy-Efficient Code Generation Using Large Language Models: A Systematic Literature Review}},
  year    = {2026},
  note    = {arXiv preprint arXiv:2603.00989},
}

@misc{chipnemo2023,
  author        = {Mingjie Liu and Teodor-Dumitru Ene and Robert Kirby and Chris Cheng and
                   Nathaniel Pinckney and Rongjian Liang and Jonah Alben and Himyanshu Anand and
                   Sanmitra Banerjee and Ismet Bayraktaroglu and others},
  title         = {{ChipNeMo}: Domain-Adapted {LLMs} for Chip Design},
  year          = {2023},
  eprint        = {2311.00176},
  archivePrefix = {arXiv},
  primaryClass  = {cs.CL},
  url           = {https://arxiv.org/abs/2311.00176},
}

@inproceedings{amdahl1967validity,
  author = {Gene M. Amdahl},
  title = {{Validity of the single processor approach to achieving large scale computing capabilities}},
  year = {1967},
  doi = {10.1145/1465482.1465560},
  booktitle = {Proceedings of the April 18--20, 1967, Spring Joint Computer Conference},
  pages = {483--485},
  publisher = {ACM},
}

@article{gonzalez1996energy,
  author = {Ricardo Gonzalez and Mark Horowitz},
  title = {{Energy dissipation in general purpose microprocessors}},
  year = {1996},
  doi = {10.1109/4.535411},
  journal = {IEEE Journal of Solid-State Circuits},
  volume = {31},
  number = {9},
  pages = {1277--1284},
}

@book{miettinen1998multiobjective,
  author = {Kaisa Miettinen},
  title = {{Nonlinear Multiobjective Optimization}},
  year = {1998},
  doi = {10.1007/978-1-4615-5563-6},
  publisher = {Springer},
  series = {International Series in Operations Research \& Management Science},
  volume = {12},
}

@article{browne2012mcts,
  author = {Cameron B. Browne and Edward Powley and Daniel Whitehouse and Simon M. Lucas and Peter I. Cowling and Philipp Rohlfshagen and Stephen Tavener and Diego Perez and Spyridon Samothrakis and Simon Colton},
  title = {{A Survey of Monte Carlo Tree Search Methods}},
  year = {2012},
  doi = {10.1109/TCIAIG.2012.2186810},
  journal = {IEEE Transactions on Computational Intelligence and AI in Games},
  volume = {4},
  number = {1},
  pages = {1--43},
}

@misc{zhao2023llmsurvey,
  author = {Zhao, Wayne Xin and Zhou, Kun and Li, Junyi and Tang, Tianyi and Wang, Xiaolei and Hou, Yupeng and Min, Yingqian and Zhang, Beichen and Zhang, Junjie and Dong, Zican and Du, Yifan and Yang, Chen and Chen, Yushuo and Chen, Zhipeng and Jiang, Jinhao and Ren, Ruiyang and Li, Yifan and Tang, Xinyu and Liu, Zikang and Liu, Peiyu and Nie, Jian-Yun and Wen, Ji-Rong},
  title = {{A Survey of Large Language Models}},
  year = {2023},
  eprint = {2303.18223},
  archivePrefix = {arXiv},
  url = {https://arxiv.org/abs/2303.18223},
}

@misc{ouyang2022instructions,
  author = {Ouyang, Long and Wu, Jeff and Jiang, Xu and Almeida, Diogo and Wainwright, Carroll L. and Mishkin, Pamela and Zhang, Chong and Agarwal, Sandhini and Slama, Katarina and Ray, Alex and Schulman, John and Hilton, Jacob and Kelton, Fraser and Miller, Luke and Simens, Maddie and Askell, Amanda and Welinder, Peter and Christiano, Paul and Leike, Jan and Lowe, Ryan},
  title = {{Training language models to follow instructions with human feedback}},
  year = {2022},
  eprint = {2203.02155},
  archivePrefix = {arXiv},
  url = {https://arxiv.org/abs/2203.02155},
}

@misc{lewis2020rag,
  author = {Lewis, Patrick and Perez, Ethan and Piktus, Aleksandra and Petroni, Fabio and Karpukhin, Vladimir and Goyal, Naman and K{\"{u}}ttler, Heinrich and Lewis, Mike and Yih, Wen-tau and Rockt{\"{a}}schel, Tim and Riedel, Sebastian and Kiela, Douwe},
  title = {{Retrieval-Augmented Generation for Knowledge-Intensive NLP Tasks}},
  year = {2020},
  eprint = {2005.11401},
  archivePrefix = {arXiv},
  url = {https://arxiv.org/abs/2005.11401},
}

@misc{wei2022chain,
  author = {Wei, Jason and Wang, Xuezhi and Schuurmans, Dale and Bosma, Maarten and Ichter, Brian and Xia, Fei and Chi, Ed and Le, Quoc and Zhou, Denny},
  title = {{Chain-of-Thought Prompting Elicits Reasoning in Large Language Models}},
  year = {2022},
  eprint = {2201.11903},
  archivePrefix = {arXiv},
  url = {https://arxiv.org/abs/2201.11903},
}

@book{sutton2018reinforcement,
  author = {Richard S. Sutton and Andrew G. Barto},
  title = {Reinforcement Learning: An Introduction},
  edition = {2},
  year = {2018},
  publisher = {MIT Press},
  isbn = {9780262039246},
  url = {https://mitpress.mit.edu/9780262039246/reinforcement-learning/},
}

@article{minaee2024large,
  title={Large language models: A survey},
  author={Minaee, Shervin and Mikolov, Tomas and Nikzad, Narjes and Chenaghlu, Meysam and Socher, Richard and Amatriain, Xavier and Gao, Jianfeng},
  journal={arXiv preprint arXiv:2402.06196},
  year={2024}
}

@misc{ding2025asap,
  author    = {Yuran Ding and
               Xinwei Chen and
               Xiaofan Zhang and
               Zongwei Zhou},
  title     = {{ASAP}: an Agentic Solution to Auto-optimize Performance of Large-Scale {LLM} Training},
  year      = {2025},
  note      = {arXiv preprint arXiv:2511.03844},
}

@misc{raina2026primus,
  author    = {Anshu Raina and
               Peyman Razaghi and
               Yuankai Chen and
               Cheng Yao and
               Yao Fu and
               Vidushi Goyal and
               Devang Patel and
               Wen Xie and
               Zhenyu Gu and
               Emad Barsoum},
  title     = {Beyond the Trial-and-Error Loop: Hybrid Projection and Automated Tuning for Distributed Training},
  year      = {2026},
  note      = {arXiv preprint arXiv:2610.19423},
}

@misc{hamadanian2025glia,
  author    = {Pouya Hamadanian and
               Pantea Karimi and
               Arash Nasr-Esfahany and
               Kimia Noorbakhsh and
               Joseph Chandler and
               Ali ParandehGheibi and
               Mohammad Alizadeh and
               Hari Balakrishnan},
  title     = {{Glia}: A Human-Inspired {AI} for Automated Systems Design and Optimization},
  year      = {2025},
  note      = {arXiv preprint arXiv:2510.27176},
}

@misc{jiang2026autopoiesis,
  author    = {Youhe Jiang and
               Ran Yan and
               You Peng and
               Wenshuang Li and
               Taiyi Wang and
               Fangcheng Fu and
               Binhang Yuan},
  title     = {{Autopoiesis}: A Self-Evolving System Paradigm for {LLM} Serving Under Runtime Dynamics},
  year      = {2026},
  note      = {arXiv preprint arXiv:2604.07144},
}

@misc{poudel2026cachecraft,
  author    = {Pratik Poudel and
               Yanzhao Wu and
               Sumit Jha and
               Jason Liu},
  title     = {Discovering {KV} Cache Eviction Policies via {LLM}-Guided Program Evolution},
  year      = {2026},
  note      = {arXiv preprint arXiv:2608.14555},
}

@misc{cheng2025adrs,
  author    = {Audrey Cheng and
               Shu Liu and
               Melissa Pan and
               Zhifei Li and
               Bowen Wang and
               Alex Krentsel and
               Tian Xia and
               Mert Cemri and
               Jongseok Park and
               Shuo Yang and
               Jeff Chen and
               Lakshya Agrawal and
               Aditya Desai and
               Jiarong Xing and
               Koushik Sen and
               Matei Zaharia and
               Ion Stoica},
  title     = {Barbarians at the Gate: How {AI} is Upending Systems Research},
  year      = {2025},
  note      = {arXiv preprint arXiv:2510.06189},
}

@misc{cheng2025letbarbarians,
  author    = {Audrey Cheng and
               Shu Liu and
               Melissa Pan and
               Zhifei Li and
               Shubham Agarwal and
               Mert Cemri and
               Bowen Wang and
               Alexander Krentsel and
               Tian Xia and
               Jongseok Park and
               Shuo Yang and
               Jeff Chen and
               Lakshya Agrawal and
               Ashwin Naren and
               Shulu Li and
               Ruiying Ma and
               Aditya Desai and
               Jiarong Xing and
               Koushik Sen and
               Matei Zaharia and
               Ion Stoica},
  title     = {Let the Barbarians In: How {AI} Can Accelerate Systems Performance Research},
  year      = {2025},
  note      = {arXiv preprint arXiv:2512.14806},
}

@misc{novikov2025alphaevolve,
  author    = {Alexander Novikov and
               Ng{\^a}n V{\~u} and
               Marvin Eisenberger and
               Emilien Dupont and
               Po-Sen Huang and
               Adam Zsolt Wagner and
               Sergey Shirobokov and
               Borislav Kozlovskii and
               Francisco J. R. Ruiz and
               Abbas Mehrabian and
               M. Pawan Kumar and
               Abigail See and
               Swarat Chaudhuri and
               George Holland and
               Alex Davies and
               Sebastian Nowozin and
               Pushmeet Kohli and
               Matej Balog},
  title     = {{AlphaEvolve}: A coding agent for scientific and algorithmic discovery},
  year      = {2025},
  note      = {arXiv preprint arXiv:2506.13131},
}

@misc{lu2026agenticsched,
  author    = {Tianxi Lu and
               Sherief Reda},
  title     = {Agentic {CPU-GPU} Scheduling for Heterogeneous {AI} Workloads},
  year      = {2026},
  note      = {arXiv preprint arXiv:2607.22242},
}

@misc{yang2026tsguard,
  author    = {Yitao Yang and
               Yangtao Deng and
               Yifan Xiong and
               Baochun Li and
               Hong Xu and
               Peng Cheng},
  title     = {{TSGuard}: Automated User-Centric Incident Diagnosis for {AI} Workloads in the Cloud},
  year      = {2025},
  note      = {arXiv preprint arXiv:2506.01481},
}

@misc{shi2024clusterdiag,
  author    = {Honghao Shi and
               Longkai Cheng and
               Wenli Wu and
               Yuhang Wang and
               Xuan Liu and
               Shaokai Nie and
               Weixv Wang and
               Xuebin Min and
               Chunlei Men and
               Yonghua Lin},
  title     = {Enhancing Cluster Resilience: {LLM}-agent Based Autonomous Intelligent Cluster Diagnosis System and Evaluation Framework},
  year      = {2024},
  note      = {arXiv preprint arXiv:2411.05349},
}

@misc{xing2026flashinferbench,
  author    = {Shanli Xing and
               Yiyan Zhai and
               Alexander Jiang and
               Yixin Dong and
               Yong Wu and
               Zihao Ye and
               Charlie Ruan and
               Yingyi Huang and
               Yineng Zhang and
               Liangsheng Yin and
               Aksara Bayyapu and
               Luis Ceze and
               Tianqi Chen},
  title     = {{FlashInfer-Bench}: Building the Virtuous Cycle for {AI}-driven {LLM} Systems},
  year      = {2026},
  note      = {arXiv preprint arXiv:2601.00227},
}

@misc{nagaitsev2025pike,
  author    = {Kirill Nagaitsev and
               Luka Grbcic and
               Samuel Williams and
               Costin Iancu},
  title     = {Optimizing {PyTorch} Inference with {LLM}-Based Multi-Agent Systems},
  year      = {2025},
  note      = {arXiv preprint arXiv:2511.16964},
}

@misc{tang2026mosaic,
  author    = {Annabelle Sujun Tang and
               Christopher Priebe and
               Lianhui Qin and
               Hadi Esmaeilzadeh},
  title     = {{MOSAIC}: More {LLMs}, Less Cost through Collaborative Reasoning for Compiler Optimizations},
  year      = {2026},
  note      = {arXiv preprint arXiv:2602.01935},
}

@misc{lysenstoen2026sloguard,
  author    = {Christian Lysenst{\o}en},
  title     = {{SLO-Guard}: Crash-Aware, Budget-Consistent Autotuning for {SLO}-Constrained {LLM} Serving},
  year      = {2026},
  note      = {arXiv preprint arXiv:2604.17627},
}

@misc{white2026scepsy,
  author    = {Otto White and
               Marcel Wagenl{\"a}nder and
               Britannio Jarrett and
               Xijin Zhou and
               Yanda Tao and
               Pedro Silvestre and
               Guo Li and
               Huanzhou Zhu and
               Llu{\'i}s Vilanova and
               Peter Pietzuch},
  title     = {{Scepsy}: Serving Agentic Workflows Using Aggregate {LLM} Pipelines},
  year      = {2026},
  note      = {arXiv preprint arXiv:2604.15186},
}

@misc{jiang2024megascale,
  author    = {Ziheng Jiang and
               Haibin Lin and
               Yinmin Zhong and
               Qi Huang and
               Yangrui Chen and
               Zhi Zhang and
               Yanghua Peng and
               Xiang Li and
               Cong Xie and
               Shibiao Nong and
               Yulu Jia and
               Sun He and
               Hongmin Chen and
               Zhihao Bai and
               Qi Hou and
               Shipeng Yan and
               Ding Zhou and
               Yiyao Sheng and
               Zhuo Jiang and
               Haohan Xu and
               Haoran Wei and
               Zhang Zhang and
               Pengfei Nie and
               Leqi Zou and
               Sida Zhao and
               Liang Xiang and
               Zherui Liu and
               Zhe Li and
               Xiaoying Jia and
               Jianxi Ye and
               Xin Jin and
               Xin Liu},
  title     = {{MegaScale}: Scaling Large Language Model Training to More Than 10,000 {GPU}s},
  year      = {2024},
  note      = {arXiv preprint arXiv:2402.15627},
}

@misc{cui2025flare,
  author    = {Weihao Cui and
               Ji Zhang and
               Han Zhao and
               Chao Liu and
               Jian Sha and
               Bingsheng He and
               Minyi Guo and
               Quan Chen},
  title     = {{Flare}: Anomaly Diagnostics for Divergent {LLM} Training in {GPU} Clusters of Thousand-Plus Scale},
  year      = {2025},
  note      = {arXiv preprint arXiv:2502.05413},
}

@misc{jiang2025l4,
  author    = {Zhihan Jiang and
               Junjie Huang and
               Zhuangbin Chen and
               Yichen Li and
               Guangba Yu and
               Cong Feng and
               Yongqiang Yang and
               Zengyin Yang and
               Michael R. Lyu},
  title     = {{L4}: Diagnosing Large-scale {LLM} Training Failures via Automated Log Analysis},
  year      = {2025},
  note      = {arXiv preprint arXiv:2503.20263},
}

@misc{duan2024trainsurvey,
  author    = {Jiangfei Duan and
               Shuo Zhang and
               Zerui Wang and
               Lijuan Jiang and
               Wenwen Qu and
               Qinghao Hu and
               Guoteng Wang and
               Qizhen Weng and
               Hang Yan and
               Xingcheng Zhang and
               Xipeng Qiu and
               Dahua Lin and
               Yonggang Wen and
               Xin Jin and
               Tianwei Zhang and
               Peng Sun},
  title     = {Efficient Training of Large Language Models on Distributed Infrastructures: A Survey},
  year      = {2024},
  note      = {arXiv preprint arXiv:2407.20018},
}

@misc{miao2023servingsurvey,
  author    = {Xupeng Miao and
               Gabriele Oliaro and
               Zhihao Zhang and
               Xinhao Cheng and
               Hongyi Jin and
               Tianqi Chen and
               Zhihao Jia},
  title     = {Towards Efficient Generative Large Language Model Serving: A Survey from Algorithms to Systems},
  year      = {2023},
  note      = {arXiv preprint arXiv:2312.15234},
}

@inproceedings{chen2026optico,
  author    = {Sheng Chen and Zhe Tang and Weixing Zhang and Fei Yang and
               Yuanyuan Wang and Tianlin Li and Yang Liu},
  title     = {{OptiCo}: Adaptive Distributed Training Optimization via
               Collaborative Agent Reasoning},
  booktitle = {Proceedings of the 64th Annual Meeting of the Association for Computational Linguistics (ACL)},
  publisher = {Association for Computational Linguistics},
  pages     = {27847--27867},
  year      = {2026},
}

@misc{kwon2023vllm,
  author        = {Woosuk Kwon and Zhuohan Li and Siyuan Zhuang and Ying Sheng and Lianmin Zheng and Cody Hao Yu and Joseph E. Gonzalez and Hao Zhang and Ion Stoica},
  title         = {{Efficient Memory Management for Large Language Model Serving with PagedAttention}},
  year          = {2023},
  eprint        = {2309.06180},
  archivePrefix = {arXiv},
  url           = {https://arxiv.org/abs/2309.06180}
}

@misc{zheng2024sglang,
  author        = {Lianmin Zheng and Liangsheng Yin and Zhiqiang Xie and Chuyue Sun and Jeff Huang and Cody Hao Yu and Shiyi Cao and Christos Kozyrakis and Ion Stoica and Joseph E. Gonzalez and Clark Barrett and Ying Sheng},
  title         = {{SGLang: Efficient Execution of Structured Language Model Programs}},
  year          = {2023},
  eprint        = {2312.07104},
  archivePrefix = {arXiv},
  url           = {https://arxiv.org/abs/2312.07104}
}

@misc{shoeybi2019megatron,
  author        = {Mohammad Shoeybi and Mostofa Patwary and Raul Puri and Patrick LeGresley and Jared Casper and Bryan Catanzaro},
  title         = {{Megatron-LM: Training Multi-Billion Parameter Language Models Using Model Parallelism}},
  year          = {2019},
  eprint        = {1909.08053},
  archivePrefix = {arXiv},
  url           = {https://arxiv.org/abs/1909.08053}
}

@misc{zhong2024distserve,
  author        = {Yinmin Zhong and Shengyu Liu and Junda Chen and Jianbo Hu and Yibo Zhu and Xuanzhe Liu and Xin Jin and Hao Zhang},
  title         = {{DistServe: Disaggregating Prefill and Decoding for Goodput-optimized Large Language Model Serving}},
  year          = {2024},
  eprint        = {2401.09670},
  archivePrefix = {arXiv},
  url           = {https://arxiv.org/abs/2401.09670}
}

@misc{jiang2024hexgen,
  author        = {Youhe Jiang and Ran Yan and Xiaozhe Yao and Yang Zhou and Beidi Chen and Binhang Yuan},
  title         = {{HexGen: Generative Inference of Large Language Model over Heterogeneous Environment}},
  year          = {2023},
  eprint        = {2311.11514},
  archivePrefix = {arXiv},
  url           = {https://arxiv.org/abs/2311.11514}
}

@misc{miao2024spotserve,
  author        = {Xupeng Miao and Chunan Shi and Jiangfei Duan and Xiaoli Xi and Dahua Lin and Bin Cui and Zhihao Jia},
  title         = {{SpotServe: Serving Generative Large Language Models on Preemptible Instances}},
  year          = {2023},
  eprint        = {2311.15566},
  archivePrefix = {arXiv},
  url           = {https://arxiv.org/abs/2311.15566}
}

@misc{liu2025chunkkv,
  author        = {Xiang Liu and Zhenheng Tang and Peijie Dong and Zeyu Li and Yue Liu and Bo Li and Xuming Hu and Xiaowen Chu},
  title         = {{ChunkKV: Semantic-Preserving KV Cache Compression for Efficient Long-Context LLM Inference}},
  year          = {2025},
  eprint        = {2502.00299},
  archivePrefix = {arXiv},
  url           = {https://arxiv.org/abs/2502.00299}
}

@misc{wang2025openhands,
  author        = {Xingyao Wang and Boxuan Li and Yufan Song and Frank F. Xu and Xiangru Tang and Mingchen Zhuge and Jiayi Pan and Yueqi Song and Bowen Li and Jaskirat Singh and Hoang H. Tran and Fuqiang Li and Ren Ma and Mingzhang Zheng and Bill Qian and Yanjun Shao and Niklas Muennighoff and Yizhe Zhang and Binyuan Hui and Junyang Lin and Robert Brennan and Hao Peng and Heng Ji and Graham Neubig},
  title         = {{OpenHands: An Open Platform for AI Software Developers as Generalist Agents}},
  year          = {2024},
  eprint        = {2407.16741},
  archivePrefix = {arXiv},
  url           = {https://arxiv.org/abs/2407.16741}
}

@misc{mouret2015mapelites,
  author        = {Jean-Baptiste Mouret and Jeff Clune},
  title         = {{Illuminating search spaces by mapping elites}},
  year          = {2015},
  eprint        = {1504.04909},
  archivePrefix = {arXiv},
  url           = {https://arxiv.org/abs/1504.04909}
}

@misc{shao2024deepseekmath,
  author        = {Zhihong Shao and Peiyi Wang and Qihao Zhu and Runxin Xu and Junxiao Song and Xiao Bi and Haowei Zhang and Mingchuan Zhang and Y. K. Li and Y. Wu and Daya Guo},
  title         = {{DeepSeekMath: Pushing the Limits of Mathematical Reasoning in Open Language Models}},
  year          = {2024},
  eprint        = {2402.03300},
  archivePrefix = {arXiv},
  url           = {https://arxiv.org/abs/2402.03300}
}

@misc{yao2023react,
  author        = {Shunyu Yao and Jeffrey Zhao and Dian Yu and Nan Du and Izhak Shafran and Karthik Narasimhan and Yuan Cao},
  title         = {{ReAct: Synergizing Reasoning and Acting in Language Models}},
  year          = {2022},
  eprint        = {2210.03629},
  archivePrefix = {arXiv},
  url           = {https://arxiv.org/abs/2210.03629}
}

@misc{pinckney2025cvdp,
  author        = {Nathaniel Pinckney and Chenhui Deng and Chia-Tung Ho and Yun-Da Tsai and Mingjie Liu and Wenfei Zhou and Brucek Khailany and Haoxing Ren},
  title         = {{Comprehensive Verilog Design Problems: A Next-Generation Benchmark Dataset for Evaluating Large Language Models and Agents on RTL Design and Verification}},
  year          = {2025},
  eprint        = {2506.14074},
  archivePrefix = {arXiv},
  url           = {https://arxiv.org/abs/2506.14074}
}

@misc{hsieh2024ruler,
  author        = {Cheng-Ping Hsieh and Simeng Sun and Samuel Kriman and Shantanu Acharya and Dima Rekesh and Fei Jia and Yang Zhang and Boris Ginsburg},
  title         = {{RULER: What's the Real Context Size of Your Long-Context Language Models?}},
  year          = {2024},
  eprint        = {2404.06654},
  archivePrefix = {arXiv},
  url           = {https://arxiv.org/abs/2404.06654}
}

@misc{xia2024evoeval,
  author        = {Chunqiu Steven Xia and Yinlin Deng and Lingming Zhang},
  title         = {{Top Leaderboard Ranking = Top Coding Proficiency, Always? EvoEval: Evolving Coding Benchmarks via LLM}},
  year          = {2024},
  eprint        = {2403.19114},
  archivePrefix = {arXiv},
  url           = {https://arxiv.org/abs/2403.19114}
}

@misc{agrawal2024vidur,
  author        = {Amey Agrawal and Nitin Kedia and Jayashree Mohan and Ashish Panwar and Nipun Kwatra and Bhargav Gulavani and Ramachandran Ramjee and Alexey Tumanov},
  title         = {{Vidur: A Large-Scale Simulation Framework For LLM Inference}},
  year          = {2024},
  eprint        = {2405.05465},
  archivePrefix = {arXiv},
  url           = {https://arxiv.org/abs/2405.05465}
}

@misc{sallou2024energibridge,
  author        = {June Sallou and Luís Cruz and Thomas Durieux},
  title         = {{EnergiBridge: Empowering Software Sustainability through Cross-Platform Energy Measurement}},
  year          = {2023},
  eprint        = {2312.13897},
  archivePrefix = {arXiv},
  url           = {https://arxiv.org/abs/2312.13897}
}

@misc{wang2026lakeqa,
  author        = {Haonan Wang and Jiaxiang Liu and Yurong Liu and Austin Senna Wijaya and Tianle Zhou and Eden Wu and Yijia Chen and Wanting You and Reya Vir and Daniela Pinto and Grace Fan and Yusen Zhang and Juliana Freire and Eugene Wu},
  title         = {{LakeQA: An Exploratory QA Benchmark over a Million-Scale Data Lake}},
  year          = {2026},
  eprint        = {2606.10460},
  archivePrefix = {arXiv},
  url           = {https://arxiv.org/abs/2606.10460}
}

@misc{zandieh2025turboquant,
  author        = {Amir Zandieh and Majid Daliri and Majid Hadian and Vahab Mirrokni},
  title         = {{TurboQuant: Online Vector Quantization with Near-optimal Distortion Rate}},
  year          = {2025},
  eprint        = {2504.19874},
  archivePrefix = {arXiv},
  url           = {https://arxiv.org/abs/2504.19874}
}

@inproceedings{ansel2024pytorch2,
  author    = {Jason Ansel and others},
  title     = {{PyTorch 2: Faster Machine Learning Through Dynamic Python Bytecode Transformation and Graph Compilation}},
  booktitle = {Proceedings of the 29th ACM International Conference on Architectural Support for Programming Languages and Operating Systems (ASPLOS)},
  year      = {2024},
  doi       = {10.1145/3620665.3640366},
  publisher = {ACM},
  address   = {New York, NY, USA}
}

@inproceedings{rasley2020deepspeed,
  author    = {Jeff Rasley and Samyam Rajbhandari and Olatunji Ruwase and Yuxiong He},
  title     = {{DeepSpeed: System Optimizations Enable Training Deep Learning Models with Over 100 Billion Parameters}},
  booktitle = {Proceedings of the 26th ACM SIGKDD International Conference on Knowledge Discovery and Data Mining (KDD)},
  year      = {2020},
  doi       = {10.1145/3394486.3406703},
  publisher = {ACM},
  address   = {New York, NY, USA}
}

@inproceedings{gamblin2015spack,
  author    = {Todd Gamblin and Matthew LeGendre and Michael R. Collette and Gregory L. Lee and Adam Moody and Bronis R. de Supinski and Scott Futral},
  title     = {{The Spack Package Manager: Bringing Order to HPC Software Chaos}},
  booktitle = {Proceedings of the International Conference for High Performance Computing, Networking, Storage and Analysis (SC)},
  year      = {2015},
  doi       = {10.1145/2807591.2807623},
  publisher = {ACM},
  address   = {New York, NY, USA}
}

@inproceedings{hu2023lucid,
  author    = {Qinghao Hu and Meng Zhang and Peng Sun and Yonggang Wen and Tianwei Zhang},
  title     = {{Lucid: A Non-intrusive, Scalable and Interpretable Scheduler for Deep Learning Training Jobs}},
  booktitle = {Proceedings of the 28th ACM International Conference on Architectural Support for Programming Languages and Operating Systems (ASPLOS)},
  year      = {2023},
  doi       = {10.1145/3575693.3575705},
  publisher = {ACM},
  address   = {New York, NY, USA}
}

@inproceedings{tramm2014xsbench,
  author    = {John R. Tramm and Andrew R. Siegel and Tanzima Islam and Martin Schulz},
  title     = {{XSBench -- The Development and Verification of a Performance Abstraction for Monte Carlo Reactor Analysis}},
  booktitle = {PHYSOR 2014 -- The Role of Reactor Physics toward a Sustainable Future},
  year      = {2014}
}

@misc{pouchet2012polybench,
  author = {Louis-No{\"e}l Pouchet},
  title  = {{PolyBench/C: The Polyhedral Benchmark Suite}},
  year   = {2012},
  url    = {https://sourceforge.net/projects/polybench/}
}

@misc{jin2023hecbench,
  author = {Zheming Jin},
  title  = {{HeCBench: Heterogeneous Computing Benchmarks}},
  year   = {2023},
  url    = {https://github.com/zjin-lcf/HeCBench}
}

@misc{sharma2025openevolve,
  author = {Asankhaya Sharma},
  title  = {{OpenEvolve: An Open-Source Evolutionary Coding Agent}},
  year   = {2025},
  url    = {https://github.com/codelion/openevolve}
}

@misc{perron2024ortools,
  author = {Laurent Perron and Vincent Furnon},
  title  = {{OR-Tools}},
  year   = {2024},
  url    = {https://developers.google.com/optimization/}
}

\clearpage
\appendix
\section{Evidence, Protocols, and Supplementary Analysis}
\label{app:supplement}
\suppressfloats[t]

\subsection{Taxonomy Reference Matrix}
\label{app:taxonomy-table}
Table~\ref{tab:taxonomy} summarizes the assignment rule. Primary categories
organize the presentation; secondary objectives and deployment dependencies
remain attributes of individual studies.

\begin{table}[htbp]
\centering\footnotesize
\caption{Lifecycle-aligned taxonomy of LLM applications to AI infrastructure (eight categories). The principal
contribution and evaluation objective determine the primary category. Secondary
objectives remain attributes of a study rather than evidence that categories are disjoint.}
\label{tab:taxonomy}
\setlength{\tabcolsep}{4pt}
\renewcommand{\arraystretch}{1.15}
\begin{tabularx}{\linewidth}{@{}p{0.45cm}p{2.2cm}p{2.8cm}>{\raggedright\arraybackslash}X@{}}
\toprule
\textbf{ID}&\textbf{Category}&\textbf{Primary concern}&\textbf{Representative outputs and evaluation}\\
\midrule
A1&Hardware Design&Hardware designs and EDA decisions
&RTL/HLS, circuit descriptions, formal properties; simulation, formal checks, and PPA\\
A2&Kernel Engineering&Device-level computation
&Accelerator kernels and transformations; numerical agreement and measured runtime\\
A3&HPC Software&Compiler and parallel software
&Compiler decisions, translated programs, and repairs; code size, execution, and defect analysis\\
\midrule
O1&System Integration&Artifact deployment in a target environment
&Configurations, dependency installation, and service bindings; deployment and integration tests\\
O2&Production Operations&System observations and operational decisions
&Parsed logs, diagnoses, recommendations, and interventions; task-specific assessment and operational outcomes\\
O3&Workload Orchestration&Resource allocation for training and serving workloads
&Parallelization and sharding plans, routing, batching, cache, and placement policies, and schedules; simulators, projection engines, traces, and cluster measurements of throughput, latency, goodput, or job completion time\\
\midrule
R1&Data Management&Data processing, access, and associated resources
&Pipelines, queries, extraction functions, and catalogs; quality, throughput, and resource use\\
R2&Energy Optimization&Energy, power, and thermal objectives
&Refactored code, power-oriented designs, and forecasts; instrumented energy or prediction error\\
\bottomrule
\end{tabularx}
\end{table}

\subsection{Coverage of AI-Specific Concerns}
\label{app:ai-concerns}
Table~\ref{tab:ai-workloads} lists the concerns that distinguish AI
infrastructure from general-purpose computing infrastructure and, for each,
records its lifecycle stage (Figure~\ref{fig:lifecycle}), its taxonomy
category, and the LLM-driven systems that address it. The count $n$
includes only systems in which an LLM acts as the engineering agent; the
coverage mark is a function of $n$ alone, so it summarizes how much LLM-driven
work exists for a concern rather than how good that work is. Concerns with
$n=0$ are the open problems referred to in Section~\ref{sec:prelim:lifecycle}
and Section~\ref{sec:conclusion}.

\begin{table}[t]
\centering
\caption{Coverage of AI-specific training and serving concerns by LLM-driven methods. $n$
counts the LLM-driven systems for each concern; \protect\lcF~$n\geq4$,
\protect\lcH~$1\leq n\leq3$, \protect\lcN~$n=0$. Systems in parentheses address the concern
without an LLM acting as the engineering agent and are not counted.}
\label{tab:ai-workloads}
\footnotesize
\setlength{\tabcolsep}{3.4pt}
\renewcommand{\arraystretch}{1.08}
\begin{tabularx}{\linewidth}{@{}>{\raggedright\arraybackslash}p{4.1cm}c c>{\raggedright\arraybackslash}X cc@{}}
\toprule
\textbf{Concern} & \textbf{Stage} & \textbf{Category} & \textbf{LLM-driven systems (context)} & $\boldsymbol{n}$ & \textbf{Cov.}\\
\midrule
\rowcolor{tabband}\multicolumn{6}{@{}l}{\textit{Distributed training}}\\
Parallelism and sharding plans & S4 & O3 & ASAP~\cite{ding2025asap}, Primus Tuning Agent~\cite{raina2026primus}, OptiCo~\cite{chen2026optico} & 3 & \lcH\\
Compute--communication overlap & S2--S4 & A2 & CUCo~\cite{rp260302376} & 1 & \lcH\\
Training-accelerator circuits & S1 & A1 & AlphaEvolve (TPU arithmetic circuit)~\cite{novikov2025alphaevolve} & 1 & \lcH\\
Failure and straggler diagnosis & S5 & O2 & TSGuard~\cite{yang2026tsguard}, Cluster Diagnosis Agent~\cite{shi2024clusterdiag} (ARGUS~\cite{zhou2026argus}, Flare~\cite{cui2025flare}, L4~\cite{jiang2025l4}) & 2 & \lcH\\
Checkpointing and recovery & S5 & O3 & none & 0 & \lcN\\
Straggler mitigation & S5 & O3 & none (MegaScale~\cite{jiang2024megascale}) & 0 & \lcN\\
\rowcolor{tabband}\multicolumn{6}{@{}l}{\textit{Inference serving}}\\
Routing, batching, and autoscaling & S4 & O3 & Glia~\cite{hamadanian2025glia}, Autopoiesis~\cite{jiang2026autopoiesis} & 2 & \lcH\\
KV-cache eviction & S4 & O3 & CacheCraft~\cite{poudel2026cachecraft} & 1 & \lcH\\
Expert and model placement & S4 & O3 & ADRS~\cite{cheng2025adrs}, Let the Barbarians In~\cite{cheng2025letbarbarians} & 2 & \lcH\\
Quantization and parallelism choice & S4 & O3 & none (AIConfigurator~\cite{xu2026aiconfigurator}, SLO-Guard~\cite{lysenstoen2026sloguard}, Scepsy~\cite{white2026scepsy}) & 0 & \lcN\\
Serving kernels and inference runtimes & S2--S4 & A2 & FlashInfer-Bench~\cite{xing2026flashinferbench}, AgentCompile~\cite{rp260607665}, PIKE~\cite{nagaitsev2025pike}, Ave~\cite{rp260418616}, CuBridge~\cite{rp260505023}, Semantic+Search GPU Tuner~\cite{wang2026fact} & 6 & \lcF\\
Inference accelerators & S1 & A1 & Design Conductor 2.0~\cite{rp260505170} & 1 & \lcH\\
\rowcolor{tabband}\multicolumn{6}{@{}l}{\textit{Shared by training and serving}}\\
Deep-learning operator kernels & S2 & A2 & other AI-evaluated A2 systems (Table~\ref{tab:cmp-a2}) & 22 & \lcF\\
DL compilers, frameworks, and testing & S3 & A3 & PassNet~\cite{chen2026passnet}, MOSAIC~\cite{tang2026mosaic}, AlignGuard~\cite{rp260408720}, FuzzGPT~\cite{rp230402014}, VibeTensor~\cite{rp260116238} & 5 & \lcF\\
GPU-cluster scheduling and placement & S5 & O3 & SchedMate~\cite{wang2025semantic}, AlphaEvolve (Borg)~\cite{novikov2025alphaevolve}, Agentic CPU-GPU Scheduling~\cite{lu2026agenticsched} & 3 & \lcH\\
Training-data preparation & S6 & R1 & DataFlow~\cite{dataflow2025prep} & 1 & \lcH\\
Data loading and input pipelines & S6 & R1 & none & 0 & \lcN\\
Accelerator power and facility cooling & S7 & R2 & AUTOGATE (NVDLA)~\cite{rp260617461}, HyperLoad~\cite{jiang2025hyperload} & 2 & \lcH\\
Power capping for AI jobs & S7 & R2 & none & 0 & \lcN\\
\bottomrule
\end{tabularx}
\end{table}

\subsection{Literature Collection Protocol}
\label{app:protocol}
\noindent\textbf{Candidate pool.}
The review starts from an arXiv candidate pool on LLM-driven infrastructure
engineering, published between 2022 and 2026, with title, authors, year, and a
provisional category for each record. Authors, titles, and years were
re-fetched from the arXiv API for every record that enters the review, and
the provisional categories were reassigned under the rules of
Section~\ref{sec:taxonomy:rule}.

\noindent\textbf{Targeted supplement.}
Because the pool under-represented the runtime decisions of training and
serving, it was supplemented in this revision by targeted web and arXiv
searches combining an LLM or agent term with each AI-specific concern of
Table~\ref{tab:ai-workloads}: parallelism or sharding configuration and
Megatron~\cite{shoeybi2019megatron} or DeepSpeed~\cite{rasley2020deepspeed} tuning; collective communication; failure, straggler,
and hang diagnosis in training clusters; serving configuration, batching,
routing, autoscaling, and SLO tuning; KV-cache eviction; mixture-of-experts
expert placement; GPU-cluster job scheduling; training-data loading; and
AI-driven systems research with OpenEvolve~\cite{sharma2025openevolve} or AlphaEvolve~\cite{novikov2025alphaevolve}. Each search was
followed by backward snowballing from the retrieved papers' related work and
by verification of the arXiv or proceedings record. Searches for checkpointing, straggler mitigation,
data loading, quantization choice, and power capping returned no system in
which an LLM acts as the engineering agent.

\noindent\textbf{Inclusion and selection.}
A system is eligible when an LLM generates an infrastructure artifact,
proposes an infrastructure action or policy, or interprets system evidence,
and when a primary paper describes its evaluation. Systems that optimize LLM
execution without an LLM agent are retained only as context and are labeled as
such. From the eligible records, systems were selected for detailed review to
cover every method family, benchmark, and AI-specific concern of each
category, preferring primary studies with recoverable evaluation settings. The
selected systems' attributes are recorded in a machine-readable table
from which Figure~\ref{fig:taxonomy-framework} and
Tables~\ref{tab:cmp-a1}--\ref{tab:cmp-r2} are generated.

\noindent\textbf{Scope of the claims.}
The selection is documented and covers each category's method families, but
it is not a pre-registered systematic review, and eligible papers outside the
selection are not individually reviewed. The quantitative synthesis in Section~\ref{sec:experiments} applies the additional
extraction rules below.

\subsection{Extraction Rules and Cross-Setting Interpretation}
\label{app:tiers}

\noindent\textbf{Provenance and arithmetic.}
The evidence archive records the primary PDF, version where identifiable,
source location, and file hash. Numerical observations are transcribed from
explicit tables, prose, or printed figure labels. Counts, rate differences, ratios, and endpoint
changes are distinguished from quoted source values. Unless underlying counts
are used explicitly, differences are calculated from the displayed values;
small rounding differences are not evidence of a different experiment.
The synthesis verifies what the sources report, not whether an independent
re-execution would reproduce their results.

\noindent\textbf{Comparison levels.}
A within-study contrast retains the study's common task definition and records
the components that change. It does not imply randomized causal identification.
Cross-setting comparisons are permitted when the native metric and differing
conditions remain explicit. Different baselines, devices, task subsets, or
budgets qualify interpretation rather than prohibit juxtaposition. Outcomes
with different meanings may be presented together as a descriptive inventory,
but are not averaged into a pooled score or used to infer relative task
difficulty. The protocol descriptions below apply to the corresponding main-text
figures and tables.

\noindent\textbf{A1: rates, revisions, and budgets.}
Original RTLLM~\cite{lu2023rtllm} evaluates 30 designs with five generations per design. Its
syntax percentage is generation-based, while functionality counts designs
with a passing candidate (Table~III). A syntax-minus-function difference is
therefore not a nested failure fraction. RTLCoder~\cite{liu2023rtlcoder} uses the 29-design RTLLM-v1.1~\cite{lu2023rtllm} set and reports
pass@5; its VerilogEval~\cite{liu2023verilogeval} table combines new evaluations and values quoted from
prior publications, including the GPT-3.5/GPT-4 VerilogEval rows. VeriPPA~\cite{thorat2025verippa} uses 29 designs and the RTLLM columns of Table~II; its four-shot
prompt describes examples in context, not four generated candidates.
EvolVE's Table~1 uses VerilogEval-v2~\cite{pinckney2024revisiting,hsin2026evolve} and RTLLM-v2 and reports computational
node budgets of 15--300. Node counts and model calls need not have equal cost.
The syntax and functional coordinates in Figure~\ref{fig:a1-plane} retain
these source definitions. Its contours are descriptive differences between
axes, not estimates of a common failure probability.

\noindent\textbf{A2: threshold, selection, and execution reference.}
KernelBench's $\mathrm{fast}_1$ uses a strict speedup threshold of one~\cite{kernelbench2025}; the
candidate budget is separately denoted by $k$ in $\mathrm{fast}_p@k$.
The refinement comparison uses NVIDIA L40S and PyTorch eager. Levels~1 and~2
contain 100 tasks each; Level~3 contains 50. The GPT-5-RL~\cite{tehrani2026gpt5} study uses an
expanded 264-task suite and TorchInductor~\cite{ansel2024pytorch2}. Its Figure~5 speedups summarize
model-specific sets of correct kernels; the source separately compares their
intersection in Figure~7. The Figure~5 means do not share a fixed solved-task
denominator. Its one-attempt condition, three
attempts with feedback, and MakoraGenerate~\cite{tehrani2026gpt5} agent are distinct protocols.
CUDA-L1's main A100 results use the source's Default reference and a speedup
threshold greater than 1.01. Its limit of 20 trials occurs in training-data
collection and is not assigned here as the evaluation budget~\cite{li2026cudal1}. Moreover,
CUDA-L1 optimizes over the benchmark tasks during training; its benchmark
results should not be interpreted as held-out task generalization. PEAK~\cite{tariq2025peak} uses
custom MatMul workloads on A6000, MI200, and Adreno hardware, with
human-selected transformation sequences. Its arithmetic mean across device,
precision, and matrix-size results differs from a geometric mean over tasks.

\noindent\textbf{A3: units and validation strength.}
LLM Compiler's 3.01\% concerns instruction counts on held-out functions~\cite{cummins2023compiler},
not runtime on the LLVM test suite. Zero additional compilations refers to
pass selection at inference, not zero training or compilation cost.
Magellan's 5.23\% and 8.79\% concern different compiler-size evaluations~\cite{chen2026magellan}.
AwareCompiler's 30.03\% is the mean of suite-level IR-count reductions in
Table~1, not a 15--25\% speedup over \texttt{-O3}~\cite{lin2025awarecompiler}. UniPar validates translated
kernels on ParaTrans~\cite{bitan2025unipar}; ACCeLLiuM's exact-match and directive-type rates measure
textual agreement~\cite{jhaveri2025accellium}. MPI detection and repair use different denominators. The repair source prints
76.21\%, whereas its reported 125/164 count rounds to 76.22\%; the latter is
used for the derived rate in this survey. CompilerGPT's maximum speedup is one result among repeated trials on five
programs~\cite{pirkelbauer2025compilergpt}. These metrics are retained individually in Figure~\ref{fig:a3-cost}.

\noindent\textbf{O1--O2: verifier and judge.}
DeployBench~\cite{wang2026deploybench} uses fresh virtual machines, the OpenHands~\cite{wang2025openhands} scaffold, hidden
verification, and task-specific time limits of 10--60 minutes. Five models
are included in the archived source. GenKubeSec's detection results are not
proof that all proposed remediations are safe or deploy successfully~\cite{malul2024genkubesec}.
TraceBench~\cite{egersdoerfer2026ioagent} compares four diagnosis tools with a GPT-4o ranker. Scores are
obtained from rank transformations and aggregation across traces; the axis
named accuracy is not ordinary root-cause classification accuracy.
Drishti~\cite{drishti} is rule-based and ION~\cite{ion} is LLM-based. HPC-LLM's BERTScore of 0.808
and 0.831 in two runs is an answer-similarity metric~\cite{hpcllm2026}, not ticket-resolution
accuracy. The Kubernetes migration study~\cite{ueno2024migrating} and the tracing, diagnosis, and
log-parsing systems ARGUS, MetaKube, StateGraph, and the HPC log
parser~\cite{zhou2026argus,sun2026metakube,xiang2025stategraph,karimi2026logs} are discussed
qualitatively and contribute no TraceBench measurements. AIConfigurator~\cite{xu2026aiconfigurator} is
retained only as contextual non-LLM optimization; CPU search time and aggregate
GPU-hours are reported separately, without presenting their quotient as a
matched wall-clock acceleration.

\noindent\textbf{O3: evaluators, references, and derived factors.}
Factors in Figure~\ref{fig:o3-evidence} and Table~\ref{tab:o3-evidence} are
derived from source values: reference divided by method for time and cost,
method divided by reference for throughput, MFU, and accuracy scores, and
$1/(1-r)$ for a reported fractional reduction $r$. ASAP's evaluation is
agreement with an expert sharding mesh~\cite{ding2025asap}; the 1.43$\times$ and 2.58$\times$
throughput gains it reports belong to the matched expert configurations, and
the latter includes a larger batch size. Primus~\cite{raina2026primus} compares an FP8 configuration
found by the agent with a published BF16 reference, so its 27\% measured gain
combines configuration and precision; its ablation (Table~11 of the source)
uses projected throughput on a different four-node cluster. OptiCo's 1.84\%
is the mean relative MFU improvement over the best expert-tuned configuration across
16 model--scale settings~\cite{chen2026optico}; the reported range is 0.08--8.65\%. Glia's 2.2$\times$
is mean request completion time in the vidur~\cite{agrawal2024vidur,hamadanian2025glia} simulator with four A10 GPUs and
Llama-3-8B, and its testbed result uses slowdown, a different metric.
Autopoiesis~\cite{jiang2026autopoiesis} reports mean and maximum reductions in total trace time, which
includes scheduling, reconfiguration, and serving time. CacheCraft's factor
uses the RULER~\cite{hsieh2024ruler} 4k score at 88\% compression on Llama-3.1-8B~\cite{poudel2026cachecraft}, where the
strongest baseline is ChunkKV~\cite{liu2025chunkkv}. ADRS's expert-placement factor is computed from
the reported 19.6\,ms and 3.7\,ms runtimes at equal imbalance~\cite{cheng2025adrs}; the source's
case-study text reports 5.0$\times$ and its summary table 2$\times$.
SchedMate's physical result uses an eight-node A800 cluster and Lucid~\cite{wang2025semantic,hu2023lucid}; its simulated
range spans four production traces. AlphaEvolve's 0.7\% is fleet-wide compute
recovered after deployment~\cite{novikov2025alphaevolve}, measured by the source rather than by a controlled
comparison released with the paper. Search times are wall-clock times stated
by the sources and exclude the construction of their evaluators.

\noindent\textbf{R1: privileged information and accounting boundaries.}
SANA's Naive, Standard, and Ideal conditions jointly vary planning, search,
and analysis~\cite{sana2026datalake}. Ideal has sanitized subquestions, gold-source retrieval, and
idealized analysis that may call a stronger model. A 30-tool-call limit and
600-second time limit do not equalize information access or total computation.
The Ideal condition is diagnostic rather than an attainable upper-bound
proof. Evaporate~\cite{arora2023evaporate} evaluates Direct on sampled documents (20\% for FDA and
Wikipedia, 2\% for SWDE, and 0.0002\% for Enron), while Code+ processes the
full datasets. These coverage differences accompany the reported quality
comparison. Its approximately $110\times$ token reduction averages per-source
ratios; dividing the rounded mean token counts, $258/1.9$, instead gives
approximately $136\times$. Monetary values retain the original model and historical prices and are not
current API-cost estimates. SANA's $\Delta$ values in Table~\ref{tab:r1-quality-cost}
are those printed by the source, computed from unrounded rates.

\noindent\textbf{R2: screened candidates, meters, and forecasting.}
LASSI-EE~\cite{dearing2026lassiee} reports ER@$k$ using its screening and energy-measurement procedure;
the screening includes an LLM-as-judge verdict. A VALID label does not provide
a formal proof of semantic equivalence. The tabulated averages span 22
HeCBench~\cite{jin2023hecbench} applications on MI100 and A100. The study of Apsan et al.~\cite{apsan2025energy} on EvoEval~\cite{xia2024evoeval} problems uses the
Functional prompting condition as the LLM reference. It uses EnergiBridge~\cite{sallou2024energibridge}
on the server and PC and a Monsoon monitor on Raspberry Pi, with repeated
measurements. The small Pi mean difference does not establish statistical
superiority; the source reports a nonsignificant aggregate comparison there.
HyperLoad~\cite{jiang2025hyperload} reference errors are selected independently as the minimum
non-HyperLoad value for each metric, horizon, and training fraction. Its MSE
and MAE describe cooling-load forecasts on DCData, not measured
energy saved by a controller. No common percentage-efficiency scale is imposed
on these studies.

\subsection{Within-Study Category Figures}
\label{app:single-study-figs}
\begin{figure}[tbp]
\centering
\newlength{\operationfigureheight}
\setlength{\operationfigureheight}{0.45\linewidth}
\begin{minipage}[t]{0.485\linewidth}
\centering
\begin{minipage}[c][\operationfigureheight][c]{\linewidth}
\centering
\includegraphics[width=\linewidth,height=\operationfigureheight,keepaspectratio]{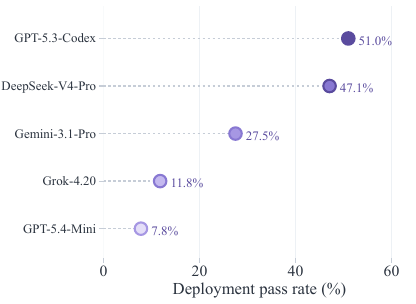}
\end{minipage}
\caption{Deployment success on DeployBench~\cite{wang2026deploybench}.
Pass rate is the percentage of tasks completed successfully under the hidden
verifier; higher is better.}
\Description{Five agents have pass rates of 51.0, 47.1, 27.5, 11.8 and 7.8 percent.}
\label{fig:o1-integration}
\end{minipage}\hfill
\begin{minipage}[t]{0.485\linewidth}
\centering
\begin{minipage}[c][\operationfigureheight][c]{\linewidth}
\centering
\includegraphics[width=\linewidth,height=\operationfigureheight,keepaspectratio]{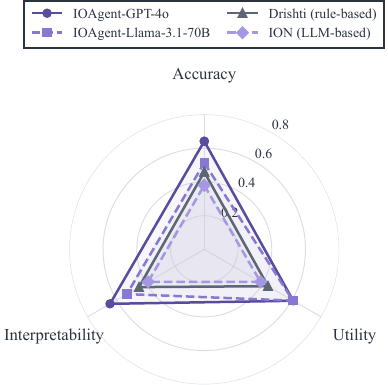}
\end{minipage}
\caption{Normalized diagnosis rankings on TraceBench~\cite{egersdoerfer2026ioagent}.
The GPT-4o judge assesses accuracy, utility, and interpretability. Axes display
$[0,0.8]$ of the underlying $[0,1]$ range; higher is better.}
\Description{A radar plot compares systems on judge-assigned ranking criteria.}
\label{fig:o2-diagnosis}
\end{minipage}
\end{figure}

\begin{figure}[tp]
\centering
\begin{subfigure}[t]{0.48656\textwidth}
\centering
\includegraphics[width=\linewidth]{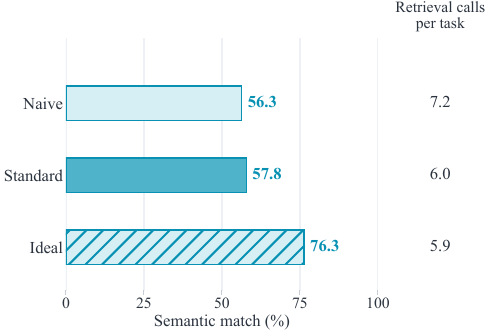}
\caption{SANA~\cite{sana2026datalake} configurations on LakeQA~\cite{wang2026lakeqa}.}
\label{fig:r1-sana}
\end{subfigure}\hfill
\begin{subfigure}[t]{0.47344\textwidth}
\centering
\includegraphics[width=\linewidth]{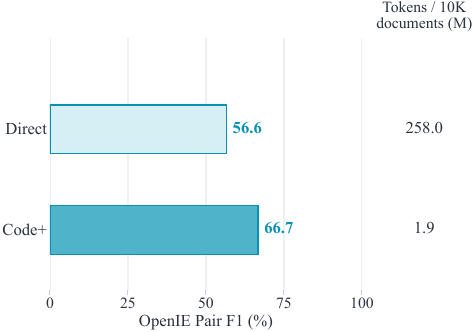}
\caption{Evaporate~\cite{arora2023evaporate} extraction methods.}
\label{fig:r1-evaporate}
\end{subfigure}
\caption{Quality and resource use under alternative data-management configurations. SANA's Ideal
condition uses oracle-like tools to diagnose headroom~\cite{sana2026datalake}; Evaporate
compares direct extraction with synthesized code~\cite{arora2023evaporate}. Semantic
match and OpenIE Pair F1 are percentages, with higher values indicating better quality.
Retrieval calls and token consumption are separate resource measures, with lower values
indicating less use. Compare configurations within each subplot, not across studies.}
\Description{Three SANA bars show semantic match with retrieval-call counts; two Evaporate bars show OpenIE Pair F1 with token consumption.}
\label{fig:r1-quality-cost}
\end{figure}

Figures~\ref{fig:o1-integration} and~\ref{fig:o2-diagnosis} display the
DeployBench~\cite{wang2026deploybench} and TraceBench~\cite{egersdoerfer2026ioagent} values. Figure~\ref{fig:r1-quality-cost} retains
separate data-management metrics. Their protocol differences are described
above; no additional normalization is applied to make the tasks interchangeable.

\subsection{Source Locations and Native Outcomes}
\label{app:ledgers}
Table~\ref{tab:source-locations} identifies the principal numerical sources.
Locations refer to the archived PDF versions. A table entry may quote results
from an earlier paper; such secondary reporting is identified in the main text.
The accompanying local evidence manifest records exact filenames and hashes.
Table~\ref{tab:o3-evidence} lists the workload-orchestration results summarized
in Figure~\ref{fig:o3-evidence} together with their evaluation environments,
references, and search costs, and Table~\ref{tab:a3-gaincost} lists the
HPC-software outcomes shown in Figure~\ref{fig:a3-cost}.

\begin{table}[tp]
\centering\footnotesize
\caption{Primary-source locations for the quantitative synthesis. Values are transcribed
from the identified tables or explicit prose; differences and ratios are survey-derived
unless the source reports them directly. Page numbers refer to PDF pages.}
\label{tab:source-locations}\label{tab:rtl-passrates}\label{tab:kernelbench-fast1}
\begin{tabularx}{\linewidth}{@{}p{3.5cm}p{3.0cm}X@{}}
\toprule
\textbf{Source}&\textbf{Location}&\textbf{Evidence used}\\
\midrule
VerilogEval~\cite{liu2023verilogeval}&Table on p.\ 6&Machine/Human pass@$k$ for GPT models and Verilog-SFT\\
RTLLM~\cite{lu2023rtllm}&Table III, p.\ 5&Syntax rates; functional counts out of 30\\
RTLCoder~\cite{liu2023rtlcoder}&Table II, p.\ 4&Model comparison, quoted prior results, and fine-tuning contrasts\\
VeriPPA~\cite{thorat2025verippa}&Table II, p.\ 5&66.89/81.37 syntax; 44.82/62.06 functionality\\
EvolVE~\cite{hsin2026evolve}&Table 1, p.\ 8&Pass rates by model, search strategy, and node budget\\
ChipGPT~\cite{chang2023chipgpt}&Evaluation, p.\ 7&Reported 35\% area reduction; contextual A1 evidence\\
KernelBench~\cite{kernelbench2025}&Tables 1--2&One-call baseline and ten-call sampling/refinement\\
GPT-5-RL~\cite{tehrani2026gpt5}&Figure 5, p.\ 16; p.\ 5&264-task correctness, performance success, speedup, three-attempt result\\
CUDA-L1~\cite{li2026cudal1}&Tables 4--5&Correct/faster task counts and arithmetic mean speedups\\
KernelEvolve~\cite{meta2025kernelevolve}&Abstract; evaluation&250/250 KernelBench functional successes\\
PEAK~\cite{tariq2025peak}&Table 1, p.\ 4&19.52 mean speedup; 78.26\% mean documented-peak fraction\\
DeployBench~\cite{wang2026deploybench}&Table 2&Five models on 51 tasks\\
GenKubeSec~\cite{malul2024genkubesec}&Detection table, p.\ 7&Precision 0.990; recall 0.999; F1 0.994\\
IOAgent~\cite{egersdoerfer2026ioagent}&Table IV&TraceBench scores for IOAgent, Drishti, and ION\\
ASAP~\cite{ding2025asap}&Section 3&Agreement with expert meshes in three TPU scenarios\\
Primus~\cite{raina2026primus}&Section 4.7; Tables 8, 9, 11&3,475 and 4,402 tok/s/GPU; 30 trials; ablation of oracle and agent\\
OptiCo~\cite{chen2026optico}&Tables 1--2; Section 3.5&MFU against expert configurations; time to best; cost-benefit estimate\\
Glia~\cite{hamadanian2025glia}&Sections 1, 4.4, 5.1--5.3&Completion time vs.\ LLQ and OpenEvolve; testbed slowdown; GPU-hours\\
Autopoiesis~\cite{jiang2026autopoiesis}&Section 7.1&Mean and maximum trace-time reductions vs.\ serving systems\\
CacheCraft~\cite{poudel2026cachecraft}&Section 6; scope paragraph&RULER scores at 88\% compression; LongBench-v2 counterexample\\
ADRS~\cite{cheng2025adrs}&Table 1; Section 5.2&Expert-placement runtimes; placement cost; search time and cost\\
SchedMate~\cite{wang2025semantic}&Section 5.2 (Table 1); Section 5.3&Physical and simulated JCT; simulator error\\
AlphaEvolve~\cite{novikov2025alphaevolve}&Sections 3.3.1--3.3.4&Fleet compute recovered; Gemini kernel and training time; TPU circuit\\
TSGuard~\cite{yang2026tsguard}&Abstract; evaluation&Diagnostic accuracy and verification time on production incidents\\
SANA~\cite{sana2026datalake}&Tables 3--4 and 7--8&LakeQA/KramaBench quality and tool calls\\
Evaporate~\cite{arora2023evaporate}&Table 3, p.\ 11&Five-source OpenIE Pair F1, tokens, and historical cost\\
LASSI-EE~\cite{dearing2026lassiee}&Table III, p.\ 9&ER@$k$ values and differences from vanilla LLM\\
Apsan et al.~\cite{apsan2025energy}&Table 4, p.\ 7&Device-specific canonical and functional-prompt means\\
HyperLoad~\cite{jiang2025hyperload}&Table 1&DCData errors across horizons and training fractions\\
\bottomrule
\end{tabularx}
\end{table}

\begin{table}[tp]
\centering
\caption{Workload-orchestration results under native evaluation conditions. Factors are derived
from the reported values as in Figure~\ref{fig:o3-evidence}; a dash means that the study reports
agreement or optimality rather than a ratio, and n/r means not reported. Glia's simulator runs
Llama-3-8B on four A10 GPUs; CacheCraft's result is at 88\% KV-cache compression.
$^{\dagger}$ADRS reports this speedup as 5.0$\times$ in its case-study text and as 2$\times$ in its
summary table; the factor shown is computed from the reported runtimes.}
\label{tab:o3-evidence}
\scriptsize
\setlength{\tabcolsep}{3pt}
\renewcommand{\arraystretch}{1.1}
\begin{tabularx}{\linewidth}{@{}>{\raggedright\arraybackslash}p{1.75cm}>{\raggedright\arraybackslash}p{1.9cm}>{\raggedright\arraybackslash}p{2.0cm}>{\raggedright\arraybackslash}p{2.0cm}>{\raggedright\arraybackslash}X r>{\raggedright\arraybackslash}p{1.75cm}@{}}
\toprule
\textbf{System} & \textbf{Decision} & \textbf{Environment} & \textbf{Reference} & \textbf{Reported result} & \textbf{Factor} & \textbf{Search cost}\\
\midrule
\rowcolor{tabband}\multicolumn{7}{@{}l}{\textit{Parallelism and sharding configuration}}\\
ASAP~\cite{ding2025asap} & TPU sharding mesh & three TPU jobs & expert mesh & matches the expert in 3/3 & --- & n/r\\
Primus~\cite{raina2026primus} & parallelism, recompute, kernels & projection; 32 MI355X GPUs & published BF16 recipe & 3{,}475$\to$4{,}402 tok/s/GPU & 1.27$\times$ & 30 trials, 30 min\\
OptiCo~\cite{chen2026optico} & Megatron-LM strategy & 8 or 16 A100 GPUs & best expert configuration & MFU $+$1.84\% (0.08--8.65\%) & 1.02$\times$ & 0.05--3.57 h\\
\midrule
\rowcolor{tabband}\multicolumn{7}{@{}l}{\textit{Serving and memory policies}}\\
Glia~\cite{hamadanian2025glia} & request router & simulator; 4$\times$A10 testbed & LLQ & completion time 2.2$\times$ lower; slowdown $>$4.5$\times$ lower & 2.2$\times$ & 20 runs, $<$2 h\\
Autopoiesis~\cite{jiang2026autopoiesis} & (re)scheduling policy & 32--64-GPU clusters & DistServe, HexGen, SpotServe & trace time $-$27\%, $-$30\%, $-$44\% & 1.37--1.79$\times$ & online\\
CacheCraft~\cite{poudel2026cachecraft} & KV-cache eviction & RULER 4k, Llama-3.1-8B & ChunkKV & score 69.2$\to$84.6 & 1.22$\times$ & n/r\\
ADRS~\cite{cheng2025adrs} & expert rebalancing & MoE simulator & internal reference & 19.6$\to$3.7 ms & 5.3$\times^{\dagger}$ & 5 h, $<$\$10\\
ADRS~\cite{cheng2025adrs} & model placement & simulator & published solution & cost $-$18.5\% & 1.23$\times$ & 40 min, $\leq$\$5\\
\midrule
\rowcolor{tabband}\multicolumn{7}{@{}l}{\textit{Cluster scheduling and placement}}\\
SchedMate~\cite{wang2025semantic} & scheduler advice & 8-node A800 cluster; trace simulation & Lucid & JCT 6.85$\to$5.53 h; 1.23--1.91$\times$ in simulation & 1.24$\times$ & 9 s per advice\\
AlphaEvolve~\cite{novikov2025alphaevolve} & Borg priority heuristic & production fleet & production heuristic & 0.7\% of fleet compute recovered & 1.007$\times$ & n/r\\
Agentic scheduler~\cite{lu2026agenticsched} & tool-to-device mapping & RTX A5500, 13 scenarios & HEFT, StarPU, brute force & optimal in 13/13 & --- & n/r\\
Incisor~\cite{laurenzano2026incisor} & instance selection & AWS EC2 & expert constraints with SkyPilot & runtime $-$54\%, cost $-$44\% & 2.17$\times$ & n/r\\
\bottomrule
\end{tabularx}
\end{table}

\begin{table}[tp]
\centering\footnotesize
\caption{HPC-software outcomes and source locations. Measures retain their native units and
evaluation populations; UniPar uses its fine-tuned Llama model and ACCeLLiuM its fine-tuned
CodeLlama model.}
\label{tab:a3-gaincost}\label{tab:hpc-ledger}
\begin{tabularx}{\linewidth}{@{}p{2.6cm}p{2.9cm}Xp{2.2cm}@{}}
\toprule
\textbf{Study}&\textbf{Benchmark}&\textbf{Reported result}&\textbf{Source location}\\
\midrule
LLM Compiler~\cite{cummins2023compiler}&Held-out LLVM-IR functions
&3.01\% fewer instructions than \texttt{-Oz}
&Table III\\
Magellan~\cite{chen2026magellan}&LLVM inlining heuristic
&5.23\% size improvement; 8.79\% in a production evaluation
&Section 3.1\\
AwareCompiler~\cite{lin2025awarecompiler}&Seven benchmark suites
&30.03\% IR-count reduction; \texttt{-Oz}: 29.98\%
&Table 1\\
UniPar~\cite{bitan2025unipar}&ParaTrans
&29.4\% (CUDA) and 38.2\% (OpenMP) validation rate
&RQ2\\
ACCeLLiuM~\cite{jhaveri2025accellium}&OpenACC pragma--loop pairs
&50.4\% exact match; 87.3\% directive-type agreement
&Introduction; RQ2\\
MPI analysis~\cite{piersall2026mpi}&Annotated MPI programs
&44.39--77.84\% detection accuracy; 125/164 repairs
&Table 2; Section 5.3\\
CompilerGPT~\cite{pirkelbauer2025compilergpt}&Five programs
&up to $6.5\times$ speedup
&Table 3\\
\bottomrule
\end{tabularx}
\end{table}

\begin{table}[tp]
\centering
\caption{HyperLoad forecast errors relative to the strongest alternative in each setting. Results
use DCData cooling-load traces with different horizons and training-set
sizes~\cite{jiang2025hyperload}. MSE and MAE are scaled by $10^3$ and $10^2$,
respectively; lower values are better and the better value in each pair is bold.
$\Delta$ is relative error change, so negative values indicate an improvement.}
\label{tab:r2-forecast}
\scriptsize
\setlength{\tabcolsep}{3.2pt}
\renewcommand{\arraystretch}{1.08}
\begin{fittable}\begin{tabular}{@{}llrrrrrr@{}}
\toprule
& & \multicolumn{3}{c}{\textbf{MSE} ($\times 10^{3}$)} & \multicolumn{3}{c}{\textbf{MAE} ($\times 10^{2}$)}\\
\cmidrule(lr){3-5}\cmidrule(lr){6-8}
\textbf{Train} & \textbf{Horizon} & Best base & HyperLoad & $\Delta$ &
Best base & HyperLoad & $\Delta$\\
\midrule
\multirow{4}{*}{100\%}
 & \phantom{0}12 & \phantom{0}6.1 & \textbf{\phantom{0}5.5} & $-9.8$\%  & \phantom{0}5.65 & \textbf{\phantom{0}5.39} & $-4.6$\%\\
 & \phantom{0}24 & \phantom{0}8.8 & \textbf{\phantom{0}8.4} & $-4.5$\%  & \phantom{0}6.61 & \textbf{\phantom{0}6.37} & $-3.6$\%\\
 & \phantom{0}48 & 14.4 & \textbf{12.4} & $-13.9$\% & \phantom{0}7.83 & \textbf{\phantom{0}7.48} & $-4.5$\%\\
 & \phantom{0}96 & 20.3 & \textbf{19.2} & $-5.4$\%  & \phantom{0}9.77 & \textbf{\phantom{0}9.31} & $-4.7$\%\\
\midrule
\multirow{2}{*}{50\%}
 & \phantom{0}12 & \phantom{0}6.2 & \textbf{\phantom{0}5.8} & $-6.5$\%  & \phantom{0}5.65 & \textbf{\phantom{0}5.44} & $-3.7$\%\\
 & \phantom{0}96 & 20.8 & \textbf{20.0} & $-3.8$\%  & \phantom{0}9.91 & \textbf{\phantom{0}9.50} & $-4.1$\%\\
\midrule
\multirow{2}{*}{25\%}
 & \phantom{0}12 & 16.8 & \textbf{14.0} & $-16.7$\% & \phantom{0}9.27 & \textbf{\phantom{0}8.39} & $-9.5$\%\\
 & \phantom{0}96 & 45.4 & \textbf{44.2} & $-2.6$\%  & 14.99 & \textbf{14.35} & $-4.3$\%\\
\bottomrule
\end{tabular}\end{fittable}
\end{table}

\begin{table}[t]
\centering
\caption{Performance across test-time budgets. EvolVE varies the number of search nodes on
VerilogEval-v2 and RTLLM-v2~\cite{hsin2026evolve}; KernelBench and GPT-5-RL vary the number of
inference calls or attempts~\cite{kernelbench2025,tehrani2026gpt5}; LASSI-EE varies the number of
sampled candidates $k$~\cite{dearing2026lassiee}. Metric values follow the listed budget order.
$\Delta$ is the endpoint change in the native metric, and $\Delta$/dec.\ is the change per tenfold
budget increase. Bold marks the better value of each matched pair or the best value within a
block.}
\label{tab:budget-ladder}
\footnotesize
\setlength{\tabcolsep}{4pt}
\renewcommand{\arraystretch}{1.08}
\begin{tabularx}{\textwidth}{@{}>{\raggedright\arraybackslash}p{0.5cm}>{\raggedright\arraybackslash}p{3.15cm}>{\raggedright\arraybackslash}p{2.35cm}>{\raggedright\arraybackslash}p{2.45cm}>{\raggedright\arraybackslash}X >{\raggedright\arraybackslash}p{0.85cm}>{\raggedright\arraybackslash}p{1.05cm}@{}}
\toprule
\textbf{Cat.} & \textbf{System / configuration} & \textbf{Metric} &
\textbf{Budget levels} & \textbf{Metric at each level} & \textbf{$\Delta$} &
\textbf{$\Delta$/dec.}\\
\midrule
\rowcolor{tabband}\multicolumn{7}{@{}l}{\textit{EvolVE}~\cite{hsin2026evolve}}\\
A1 & Gemini-2.0-Flash $+$ MCTS & VerilogEval-v2 pass & 15/50/100/200/300 nodes & \textbf{77.6} / \textbf{81.4} / \textbf{84.6} / \textbf{86.5} / \textbf{87.8} & \gain{10.2} & \gain{7.8}\\
A1 & Gemini-2.0-Flash, no search & VerilogEval-v2 pass & 15/50/100/200/300 nodes & 72.4 / 74.4 / 77.6 / 78.8 / 80.1 & \gain{7.7} & \gain{5.9}\\
A1 & GPT-OSS-120B $+$ IGR & VerilogEval-v2 pass & 15/50/100/200/300 nodes & 89.1 / \textbf{95.5} / \textbf{96.2} / \textbf{96.8} / \textbf{96.8} & \gain{7.7} & \gain{5.9}\\
A1 & GPT-OSS-120B, no search & VerilogEval-v2 pass & 15/50/100/200/300 nodes & \textbf{92.9} / 93.6 / 94.2 / 94.2 / 94.2 & \gain{1.3} & \gain{1.0}\\
A1 & DeepSeek-R1-FP4 $+$ MCTS & VerilogEval-v2 pass & 15/50/100/200/300 nodes & \textbf{94.2} / \textbf{96.8} / \textbf{96.8} / \textbf{97.4} / \textbf{98.1} & \gain{3.9} & \gain{3.0}\\
A1 & DeepSeek-R1-FP4, no search & VerilogEval-v2 pass & 15/50/100/200/300 nodes & 91.0 / 94.2 / 94.9 / 94.9 / 95.5 & \gain{4.5} & \gain{3.5}\\
A1 & DeepSeek-R1-FP4 $+$ MCTS & RTLLM-v2 pass & 15/50/100/200/300 nodes & 82.0 / 84.0 / 86.0 / 90.0 / 92.0 & \gain{10.0} & \gain{7.7}\\
\midrule
\rowcolor{tabband}\multicolumn{7}{@{}l}{\textit{KernelBench}~\cite{kernelbench2025} \textit{and GPT-5-RL}~\cite{tehrani2026gpt5}}\\
A2 & DeepSeek-R1, Level 2 & $\mathrm{fast}_1$, 100 tasks & 1 / 10 inference calls & 36 $\rightarrow$ \textbf{72} & \gain{36} & \gain{36.0}\\
A2 & DeepSeek-R1, Level 1 & $\mathrm{fast}_1$, 100 tasks & 1 / 10 inference calls & 12 $\rightarrow$ 43 & \gain{31} & \gain{31.0}\\
A2 & DeepSeek-V3, Level 1 & $\mathrm{fast}_1$, 100 tasks & 1 / 10 inference calls & \phantom{0}6 $\rightarrow$ 19 & \gain{13} & \gain{13.0}\\
A2 & DeepSeek-V3, Level 2 & $\mathrm{fast}_1$, 100 tasks & 1 / 10 inference calls & \phantom{0}4 $\rightarrow$ \phantom{0}6 & \gain{2} & \gain{2.0}\\
A2 & GPT-5-RL & functional \%, 264 tasks & 1 / 3 attempts & 77.0 $\rightarrow$ 83.7 & \gain{6.7} & \gain{14.0}\\
\midrule
\rowcolor{tabband}\multicolumn{7}{@{}l}{\textit{LASSI-EE}~\cite{dearing2026lassiee}}\\
R2 & LASSI-EE & ER@$k$, cross-device & $k =$ 1 / 3 / 5 candidates & \textbf{29.1} / \textbf{48.2} / \textbf{54.6} & \gain{25.5} & \gain{36.5}\\
R2 & vanilla LLM baseline & ER@$k$, cross-device & $k =$ 1 / 3 / 5 candidates & 10.5 / 26.3 / 34.6 & \gain{24.1} & \gain{34.5}\\
\bottomrule
\end{tabularx}
\end{table}

For budgets $b_1$ and $b_2$ with corresponding metric values $y_1$ and $y_2$,
the descriptive endpoint change per
decade is $(y_2-y_1)/\log_{10}(b_2/b_1)$. It does not estimate a fitted scaling
law or marginal return. EvolVE's displayed values come from Table~1 without
structured testbench generation~\cite{hsin2026evolve}. The GPT-5-RL~\cite{tehrani2026gpt5} three-attempt result is
221/264; its displayed 83.7\% is rounded.

\subsection{Cross-Category Evidence Maps}
\label{sec:appendix:evidence}

\begin{figure}[tp]
\centering
\includegraphics[width=\textwidth]{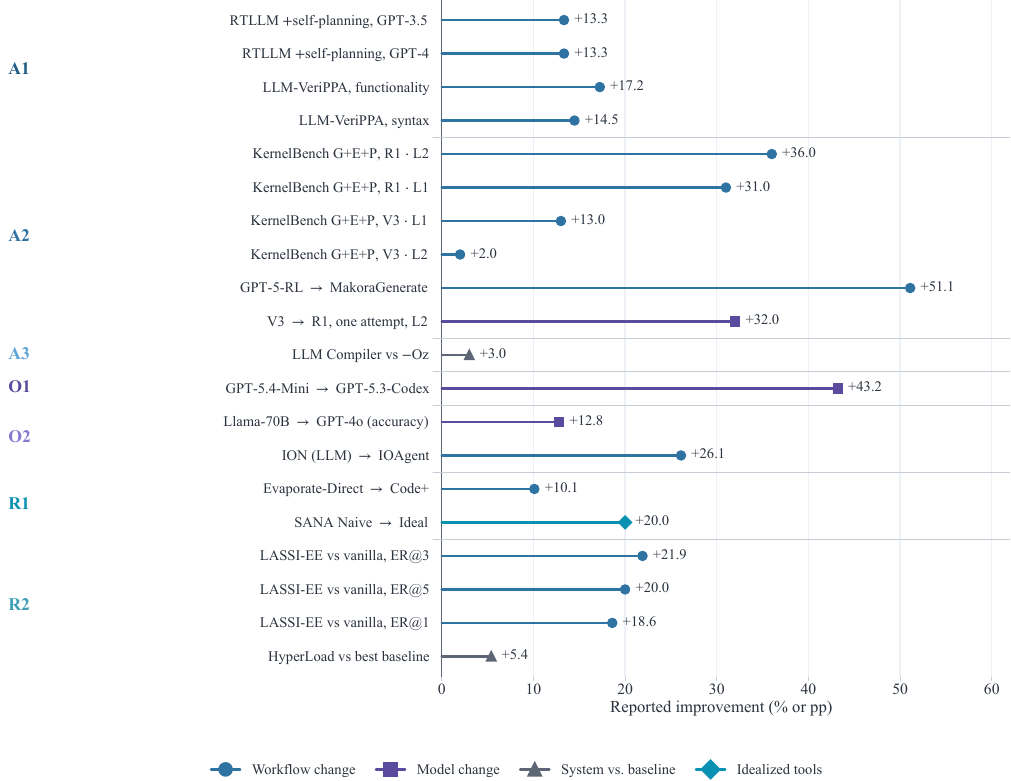}
\caption{Reported within-study improvements grouped by intervention type. Sources and evaluation
budgets are given in Table~\ref{tab:loop-closure}; SANA's idealized tools represent a
diagnostic condition~\cite{sana2026datalake}. Rate differences are percentage points,
while compiler and forecasting improvements are relative percentages. Positive values
indicate improvement in the native metric. Distances across unlike metrics do not
support a ranking of studies.}
\Description{A grouped dot-and-line plot shows study-specific improvements with separate symbols for workflow changes, model changes, baseline comparisons and idealized tools.}
\label{fig:loop-closure}
\end{figure}

Figure~\ref{fig:loop-closure} shows selected source-supported contrasts.
Percentage-point differences and relative percentage reductions retain their
native meanings. The figure does not infer causal effects from cross-study
values. Table~\ref{tab:loop-closure} identifies the quantities used.

\begin{table}[tp]
\centering\scriptsize
\caption{Within-study contrasts underlying Figure~\ref{fig:loop-closure}. Rate changes are
percentage points; compiler and forecast-error reductions are relative percentages. RTLLM
contrasts use five generations, VeriPPA a four-shot GPT-4 prompt, KernelBench contrasts ten
inference calls unless a single attempt is stated, LASSI-EE the stated $k$, and HyperLoad a
96-step horizon; SANA's tool comparison is diagnostic. No pooled effect or cross-metric ordering
is inferred.}
\label{tab:loop-closure}
\begin{tabularx}{\linewidth}{@{}p{.55cm}Xp{2.2cm}rrr@{}}
\toprule
\textbf{Cat.}&\textbf{Contrast / source}&\textbf{Metric}&\textbf{Before}&\textbf{After}&\textbf{Change}\\
\midrule
A1 & RTLLM $+$self-planning, GPT-3.5~\cite{lu2023rtllm} & functional, n/30 & 33.3 & 46.7 & +13.3 \\
A1 & RTLLM $+$self-planning, GPT-4~\cite{lu2023rtllm} & functional, n/30 & 50.0 & 63.3 & +13.3 \\
A1 & LLM-VeriPPA, functionality~\cite{thorat2025verippa} & functionality \% & 44.8 & 62.1 & +17.2 \\
A1 & LLM-VeriPPA, syntax~\cite{thorat2025verippa} & syntax \% & 66.9 & 81.4 & +14.5 \\
A2 & KernelBench G+E+P, R1 $\cdot$ L2~\cite{kernelbench2025} & $\mathrm{fast}_1$ & 36.0 & 72.0 & +36.0 \\
A2 & KernelBench G+E+P, R1 $\cdot$ L1~\cite{kernelbench2025} & $\mathrm{fast}_1$ & 12.0 & 43.0 & +31.0 \\
A2 & KernelBench G+E+P, V3 $\cdot$ L1~\cite{kernelbench2025} & $\mathrm{fast}_1$ & 6.0 & 19.0 & +13.0 \\
A2 & KernelBench G+E+P, V3 $\cdot$ L2~\cite{kernelbench2025} & $\mathrm{fast}_1$ & 4.0 & 6.0 & +2.0 \\
A2 & GPT-5-RL $\rightarrow$ MakoraGenerate~\cite{tehrani2026gpt5} & \% faster & 21.8 & 72.9 & +51.1 \\
A2 & V3 $\rightarrow$ R1, one attempt, L2~\cite{kernelbench2025} & $\mathrm{fast}_1$ & 4.0 & 36.0 & +32.0 \\
A3 & LLM Compiler vs $-$Oz~\cite{cummins2023compiler} & instr. count & --- & --- & +3.0 \\
O1 & GPT-5.4-Mini $\rightarrow$ GPT-5.3-Codex~\cite{wang2026deploybench} & pass rate \% & 7.8 & 51.0 & +43.2 \\
O2 & Llama-70B $\rightarrow$ GPT-4o (accuracy)~\cite{egersdoerfer2026ioagent} & ranker score $\times$100 & 51.3 & 64.1 & +12.8 \\
O2 & ION (LLM) $\rightarrow$ IOAgent~\cite{egersdoerfer2026ioagent} & ranker score $\times$100 & 38.0 & 64.1 & +26.1 \\
R1 & Evaporate-Direct $\rightarrow$ Code+~\cite{arora2023evaporate} & OpenIE F1 & 56.6 & 66.7 & +10.1 \\
R1 & SANA Naive $\rightarrow$ Ideal, LakeQA~\cite{sana2026datalake} & semantic match \% & 56.3 & 76.3 & +20.0 \\
R2 & LASSI-EE vs vanilla, ER@3~\cite{dearing2026lassiee} & ER@3 \% & 26.3 & 48.2 & +21.9 \\
R2 & LASSI-EE vs vanilla, ER@5~\cite{dearing2026lassiee} & ER@5 \% & 34.6 & 54.6 & +20.0 \\
R2 & LASSI-EE vs vanilla, ER@1~\cite{dearing2026lassiee} & ER@1 \% & 10.5 & 29.1 & +18.6 \\
R2 & HyperLoad vs best baseline~\cite{jiang2025hyperload} & relative MSE & --- & --- & +5.4 \\
\bottomrule
\end{tabularx}
\end{table}

\begin{figure}[tp]
\centering
\includegraphics[width=\textwidth]{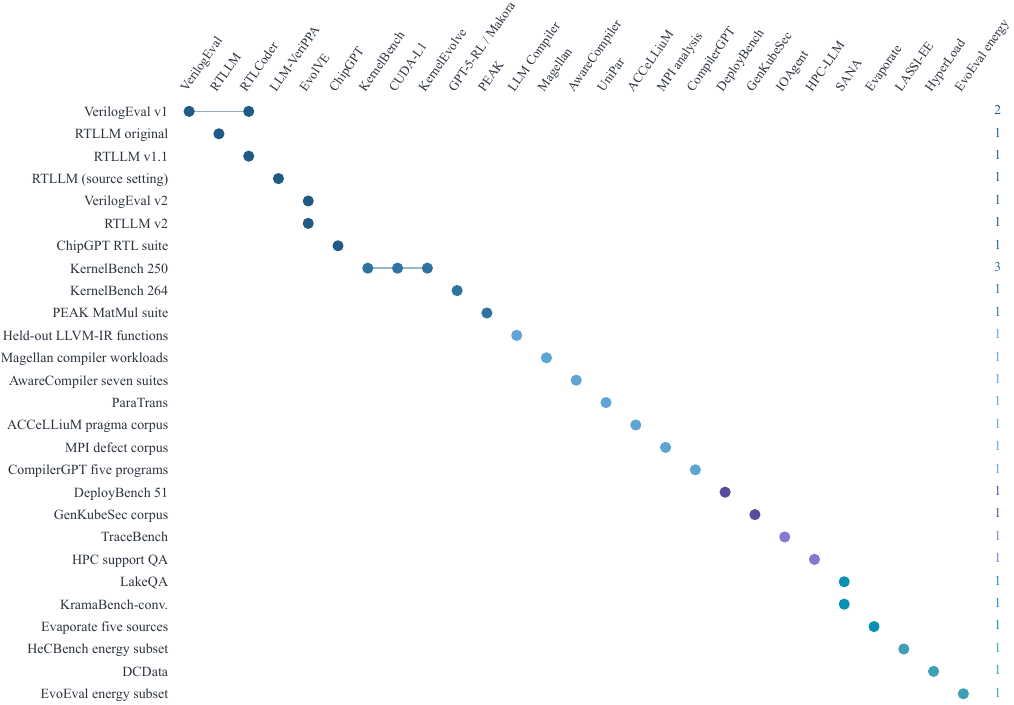}
\caption{Benchmark-family associations in the selected evidence. Each column represents a
source paper, and each marker links a paper to an evaluated benchmark family or a
study-specific task collection. Family reuse does not imply identical versions, tasks,
or execution conditions.}
\Description{A benchmark-by-paper matrix shows source associations for the selected evidence.}
\label{fig:benchmark-usage}\label{tab:benchmarks}
\end{figure}

Figure~\ref{fig:benchmark-usage} records benchmark-family associations by
source paper. The GPT-5-RL and MakoraGenerate~\cite{tehrani2026gpt5} configurations are treated as one
study, while distinct custom HPC collections remain separate. This author-coded
map summarizes the selected evidence set, not the prevalence of practices
across the entire literature.

\FloatBarrier
\subsection{Research Agenda}
\label{sec:appendix:agenda}

\paragraph{Validation beyond a single test configuration.}
Artifact benchmarks should document shapes, data types, tolerances, program
inputs, and concurrency settings. Perturbed inputs and independent validation
can reveal errors that a single reference configuration misses. Formal methods
can strengthen selected guarantees, while their assumptions and coverage remain
part of the reported result.

\paragraph{Operational outcomes and recovery.}
Diagnosis reports should be complemented by evaluations of executed actions
where appropriate. Recovery time, service impact, failed interventions, and
rollback outcomes provide evidence that rank-based report assessment cannot
supply. Shared telemetry should retain enough environmental context to make
incident interpretation reproducible.

\paragraph{Uncovered training and serving concerns.}
LLM agents have not yet been applied to checkpoint frequency and placement,
straggler mitigation, the joint choice of quantization and parallelism for a
deployment, training data loaders, or power capping of AI jobs
(Table~\ref{tab:ai-workloads}). Each has a natural evaluator, such as goodput
under failure traces, step-time variance, SLO attainment, input-pipeline
stalls, or energy per token, and each is currently addressed by non-LLM
optimizers that could serve as baselines.

\paragraph{Evaluator fidelity for orchestration.}
Workload-orchestration results should report the fidelity of the simulator or
projection engine against measured runs, the trace or cluster used, and the
result of a strong non-LLM optimizer given the same evaluator. Policies
discovered in simulation should be validated on real clusters, and their
behavior under workload shift should be measured, since an evolved policy can
exploit evaluator artifacts.

\paragraph{Cross-stage hand-offs.}
The lifecycle map shows no evaluated hand-off between accelerator design and
kernel engineering or between data supply and power. Benchmarks that carry an
artifact across stages, for example generating a kernel for a newly designed
accelerator and measuring end-to-end training time, would test whether LLM
methods can coordinate decisions that are currently optimized in isolation.

\paragraph{Resource and lifecycle accounting.}
Data and energy evaluations should report output quality, runtime, resource
consumption, and the cost of model inference and search. Measuring a complete
generate--validate--deploy--operate workflow would establish whether local
artifact gains persist after integration and amortize their development cost.
Cross-platform evaluations should report native results for each device before
summarizing them across platforms.

\end{document}